%% file: main.tex
\documentclass[10pt,journal,compsoc]{IEEEtran}

\ifCLASSOPTIONcompsoc
  \usepackage[nocompress]{cite}
\else
  \usepackage{cite}
\fi

\usepackage{blindtext}

\usepackage{amsmath,amssymb,amsfonts,bm,bbm}
\DeclareMathAlphabet\mathbfcal{OMS}{cmsy}{b}{n}
\usepackage{lipsum}
\usepackage{times}
\usepackage{epsfig}

\usepackage{stfloats}
\usepackage{multicol}
\usepackage[inline]{enumitem}
\usepackage{algorithmic}
\usepackage{graphicx}
\usepackage{textcomp}
\usepackage{caption}
\usepackage{subcaption}
\usepackage{marvosym}
\usepackage{cite}
\usepackage{mathtools}

\usepackage{booktabs}
\usepackage{multirow}
\usepackage[pagebackref=true,breaklinks=true,colorlinks,bookmarks=true,bookmarksopen=true,citecolor=blue]{hyperref}

\usepackage[utf8]{inputenc}
\usepackage[T1]{fontenc}
\usepackage{float}
\usepackage{adjustbox}
\usepackage{array}
\usepackage{dsfont}
\usepackage{icomma} 

\usepackage{paralist}
\usepackage[ruled,linesnumbered]{algorithm2e}

\usepackage[accsupp]{axessibility}  

\usepackage{tabulary}
\usepackage[table]{xcolor}

\definecolor{gray}{rgb}{0.3,0.3,0.3}
\definecolor{blue}{rgb}{0,0.5,1}
\definecolor{mask_red}{rgb}{1,0,0.8}
\definecolor{green}{rgb}{0.2,1,0.2}
\definecolor{rblue}{rgb}{0,0,1}
\definecolor{lightblue}{HTML}{6495ed}
\definecolor{lightred}{HTML}{F19C99}

\newcommand{\green}[1]{\textcolor[RGB]{96,177,87}{#1}}

\newcommand{\fn}[1]{\footnotesize{#1}}
\newcommand{\gbf}[1]{\green{\bf{\fn{(#1)}}}}

\definecolor{graytablerow}{gray}{0.6}

\definecolor{revised_color}{HTML}{0066CC}
\definecolor{revised_color_ZJM}{HTML}{9932CC}
\definecolor{revised_color_PKY}{HTML}{FF2E82}
\definecolor{revised_color_SH}{HTML}{007FFF}

\usepackage{tikz}
\usetikzlibrary{plotmarks}
\newcommand\marksymbol[2]{\tikz[#2,scale=1.5]\pgfuseplotmark{#1};}
\newcommand*\circled[1]{\tikz[baseline=(char.base)]{
\node[shape=circle,fill=gray,inner sep=0.5pt] (char) {\textcolor{white}{\footnotesize \textbf{#1}}};}}
\newcommand*\circledbrown[1]{\tikz[baseline=(char.base)]{
\node[shape=circle,fill=brown,inner sep=0.5pt] (char) {\textcolor{white}{\footnotesize \textbf{#1}}};}}

\usepackage{ragged2e}

\makeatletter
\DeclareRobustCommand\onedot{\futurelet\@let@token\@onedot}
\def\@onedot{\ifx\@let@token.\else.\null\fi\xspace}

\def\eg{\emph{e.g}\onedot} 
\def\ie{\emph{i.e}\onedot} 
 
 \def\vs{\emph{vs}\onedot}

\begin{document}

\title{Behind Every Domain There is a Shift:\\Adapting Distortion-aware Vision Transformers for Panoramic Semantic Segmentation
}

\author{Jiaming Zhang, Kailun Yang\IEEEauthorrefmark{1}, Hao Shi, Simon Reiß, Kunyu Peng, Chaoxiang Ma, Haodong Fu,\\Philip H. S. Torr, Kaiwei Wang, and Rainer Stiefelhagen
\IEEEcompsocitemizethanks{\IEEEcompsocthanksitem Jiaming Zhang, Simon Reiß, Kunyu Peng, and Rainer Stiefelhagen are with Karlsruhe Institute of Technology, Germany.
\IEEEcompsocthanksitem Kailun Yang is with Hunan University, China.
\IEEEcompsocthanksitem Jiaming Zhang and Philip H. S. Torr are with University of Oxford, UK.
\IEEEcompsocthanksitem Hao Shi and Kaiwei Wang are with Zhejiang University, China.
\IEEEcompsocthanksitem Chaoxiang Ma is with ByteDance Inc., China.
\IEEEcompsocthanksitem Haodong Fu is with Beihang University, China.
\IEEEcompsocthanksitem \IEEEauthorrefmark{1} corresponding author (kailun.yang@hnu.edu.cn).
}
}

\IEEEtitleabstractindextext{%
\begin{abstract} \justifying
\input{Tex_content/abstract}
\end{abstract}

}

\maketitle

\IEEEdisplaynontitleabstractindextext

\IEEEpeerreviewmaketitle

\IEEEraisesectionheading{\section{Introduction}\label{sec:introduction}}

\begin{figure*}[!t]
	\centering
    \begin{minipage}[t]{.6\textwidth}
    \centering
    \includegraphics[width=1\textwidth]{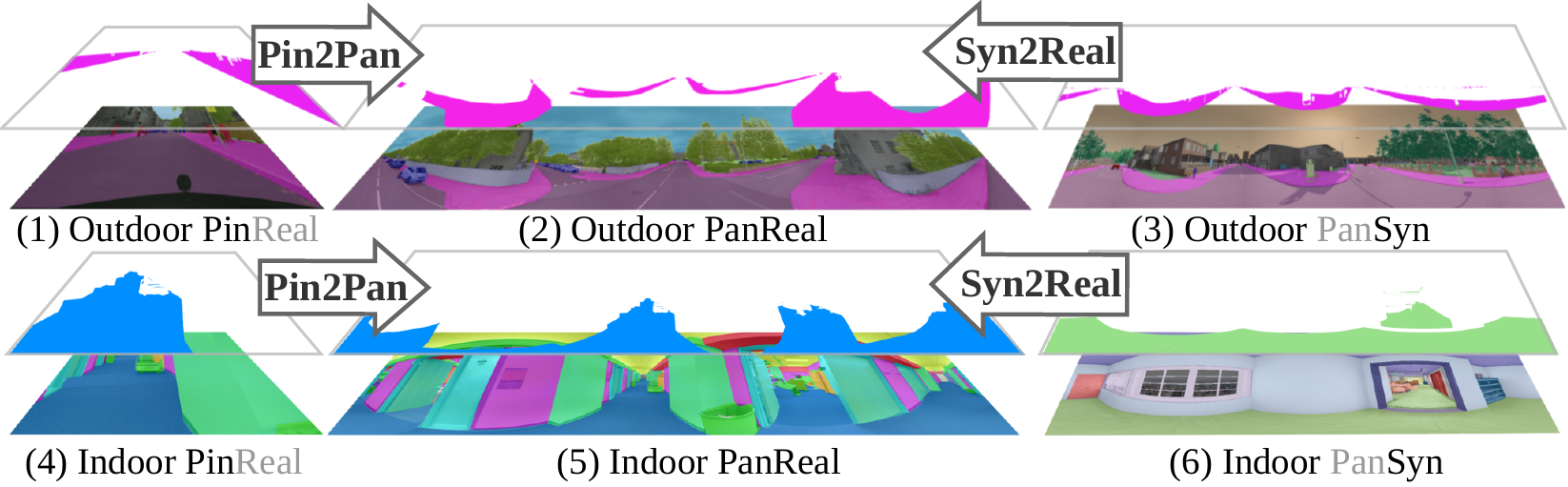}
    \subcaption{{Domain adaptation paradigms (\textsc{Pin2Pan} and \textsc{Syn2Real})}}\label{fig:da_settings1}
    \end{minipage}%
    \begin{minipage}[t]{.18\textwidth}
    \centering
    \includegraphics[width=1\textwidth]{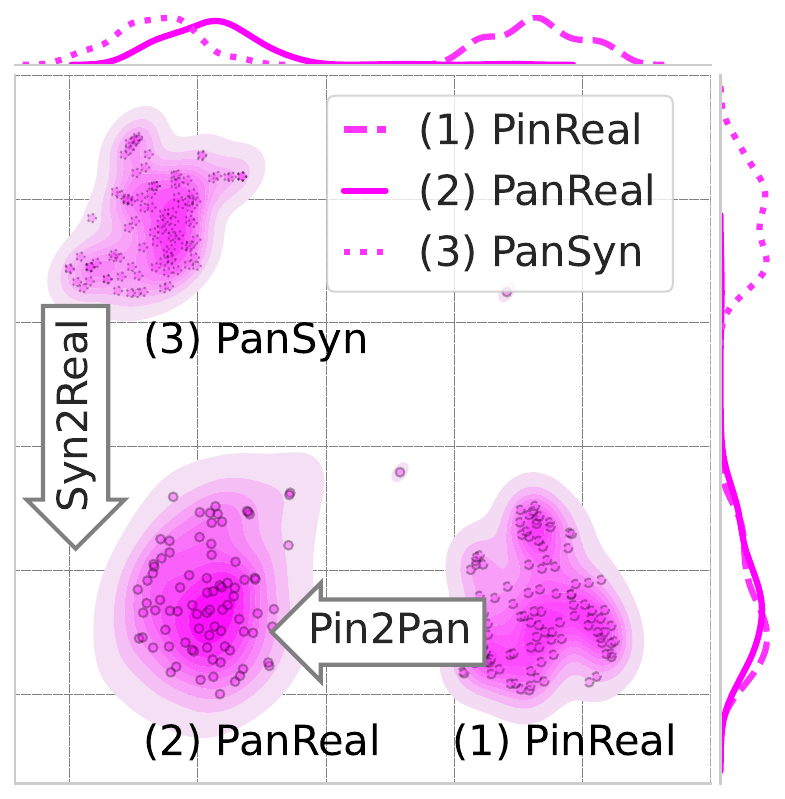}
    \subcaption{Distribution \textit{sidewalk}}\label{fig:da_settings2}
    \end{minipage}%
    \begin{minipage}[t]{.18\textwidth}
    \includegraphics[width=1\textwidth]{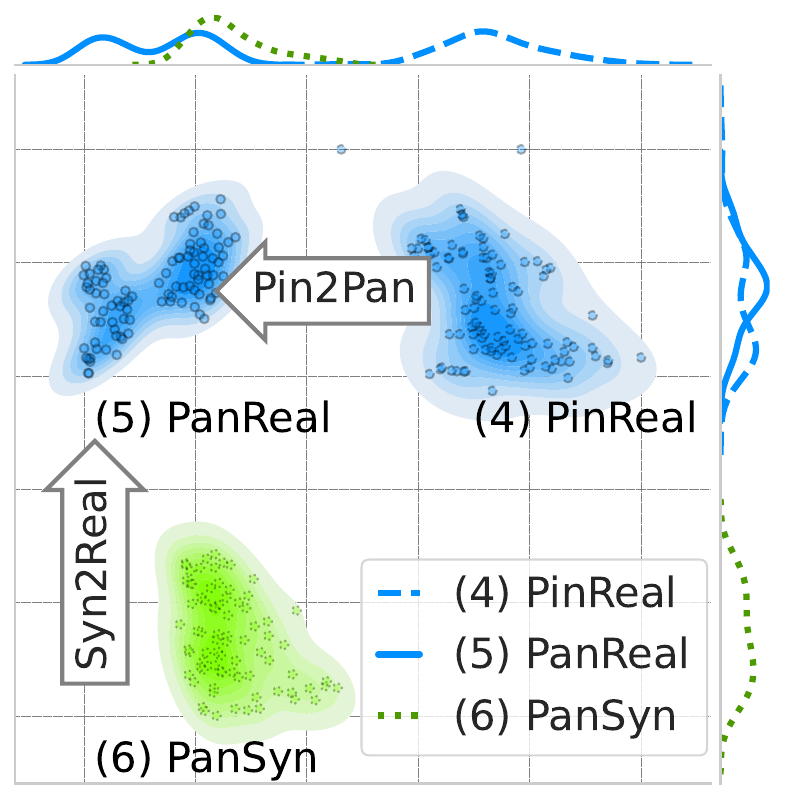}
    \subcaption{Distribution \textit{floor}}\label{fig:da_settings3}
    \end{minipage}%
	\caption{\small \textbf{Domain adaptations for panoramic semantic segmentation} include Pinhole-to-Panoramic (\textsc{Pin2Pan}) and Synthetic-to-Real (\textsc{Syn2Real}) paradigms in both indoor and outdoor scenarios. 
	The feature distributions between the target domain and two source domains are compared in the tSNE-reduced manifold space, including \textit{sidewalks} and \textit{floors}. The marginal distributions are plotted along respective axes.
	} 
    \label{fig:da_settings}
\end{figure*}

\IEEEPARstart{P}{anoramic} semantic segmentation offers an omnidirectional and dense visual understanding regimen that integrates $360^\circ$ perception of surrounding scenes and pixel-wise predictions of input images~\cite{wildpass}. 
{The attracted attention of $360^\circ$ cameras is manifesting, with an increasing number of learning systems and practical applications, such as holistic sensing in autonomous vehicles~\cite{Garanderie_2018_ECCV,densepass,gao2022review} and immersive viewing in augmented- and virtual reality (AR/VR) devices~\cite{xu2018predicting_head_movement,xu2021spherical,ai2022deep_omnidirectional_survey}.
In contrast to images captured by pinhole cameras (see Fig.~\ref{fig:da_settings1}-(1)\&(4)) with a narrow Field of View~(FoV), panoramic images with an ultra-wide FoV of $360^\circ$, deliver complete scene perception in outdoor driving environments (Fig.~\ref{fig:da_settings1}-(2)) and indoor scenarios (Fig.~\ref{fig:da_settings1}-(5)).}

\begin{figure}[!t]
	\centering
    \begin{minipage}[t]{.5\columnwidth}
    \centering
    \includegraphics[width=0.95\columnwidth]{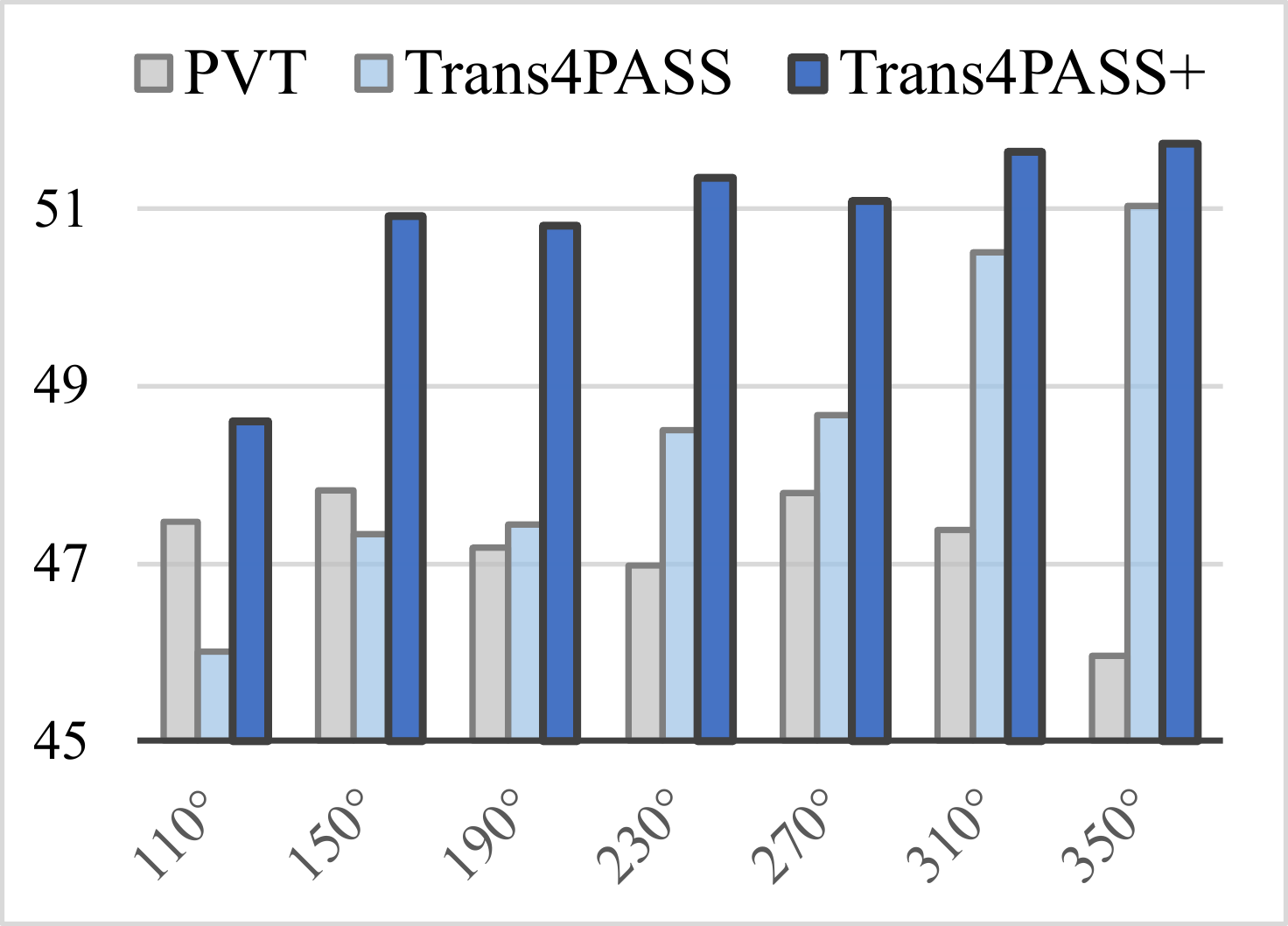}
    \vskip -1ex
    \subcaption{\small Performance of \textsc{Pin2Pan}}\label{fig:fov_1}
    \end{minipage}%
    \begin{minipage}[t]{.5\columnwidth}
    \centering
    \includegraphics[width=0.95\columnwidth]{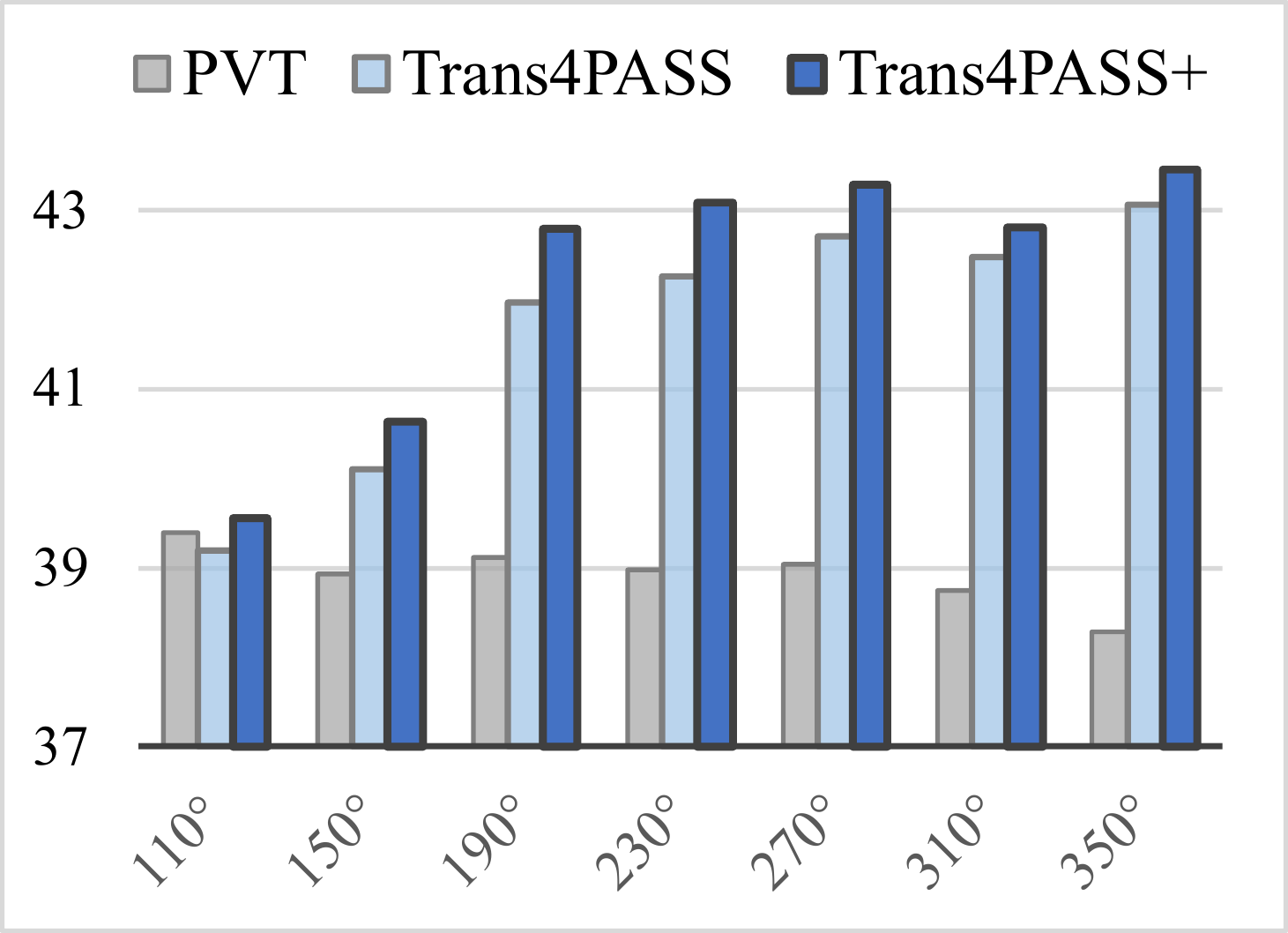}
    \vskip -1ex
    \subcaption{\small Performance of \textsc{Syn2Real}}\label{fig:fov_2}
    \end{minipage}%
    \vskip -2ex
	\caption{\small {Model performance (mIoU) against changes in Fields of View (FoV) in both (a) Pinhole-to-Panoramic (\textsc{Pin2Pan}) and (b) Synthetic-to-Real (\textsc{Syn2Real}) settings. Trans4PASS+ models perform stably.}
    }
    \label{fig:fov}
\end{figure}

\begin{figure}[!t]
    \vskip -1ex
	\centering
    \includegraphics[width=1.0\columnwidth]{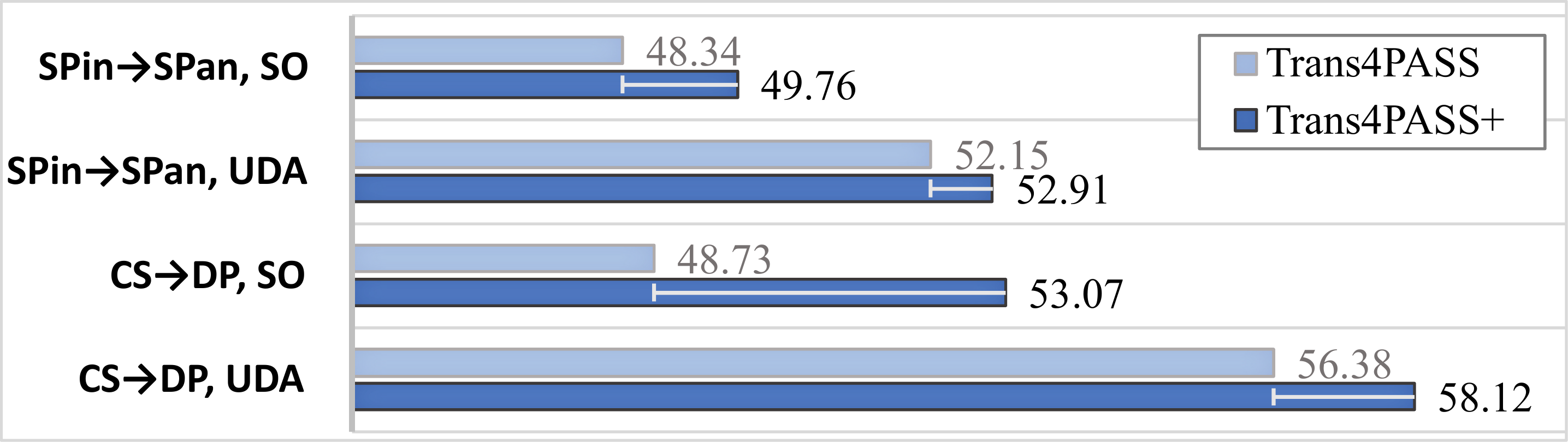}
    \vskip -2ex
	\caption{{\small Performance gains of Trans4PASS+ in the settings of Source-Only (SO) and panoramic Unsupervised Domain Adaptation (UDA), transferring from Stanford2D3D Pinhole to Panoramic (SPin{\MVRightarrow}SPan) and from Cityscapes to DensePASS (CS{\MVRightarrow}DP) domains.}
	} 
    \label{fig:plus}
    \vskip -3ex
\end{figure}

However, panoramic images often have large image distortions and object deformations due to the intrinsic equirectangular projection~\cite{hohonet,omnirange}. 
This renders a vast number of methods a sub-optimal solution for panoramic segmentation~\cite{pass,zhang2022trans4pass}, as they are tailored for pinhole images and cannot handle severe deformations. {In Fig.~\ref{fig:fov}, we observed that with an increase in FoV, the performance of the baseline model~\cite{pvt} drops.}
To solve this problem, we propose a novel distortion-aware model, \ie, \emph{Transformer for PAnoramic Semantic Segmentation (Trans4PASS)}.
{Specifically, compared with standard Patch Embedding (PE), %
our newly designed \emph{Deformable Patch Embedding (DPE)} %
helps to learn the prior knowledge of panorama characteristics during patchifying the image.} In addition, the proposed \emph{Deformable MLP (DMLP)} enables the model to better adapt to panoramas during feature parsing.
{Building upon the success of the deformable structure, we further propose a simple yet highly effective version, \ie, Trans4PASS+, which is enhanced by DMLPv2 with parallel token mixing mechanisms. %
{Compared to Trans4PASS, the new model can benefit more from larger FoV thanks to its advanced token mixing, showing inherent robustness against FoV changes in both \textsc{Pin2Pan} and \textsc{Syn2Real} settings (Fig.~\ref{fig:fov}). Meanwhile, it shows significant performance gains in the source-only setting (Fig.~\ref{fig:plus}) in both indoor (SPin{\MVRightarrow}SPan) and outdoor (CS{\MVRightarrow}DP) scenarios.}} %

Apart from the deformation of panoramic images, the scarcity of annotated data is another key difficulty that hinders the progress of panoramic semantic segmentation. Notoriously, it is extremely time-consuming and expensive to produce dense annotations %
for training success~\cite{cityscapes,setr}, and this difficulty is further exacerbated for panoramas with ultra-wide FoV and many small and distorted scene elements concurrently appearing in complex environments. {Unsupervised Domain Adaptation (UDA) is a strategy commonly employed to adapt models from a source domain to a target domain, with its extensive application observed in pinhole imagery. However, UDA in the panoramic domain remains limited within the existing literature.}
In this work, we propose a \emph{Mutual Prototypical Adaptation (MPA)} strategy for domain adaptive panoramic segmentation. Compared with other adversarial-learning~\cite{clan} and pseudo-label self-learning~\cite{crst} methods, the advantage of MPA is that mutual prototypes are generated from both source and target domains. In this manner, large-scale labeled source data and unlabeled target data are taken into account at the same time.
{MPA can further unleash the potential of our adapted model when combining stronger mask generators (\eg, SAM~\cite{kirillov2023SAM}), which alleviates the negative effect of target samples with noisy and incomplete pseudo labels. It enables our unsupervised models to outperform SAM combined with self-supervised learning (SSL) or superior to previous fully-supervised models. Furthermore, this SAM-based study shows that MPA can be flexibly adopted with different segmenters. More comparisons are presented in Sec.~\ref{sec:ablation_unsupervised_domain_adaptation}.}

Based on the MPA strategy, we first revisit the Pinhole-to-Panoramic (\textsc{Pin2Pan}) paradigm as previous works~\cite{densepass,omnirange}, by considering the label-rich pinhole images as the source domain and the label-scare panoramic images as the target domain. {Furthermore, a new dataset (SynPASS) with $9,080$ synthetic panoramic images is created, which %
brings two benefits}: (1) The large-scale annotations enable training data-hungry models for panoramic semantic segmentation; (2) A new Synthetic-to-Real (\textsc{Syn2Real}) domain adaptive panoramic segmentation scheme is established apart from \textsc{Pin2Pan}. {Based on the new dataset, we thoroughly investigate the two adaptation paradigms in both indoor and outdoor scenarios (Fig.~\ref{fig:da_settings1}).} %
{The feature distributions of the \emph{sidewalk} class from two sources (S1, S2) and one target (T) domains are presented in Fig.~\ref{fig:da_settings2}, and the \emph{floor} class from indoor domains are in Fig.~\ref{fig:da_settings3}.} Upon close inspection, two insights become clear: (1) The marginal distributions of the synthetic and real domains are close in one dimension (\eg, the shape), whereas the marginal distributions in another dimension (\eg, the appearance) are far apart. (2) The patterns are reversed between the pinhole and panoramic domains. The insights are intuitive and consistent with common observations, as objects (\eg, \emph{sidewalks} or \emph{floors}) in synthetic- and real images are shape-deformed, while real pinhole- and panoramic images are similar in appearance. We unfold a comprehensive discussion and results in Sec.~\ref{sec:pin2pan_and_syn2real_adaptation}. 

{Extensive experiments -- both indoor and outdoor scenarios and each investigated under \textsc{Pin2Pan} and \textsc{Syn2Real} paradigms -- demonstrate the superiority of the proposed distortion-aware architecture.}
{Our new Trans4PASS+ model with the MPA strategy attains state-of-the-art performances on four panoramic segmentation benchmarks.}
{On the
Stanford2D3D dataset~\cite{stanford2d3d}, our unsupervised model outperforms the fully-supervised methods for the first time.
On the
Structured3D dataset~\cite{structured3d}, our \textsc{Syn2Real}-adapted model surpasses the model trained with extra $1,400$ annotated data.}
{On the
DensePASS dataset~\cite{densepass}, our source-only model obtains $49.94\%$ in mIoU with a ${+}10.92\%$ gain over the baseline, %
and our \textsc{Pin2Pan}-adapted model obtains $59.43\%$ in mIoU with a ${+}17.44\%$ boost over the previous best method~\cite{p2pda_trans}.}

{This work is built upon our previous conference version~\cite{zhang2022trans4pass} by introducing an improved model architecture, a new panoramic segmentation benchmark, a SAM-enhanced adaptation method, and a more comprehensive study on panoramic semantic segmentation with various UDA schemes.} 
At a glance, the additional contributions of this work can be summarized as follows: %
\begin{compactitem}
\item[(1)] {A new panoramic semantic segmentation benchmark \emph{SynPASS} is established with $9,080$ images. It delivers an alternative Synthetic-to-Real (\textsc{Syn2Real}) UDA paradigm in panoramic segmentation, which is compared with the Pinhole-to-Panoramic (\textsc{Pin2Pan}) one.}
\item[(2)] {We advance the Trans4PASS+ model with a more lightweight yet effective decoder, which is extended with a DMLPv2 module with parallel token mixing mechanisms to reinforce the flexibility in modeling discriminative information.}
\item[(3)] {We present a \emph{Mutual Prototypical Adaptation (MPA)} strategy for domain adaptive panoramic segmentation via dual-domain prototypes. For the first time, we boost MPA by using the Segment Anything Model (SAM) as a pseudo-label rectification strategy, which shows better performance over standalone SAM or combined with the SSL method.}
\item[(4)] We conduct more comprehensive comparative experiments. Our proposed method outperforms recent state-of-the-art token mixing~\cite{yu2021metaformer,cyclemlp,asmlp,zhou2022fan}, deformable patch-based learning~\cite{chen2021dpt}, transformer domain adaptation~\cite{hoyer2021daformer}, and panoramic segmentation~\cite{pass,omnirange,densepass,p2pda_trans} methods.
\item[(5)] {On four panoramic segmentation datasets, our framework yields superior results, spanning indoor and outdoor scenes, before and after \textsc{Pin2Pan} and \textsc{Syn2Real} adaptation.}

\end{compactitem}

\section{Related Work}
\label{sec:relaed_work}
\input{Tex_content/related_work}

\section{Methodology}
\label{sec:methodology}
Here, we detail the proposed panoramic semantic segmentation framework in the following structure: the \emph{Trans4PASS+} architecture in Sec.~\ref{sec:trans4pass}; the \emph{deformable patch embedding} module in Sec.~\ref{sec:dpe}; two \emph{deformable MLP} variants in Sec.~\ref{sec:dmlp}; and the \emph{mutual prototypical adaptation} in Sec.~\ref{sec:mpa}.

\begin{figure*}[!t]
	\centering
    \includegraphics[width=1.0\textwidth]{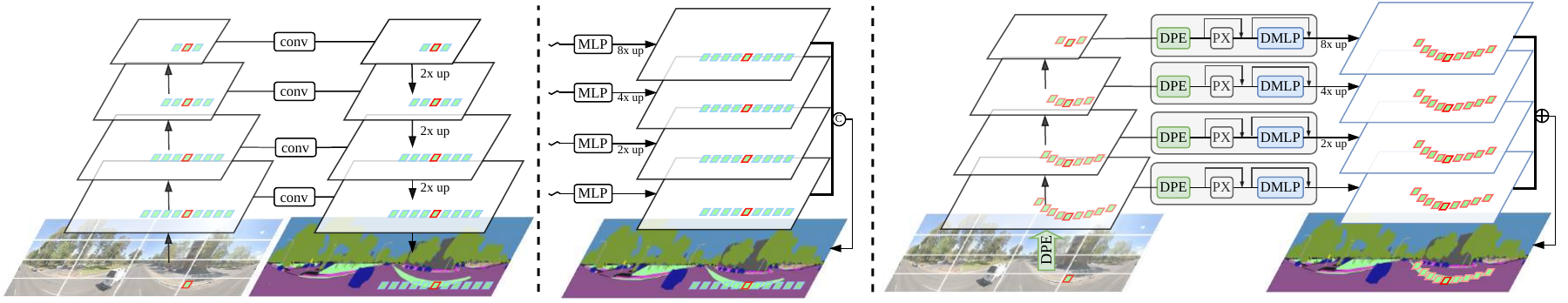}
    \begin{minipage}[t]{.3\textwidth}
    \centering
    \vskip -3ex
    \subcaption{Transformer with FPN-like decoder}\label{fig:fpn}
    \end{minipage}%
    \begin{minipage}[t]{.3\textwidth}
    \centering
    \vskip -3ex
    \subcaption{Transformer with vanilla-MLP}\label{fig:vanilla-mlp}
    \end{minipage}%
    \begin{minipage}[t]{.4\textwidth}
    \centering
    \vskip -3ex
    \subcaption{Trans4PASS with DPE and DMLP}\label{fig:trans4pass}
    \end{minipage}%
    \vskip -2ex
	\caption{\small \textbf{Comparison of segmentation transformers.}
	{Transformers (a) borrow an FPN-like decoder~\cite{setr} from CNN counterparts or (b) adopt a vanilla-MLP decoder~\cite{segformer} for feature fusion.}
	(c) \emph{Trans4PASS} integrates Deformable Patch Embeddings~(DPE) and the Deformable MLP~(DMLP) module for capabilities to handle distortions (see warped \emph{terrain}) and mix patches.
	} 
	\label{fig:decoder_structures}
\vskip -2ex
\end{figure*}

\subsection{Trans4PASS+ Architecture}
\label{sec:trans4pass}
As newly emerged learning architectures, transformer models are evolving and have attained outstanding performance in vision tasks~\cite{vit,setr}. In this work, we put forward a novel distortion-aware Trans4PASS architecture in order to explore the panoramic semantic segmentation task. Considering the trade-off between efficiency and accuracy, there are two different model sizes: the tiny (T) model and the small (S) model. Following traditional CNN/Transformer models~\cite{resnet,pvt,segformer}, both versions of Trans4PASS keep the multi-scale pyramid feature structure in the form of four stages.
{The layer numbers of four stages in the tiny model are $\{2, 2, 2, 2\}$, while in the small model, they are $\{3, 4, 6, 3\}$. In one segmentation process, given an input image in the shape of $H{\times}W{\times}3$, the Trans4PASS model first performs image patchifying.}
The encoder gradually down-samples feature maps $\boldsymbol{f}_l{\in}\{\boldsymbol{f}_1,\boldsymbol{f}_2,\boldsymbol{f}_3,\boldsymbol{f}_4\}$ in the $l^{th}$ stage with strides $s_l{\in}\{4, 8, 16, 32\}$ and channel dimensions $C_l{\in}\{64,128,320,512\}$. Then, the decoder parses multi-scale feature maps $\boldsymbol{f}_l$ into a unified shape of $\frac{H}{4}{\times}\frac{W}{4}{\times}C_{emb}$, where the number of resulting embedding channels is set as $C_{emb}{=}128$. Finally, a prediction layer outputs the final semantic segmentation result according to the number of semantic classes of the respective task, and with the same size as the input image.

However, the raw $360^\circ$ data is generally formulated in the spherical coordinate system (the latitude $\theta{\in}[0, 2\pi)$ and longitude $\phi {\in} [-\frac{1}{2} \pi, \frac{1}{2} \pi]$). To convert it to the Cartesian coordinate system (the x- and y-axes), the equirectangular projection in Eq.~(\ref{eq:equi_proj}) is commonly used to transfer $360^\circ$ data as a 2D flat panorama.
\begin{equation}\label{eq:equi_proj}
\begin{aligned}
    \left \{
    \begin{array}{rcl}
    x & = & (\theta-\theta_0)\cos\phi_1,\\
    y & = & (\phi-\phi_1),\\
    \end{array} \right.
\end{aligned}
\end{equation}
where $(\theta_0$, $\phi_1){=}(0, 0)$ is the central latitude and central longitude.

{Considering the simple equirectangular projection from Eq.~(\ref{eq:equi_proj}) as $x{=}\theta$ and $y{=}\phi$, the Area Distortion (AD) is approximated by the Jacobian determinant~\cite{lai2017semantic} in Eq.~(\ref{eq:jacobian}) and Eq.~(\ref{eq:distortion}).} 
\begin{equation}\label{eq:jacobian}
\begin{aligned}
\mathcal{J}(\theta, \phi)=\begin{vmatrix}
\frac{\partial(x)}{\partial(\theta)} & \frac{\partial(x)}{\partial(\phi)} \\
\frac{\partial(y)}{\partial(\theta)} & \frac{\partial(y)}{\partial(\phi)} \\
\end{vmatrix}.
\end{aligned}
\end{equation}
\begin{equation}\label{eq:distortion}
\begin{aligned}
\textbf{AD}(x, y)=\frac{\cos(\phi)|d\theta d\phi|}{|dxdy|}=\frac{\cos(\phi)}{\begin{vmatrix}
\mathcal{J}(\theta, \phi)\end{vmatrix}}.
\end{aligned}
\end{equation}
The AD is associated with $\cos(\phi)$. Thus, the areas ($\phi{\neq}0$) located in any panoramic image all include object distortions and deformations. 
{These observations motivate us to design a distortion-aware vision transformer model for parsing panoramic scenes.}

Compared with previous state-of-the-art segmentation transformers~\cite{setr,segformer} shown in Fig.~\ref{fig:fpn} and Fig.~\ref{fig:vanilla-mlp}, our Trans4PASS model (Fig.~\ref{fig:trans4pass}) is able to address the severe distortions in panoramas via two vital designs: (1) a \emph{Deformable Patch Embedding~(DPE)} module is proposed and applied in the encoder and decoder, enabling the model to extract and parse the feature hierarchy uniformly; (2) a \emph{Deformable MLP (DMLP)} module is proposed to better collaborate with DPE in the decoder, by adaptively mixing and interpreting the feature token extracted via DPE. Furthermore, a new DMLPv2 module is constructed with a parallel token mixing mechanism. Based on DMLPv2, our architecture is upgraded to Trans4PASS+, being more lightweight yet more effective for panoramic semantic segmentation. The DPE and DMLPs are detailed in the following sections.

\subsection{Deformable Patch Embedding}
\label{sec:dpe}
\noindent\textbf{Preliminaries on Patch Embedding.} 
{Given a 2D $C_{in}$-channel input image or intermediate feature map $\boldsymbol{f}{\in}\mathbb{R}^{H{\times}W{\times}C_{in}}$, the standard Patch Embedding (PE) module reshapes it into a sequence of flattened patches $\boldsymbol{z}{\in}\mathbb{R}^{(\frac{HW}{s^2}){\times}(s^2 \cdot C_{in})}$, where $(H, W)$ is the size of the input, $(s, s)$ is the size of each patch, and $\frac{HW}{s^2}$ is the number of patches (\ie, the length of the patch sequence).} Each element in this sequence is passed through a trainable linear projection layer transforming it into $C_{out}$ dimensional embeddings. The number of input channels $C_{in}$ is equal to the one of output channels $C_{out}$ in a typical patchifying process.

Consider one patch in $\boldsymbol{z}$ representing a rectangle area $s{\times}s$ with $s^2$ positions. The position offset relative to the patch-center $(s/2, s/2)$ at 
a location $(i,j) | i, j{\in}[1,s]$ in the patch is defined as $\boldsymbol{\Delta}_{(i,j)} {\in} \mathbb{N}^2$.  
In standard PE, the offsets of a single patch grid are fixed as:
\begin{equation}\label{eq:fixed_offset}
\begin{aligned}
\boldsymbol{\Delta}^{fixed}_{(i, j)}{\in}[\lfloor-\frac{s}{2}\rfloor, \lfloor+\frac{s}{2}\rfloor]^2.
\end{aligned}
\end{equation}
Take \eg a $3{\times}3$ patch, offsets $\boldsymbol{\Delta}^{fixed}_{(i,j)}$ relative to the patch center $(1, 1)$ will lie in $[-1, 1] {\times} [-1, 1]$. They are fixed as:
\begin{equation}\label{eq:fixed_offset_sample}
\begin{aligned}
\boldsymbol{\Delta}^{fixed}=\{(-1, -1),(-1, 0), ...,(1, 0),(1, 1)\}.
\end{aligned}
\end{equation}

However, the aforementioned equirectangular projection process leads to severe shape distortions in the projected panoramic image, as seen in Fig.~\ref{fig:da_settings}.
A standard PE module with fixed patchifying positions makes the Transformer model neglect these shape distortions of objects and the panoramas. 
Inspired by deformable convolution~\cite{dai2017deformable} and overlapping PE~\cite{segformer}, we propose \emph{Deformable Patch Embeddings (DPE)} to perform the patchifying process respectively for the input image in the encoder and the feature maps in the decoder. The DPE module enables the model to learn a data-dependent and distortion-aware offset $\boldsymbol{\Delta}^{DPE}{\in} \mathbb{N}^{H{\times}W{\times}2}$, thus, the spatial connections of objects presenting in distorted patches can be featured by the model.
{DPE for patchifying is learnable and can predict adaptive offsets according to the given input $\boldsymbol{f}$.
Compared to fixed offsets in Eq.~\eqref{eq:fixed_offset}, the adaptive offsets $\boldsymbol{\Delta}^{DPE}_{(i,j)}$ are predicted as in Eq.~\eqref{eq:offset}.} 
\begin{equation}\label{eq:offset}
\begin{aligned}
    \boldsymbol{\Delta}^{DPE}_{(i,j)} &= \begin{bmatrix}
           \min(\max(-\frac{H}{r}, g(\boldsymbol{f})_{(i,j)}), \frac{H}{r}) \\
           \min(\max(-\frac{W}{r}, g(\boldsymbol{f})_{(i,j)}), \frac{W}{r})
         \end{bmatrix},
\end{aligned}
\end{equation}
where $g(\cdot)$ is the offset prediction function, which we implement via the deformable convolution operation~\cite{dai2017deformable}. The hyperparameter $r$ in Eq.~\eqref{eq:offset} puts a constraint onto the leaned offsets and is better set as $4$ based on our experiments. 
The learned offsets make DPE adaptive and as a result distortion-aware.

\subsection{Deformable MLP}
\label{sec:dmlp}

Token mixers play a major role in the competitive modeling ability of attention-based Transformer models. The recent MLP-based models~\cite{mlp_mixer,cyclemlp} heuristically relax attention-based feature constraints by spatially mixing tokens via MLP projections. Inspired by the success of MLP-based mixers, we design a Deformable MLP~(DMLP) token mixer to conduct the adaptive feature parsing for panoramic semantic segmentation. Vanilla-MLP~\cite{mlp_mixer} based modules lack adaptivity which weakens the token mixing of panoramic data.
{In contrast, linked with the aforementioned DPE module, our DMLP-based decoder performs adaptive token mixing during the overall feature parsing, being aware of the deformation properties in $360^\circ$ images.}

\begin{figure}[!t]
	\centering
    \includegraphics[width=1.0\columnwidth]{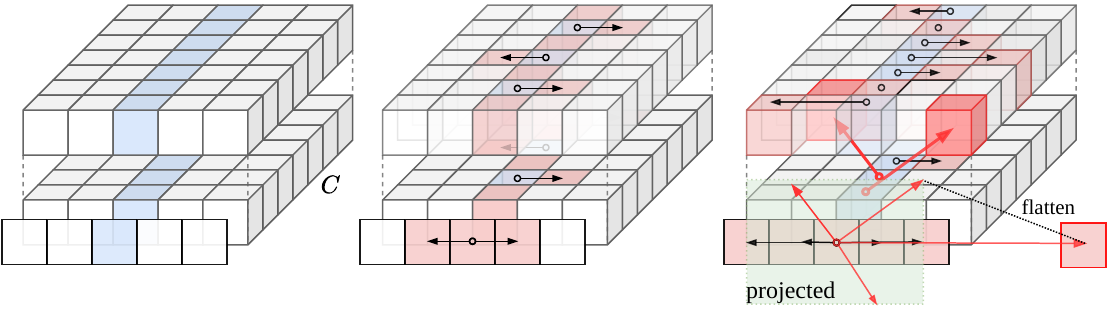}
    \begin{minipage}[t]{.2\columnwidth}
    \centering
    \vskip -4ex
    \subcaption{MLP}\label{fig:MLP}
    \end{minipage}%
    \begin{minipage}[t]{.45\columnwidth}
    \centering
    \vskip -4ex
    \subcaption{CycleMLP}\label{fig:CycleMLP}
    \end{minipage}%
    \begin{minipage}[t]{.35\columnwidth}
    \centering
    \vskip -4ex
    \subcaption{DMLP}\label{fig:OurMLP}
    \end{minipage}%
    \vskip -2ex
	\caption{\small {\textbf{Comparison of MLP modules.} The spatial-wise offsets of DMLP are learned adaptively according to the input feature map.}} 
	\label{fig:MLPBlocks}
\end{figure}
\begin{figure}[!t]
	\centering
    \includegraphics[width=1.0\columnwidth]{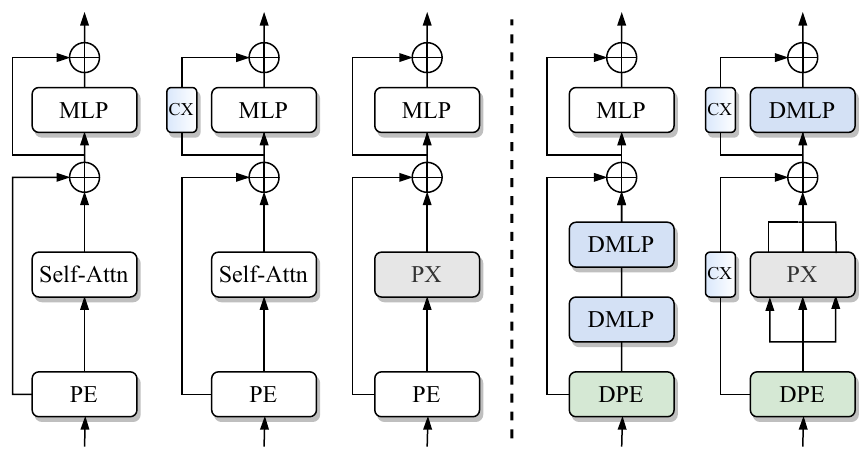}
    \begin{minipage}[t]{.20\columnwidth}
    \centering
    \vskip -4ex
    \subcaption{\scriptsize Transformer}\label{fig:mlp_block_trans}
    \end{minipage}%
    \begin{minipage}[t]{.17\columnwidth}
    \centering
    \vskip -4ex
    \subcaption{\scriptsize FAN}\label{fig:mlp_block_fan}
    \end{minipage}%
    \begin{minipage}[t]{.20\columnwidth}
    \centering
    \vskip -4ex
    \subcaption{\scriptsize PoolFormer}\label{fig:mlp_block_pool}
    \end{minipage}%
    \begin{minipage}[t]{.20\columnwidth}
    \centering
    \vskip -4ex
    \subcaption{\scriptsize Trans4PASS}\label{fig:mlp_block_dmlpv1}
    \end{minipage}%
    \begin{minipage}[t]{.23\columnwidth}
    \centering
    \vskip -4ex
    \subcaption{\scriptsize Trans4PASS+}\label{fig:mlp_block_dmlpv2}
    \end{minipage}%
    \vskip -1ex
	\caption{\small {\textbf{Comparison of token mixing structures.} PE: Patch Embedding, DPE: Deformable PE, Self-Attn: Self-Attention, CX: Channel Mixer, PX: Pooling Mixer, and DMLP: Deformable MLP.}} 
	\label{fig:MLPstructures}
\end{figure}

\noindent\textbf{DMLPv1 token mixer.} Concerning the comparison between MLP-based modules depicted in Fig.~\ref{fig:MLPBlocks}, the vanilla MLP~(Fig.~\ref{fig:MLP}) lacks the spatial context modeling, CycleMLP~(Fig.~\ref{fig:CycleMLP}) has the narrow projected receptive field due to fixed offsets, and our DMLP module~(Fig.~\ref{fig:OurMLP}) generates learned adaptive spatial offsets during mixing tokens and leads to a wider projected grid (\ie, the green panel).
Specifically, given a DPE-processed $C_{in}$-dimensional feature map $\boldsymbol{f}{\in}\mathbb{R}^{H{\times}W{\times}C_{in}}$, the spatial offset $\boldsymbol{\Delta}^{DMLP}_{(i,j,c)}$ is first predicted channel-wise by using Eq.~\eqref{eq:offset}.
{Then, the offset is flattened as a sequence in the shape of $\boldsymbol{\Delta}^{DMLP}_{(k,c)}$, where $k{\in}{H{\times}W}$ and $c{\in}C_{in}$.}
While the given feature map is projected into a sequence $\boldsymbol{z}$ with the equal shape, the offsets are used to select tokens during mixing the flattened token/patch features $\boldsymbol{z}{\in}\mathbb{R}^{HW{\times}C_{in}}$.
The mixed token is calculated as:
\begin{equation}\label{eq:dmlp}
\begin{aligned}
    \hat{\boldsymbol{z}}_{(k,c)}=\sum_{k=1}^{HW}\sum_{c=1}^{C_{in}}w^{T}_{(k,c)}\cdot{\boldsymbol{z}_{(k+\boldsymbol{\Delta}^{DMLP}_{(k,c)},c)}},
\end{aligned}
\end{equation}
where $w{\in}\mathbb{R}^{C_{in}{\times}C_{out}}$ is the weight matrix of a fully-connected layer. 
{As shown in Fig~\ref{fig:mlp_block_dmlpv1}, the DMLPv1 token mixer has a residual structure, consisting of DPE, two DMLPs, and an MLP module.}
Formally, the entire four-stage decoder is constructed by DMLPv1 token mixers and is denoted as:
\begin{equation}\label{eq:dmlpv1}
\begin{aligned}
    \hat{\boldsymbol{z}_{l}}&\coloneqq\textbf{DPE}(C_l, C_{emb})(\boldsymbol{z}_l), \forall_l{\in}\{1,2,3,4\} \\
    \hat{\boldsymbol{z}_{l}}&\coloneqq\textbf{DMLP}(C_{emb}, C_{emb})(\hat{\boldsymbol{z}_{l}}) + \hat{\boldsymbol{z}_{l}}, \forall_l \\
    \hat{\boldsymbol{z}_{l}}&\coloneqq\textbf{MLP}(C_{emb}, C_{emb})(\hat{\boldsymbol{z}_{l}}) + \hat{\boldsymbol{z}_{l}}, \forall_l \\
    \hat{\boldsymbol{z}_{l}}&\coloneqq\textbf{Up}(H/4, W/4)(\hat{\boldsymbol{z}_{l}}), \forall_l \\
    p&\coloneqq\textbf{LN}(C_{emb}, C_K)(\sum_{l=1}\hat{\boldsymbol{z}_{l}}),
\end{aligned}
\end{equation}
{where \textbf{Up}($\cdot$) and \textbf{LN}($\cdot$) refer to operations of Upsample and LayerNorm, and $p$ is the prediction of $K$ classes.}

\noindent\textbf{DMLPv2 token mixer.} Achieving the distortion-aware property and maintaining manageable computational complexity, we put forward a simple yet effective DMLPv2 toking mixer structure, which is demonstrated in Fig.~\ref{fig:mlp_block_dmlpv2}. {Compared to recent token mixers, such as PoolFormer~\cite{yu2021metaformer}~(Fig.~\ref{fig:mlp_block_pool}) and FAN~\cite{zhou2022fan}~(Fig.~\ref{fig:mlp_block_fan}), the advanced DMLPv2 is upgraded to a novel parallel token mixing mechanism by using a Squeeze\&Excite (SE)~\cite{hu2018SE_net} based Channel Mixer~(CX) and a non-parametric multi-scale Pooling Mixer~(PX).
Such a parallel token mixing mechanism brings two vital perspectives in the advanced DMLPv2 module: (1) the CX considers space-consistent but channel-wise feature reweighting, enhancing the feature by spotlighting informative channels; (2) the multi-scale PX and DMLP focus on spatial-wise sampling via fixed or adaptive offsets, yielding mixed tokens highlighted in relevant positions.} Thus, it improves the flexibility in modeling discriminative information and thereby reinforces the generalization capacity against domain shifts.
Furthermore, thanks to the sufficient token mixing of PX, the new DMLPv2 structure reduces the model complexity by using a single MLP layer, thus making the model more lightweight. Based on Eq.~\eqref{eq:dmlpv1}, the DMLPv2-based decoder is upgraded to:
\begin{equation}\label{eq:dmlpv2}
\begin{aligned}
    \hat{\boldsymbol{z}_{l}}&\coloneqq\textbf{DPE}(C_l, C_{emb})(\boldsymbol{z}_l), \forall_l{\in}\{1,2,3,4\} \\
    \hat{\boldsymbol{z}_{l}}&\coloneqq\sum_{s{\in}\{3,5,11\}}\textbf{PX}_{s{\times}s}(C_{emb}, C_{emb})(\hat{\boldsymbol{z}_{l}}) + \textbf{CX}(\hat{\boldsymbol{z}_{l}}), \forall_l \\
    \hat{\boldsymbol{z}_{l}}&\coloneqq\textbf{DMLP}(C_{emb}, C_{emb})(\hat{\boldsymbol{z}_{l}}) + \textbf{CX}(\hat{\boldsymbol{z}_{l}}), \forall_l \\
    \hat{\boldsymbol{z}_{l}}&\coloneqq\textbf{Up}(H/4, W/4)(\hat{\boldsymbol{z}_{l}}), \forall_l \\
    p&\coloneqq\textbf{LN}(C_{emb}, C_K)(\sum_{l=1}\hat{\boldsymbol{z}_{l}}),
\end{aligned}
\end{equation}
{where \textbf{PX}$_{s{\times}s}$($\cdot$) represents the average pooling operator in size of ${s{\times}s}$ and \textbf{CX}($\cdot$) denotes the channel-wise attention operator. {Based on our experiments, we observed that setting the multi-scale pooling as \{3,5,11\} yields superior results. The DMLP-based decoder delivers a spatial- and channel-wise token mixing in an efficient manner, but with a larger receptive field, which improves the expressivity of features in the panoramic imagery. More analysis about multi-scale pooling operations and the ablation study of PX and CX in DMLPv2 are presented in Sec.~\ref{sec:trans4pass_structural_analysis}.}

\subsection{Mutual Prototypical Adaptation}
\label{sec:mpa}
{To unfold the potential of panoramic segmentation models, a large-scale labeled dataset is crucial for success.
{However, labeling panoramic images is extremely time-consuming and expensive~\cite{densepass}, due to the ultra-wide FoV and small elements of panoramas.}
In this work, we dive deep into Unsupervised Domain Adaptation~(UDA) and exploit the sub-optimal but label-rich resources for training panoramic models, \ie, exploring Pinhole-to-Panoramic~(\textsc{Pin2Pan}) adaptation and the Synthetic-to-Real (\textsc{Syn2Real}) adaptation for panoramic segmentation.} 

\noindent\textbf{Preliminaries on domain adaptation.} Given the source (\ie, the pinhole or the synthetic) dataset with a set of labeled images $\mathcal{D}^s{=}\{(x^s, y^s)|x^s {\in}\mathbb{R}^{H{\times}W{\times}3}, y^s{\in} \{0,1\}^{H{\times}W{\times}K}\}$ and the target (\ie, the panoramic) dataset $\mathcal{D}^t{=}\{(x^t)|x^t{\in} \mathbb{R}^{H{\times}W{\times}3}\}$ without annotations, the objective of DA is to adapt models from the source to the target domain with $K$ shared classes. The model is trained in the source domain $\mathcal{D}^s$ via the segmentation loss:
\begin{equation}\label{eq:seg}
\begin{aligned}
    \mathcal{L}_{SEG}^s = -\sum_{i,j,k=1}^{H,W, K}y^s_{(i,j,k)}\text{log}(p^s_{(i,j,k)}),
\end{aligned}
\end{equation}
where $p^s_{(i,j,k)}$ is the probability of the source pixel $x^s_{(i,j)}$ predicted as the $k$-th class.
To transfer models to the target data, the target pseudo label $\hat{y}^t_{(i,j,k)}$ of pixels $x^t_{(i,j)}$ in Eq.~\eqref{eq:pseudo} is calculated based on the most probable class given by the source pre-trained model. 
\begin{equation}\label{eq:pseudo}
\begin{aligned}
    \hat{y}^t_{(i,j,k)}=\mathbbm{1}_{k\doteq\text{arg}\max p^t_{(i,j,:)}}.
\end{aligned}
\end{equation}
The Self-Supervised Learning (SSL) in Eq.~\eqref{eq:ssl} is used to optimize the model based on the target pseudo labels $\hat{y}^t_{(i,j,k)}$. 
\begin{equation}\label{eq:ssl}
\begin{aligned}
    \mathcal{L}_{SSL}^t = -\sum_{i,j,k=1}^{H,W,K}\hat{y}^t_{(i,j,k)}\text{log}(p^t_{(i,j,k)}).
\end{aligned}
\end{equation}

\noindent\textbf{Proposed Mutual Prototypical Adaptation.} {As shown in Fig.~\ref{fig:mpa}, a novel \emph{Mutual Prototypical Adaptation (MPA)} method is proposed and applied to distill mutual knowledge via the dual-domain prototypes, \ie, source and target domain prototypes.} Using hard pseudo-labels in the output space results in a limited adaptation of SSL methods. {To reduce the negative effect of hard pseudo-labels, our prototype-based method has two benefits: (1) it \emph{softens} the hard pseudo-labels by using them in feature space instead of as targets; (2) it performs \emph{complementary} alignment of semantic similarities in feature space. Thus, it makes the SSL more robust by using prototypes.} Further, the non-trivial design of prototype construction includes: (1) Prototypes are generated by using the source ground truth labels and the target pseudo labels, making full usage of labeled and unlabeled data and maintaining similar properties between domains, such as appearance cues of \textsc{Pin2Pan} and shape priors of \textsc{Syn2Real}; (2) Prototypes are constructed by using multi-scale feature embeddings, becoming more robust and more expressive; (3) Prototypes are stored in memory and updated along with the model optimization process, keeping the mechanism adaptable between iterations. {Furthermore, to eliminate the effect of noisy and incomplete pseudo labels, we boost MPA by using the Segment Anything Model~\cite{kirillov2023SAM}. Specifically, as shown in Fig.~\ref{fig:mpa}, the pseudo labels generated by source-only models are incomplete and affect the domain adaptation.} SAM trained with SA-1B dataset~\cite{kirillov2023SAM} is adopted as a readily available mask generator, which can effectively reconstruct the missing parts of the pseudo-label, such as \emph{roads} in Fig.~\ref{fig:mpa}. Compared to using standalone SAM or combined with SSL-based methods, our MPA strategy with SAM can achieve better results on UDA due to the robust mutual prototypical design.

\begin{figure}[!t]
	\centering
    \includegraphics[width=1.0\columnwidth]{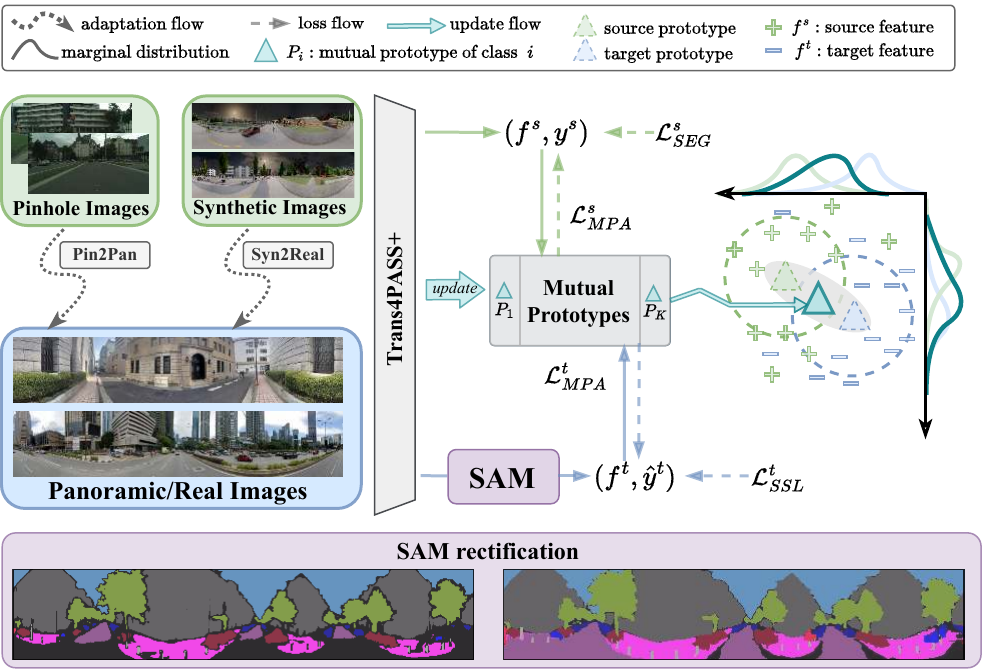}
	\caption{{\textbf{Diagram of Mutual Prototypical Adaptation (MPA).} The mutual prototypes are reconstructed by using dual-domain features of source and target domains. The target pseudo-labels are rectified by SAM~\cite{kirillov2023SAM} to eliminate the negative effect of missing parts, \eg, \emph{roads}. Zoom in for a better view.}}
    \label{fig:mpa}
\end{figure}

Specifically, a set of $n_s$ source- and $n_t$ target feature maps is constructed as $\boldsymbol{F} {=} \{\boldsymbol{f}^s_{1},\dots,\boldsymbol{f}^s_{n_s}\}{\bigcup}\{\boldsymbol{f}^t_{1},\dots,\boldsymbol{f}^t_{n_t}\}$, where $\boldsymbol{f}$ is fused from four-stage multi-scale features $\boldsymbol{f}{=}\sum_{l=1}^4f_l$ and is associated either with its respective source ground-truth label or a target pseudo-label. Each prototype $P_k$ is calculated by the mean of all feature vectors (pixel-embeddings) from $\boldsymbol{F}$ that share the class label $k$. 
{We initialize the mutual prototype memory $\mathbfcal{M}{=}\{P_1,...,P_K\}$ by computing the class-wise mean embeddings through the whole dataset. During the training process, the prototype $P_k$ is updated at timestep $t$ by $P^{t+1}_k{\leftarrow}m{P^{t-1}_k}{+}(1{-}m)P^{t}_k$ with momentum $m{=0.999}$, where $P^{t}_k$ is the mean feature vector among embeddings that share the class-label $k$ in the current mini-batch.}
Based on the dynamic memory, the prototypical feature map $\hat{\boldsymbol{f}}$ is reconstructed by stacking the prototypes $P_k {\in} \mathbfcal{M}$ according to the pixel-wise class distribution in either the source label or the pseudo-label. Inspired by the knowledge distillation loss~\cite{chen2020simclr_v2}, the MPA loss is applied to drive the feature alignment between the feature embedding $\boldsymbol{f}$ and the reconstructed feature map $\hat{\boldsymbol{f}}$.
The MPA loss only in the source domain is depicted in Eq.~\eqref{eq:loss_mpa}: 
\begin{equation}\label{eq:loss_mpa}
\begin{aligned}
    \mathcal{L}_{MPA}^s=&-\lambda{\mathcal{T}^2}\textbf{KL}(\phi(\hat{\boldsymbol{f}}^s/\mathcal{T})||\phi(\boldsymbol{f}^s/\mathcal{T}))\\
    &-(1-\lambda)\textbf{CE}(y^s,\phi(\boldsymbol{f}^s)), \\
\end{aligned}
\end{equation}
where $\textbf{KL}(\cdot)$, $\textbf{CE}(\cdot)$, and $\phi(\cdot)$ are Kullback–Leibler divergence, Cross-Entropy, and Softmax function, respectively. The temperature $\mathcal{T}$ and hyper-parameter $\lambda$ are $20$ and $0.9$ in our experiments. Similarly, the target MPA loss is constructed as in Eq.~\eqref{eq:loss_mpa_t}.
\begin{equation}\label{eq:loss_mpa_t}
\begin{aligned}
    \mathcal{L}_{MPA}^t=&-\lambda{\mathcal{T}^2}\textbf{KL}(\phi(\hat{\boldsymbol{f}}^t/\mathcal{T})||\phi(\boldsymbol{f}^t/\mathcal{T}))\\
    &-(1-\lambda)\textbf{CE}(\hat{y}^t,\phi(\boldsymbol{f}^t)), \\
\end{aligned}
\end{equation}
where the pseudo label $\hat{y}^t$ is generated by Eq.~\eqref{eq:pseudo}.

The final loss is combined by Eq.~\eqref{eq:seg}~\eqref{eq:ssl}~\eqref{eq:loss_mpa}~\eqref{eq:loss_mpa_t} with a weight of $\alpha{=}0.001$ as:
\begin{equation}\label{eq:loss_final}
\begin{aligned}
    \mathcal{L}{=}\mathcal{L}^s_{SEG}{+}\mathcal{L}^t_{SSL}{+}\alpha(\mathcal{L}^{s}_{MPA}{+}\mathcal{L}^{t}_{MPA}).
\end{aligned}
\end{equation}

\begin{figure*}[!t]
	\centering
    \includegraphics[width=1.0\textwidth]{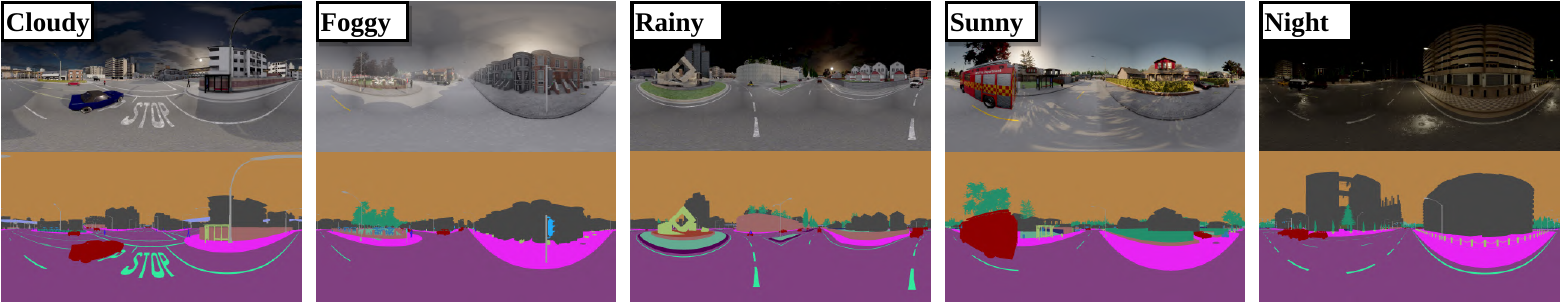}
    \vskip -2ex
	\caption{Examples of images and semantic labels in different conditions from the established SynPASS dataset.} 
    \label{fig:vis_synpass}
\end{figure*}

\begin{figure*}[!t]
	\centering
    \includegraphics[width=\textwidth]{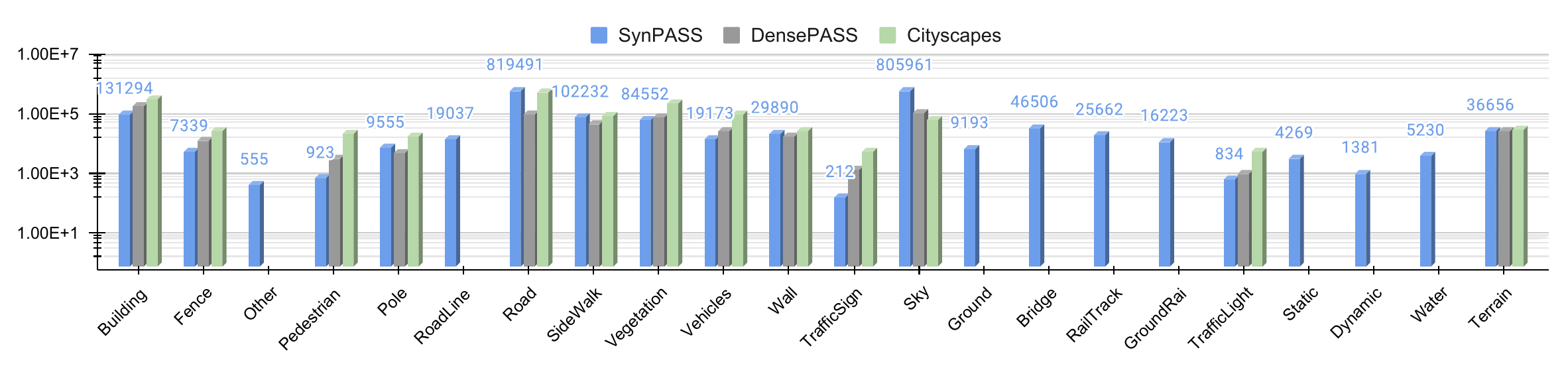}
	\vskip -2ex
	\caption{\small Distributions of SynPASS, DensePASS, and Cityscapes in terms of class-wise pixel counts per image. We use the logarithmic scaling of the vertical axis and insert the pixel count above the bar. There are $13$ classes overlapping across three datasets.
	}
    \label{fig:dataset_comp}
\end{figure*}

\section{SynPASS: Proposed Synthetic Dataset}
\label{sec:SynPASS}
Recently, the continuous emergence of panoramic semantic segmentation datasets~\cite{woodscape,densepass,stanford2d3d,structured3d,liao2022kitti360} has facilitated the development of surrounding perception, and simulators have been used to generate multi-modal data~\cite{testolina2022selma,sekkat2022synwoodscape}. However, there is currently not a readily available large-scale semantic segmentation dataset for outdoor synthetic panoramas, considering that the OmniScape dataset~\cite{omniscape} is still not released as of writing this paper. To explore the domain adaptation problem of \textsc{Syn2Real} under urban street scenes, we create the SynPASS dataset using the CARLA simulator~\cite{carla}. Our virtual sensor suite consists of $6$ pinhole cameras located at the same viewpoint to obtain a cubemap panorama image~\cite{spherephd}. The FoV of each pinhole camera was set to $91^\circ{\times}91^\circ$ to ensure the overlapping area between adjacent images. We then re-project the acquired cubemap panorama into a common equirectangular format using the cubemap-to-equirectangular projection algorithm. Given a equirectangular image grid $(\phi, \theta)$, we need to find the corresponding coordinates $(x, y)$ of each grid position on the cubemap $C{=}\{I_F, I_R, I_B, I_L, I_U, I_D\}$ to look up the value, where $\phi {\in} (-\pi, \pi)$, $\theta {\in} (-\frac{1}{2} \pi, \frac{1}{2} \pi)$, $\{I_F, I_R, I_B, I_L, I_U, I_D\} {\in} \mathbb{R} ^ {H{\times}W}$ are the \emph{front}, \emph{right}, \emph{back}, \emph{left}, \emph{top}, and \emph{bottom} view in the cubemap format, respectively. 
For $\{I_F, I_R, I_B, I_L\}$ indexed by $i{=}\{1,2,3,4\}$, we have:
\begin{equation}\label{eq:C2E_horizontal}
    \begin{aligned}
    \left\{
    \begin{array}{rcl}
    x & = & \frac{W}{2} \cdot tan(\phi - i\frac{\pi}{2}),\\
    y & = & -\frac{H \cdot tan\theta}{2cos(\phi - i\frac{\pi}{2})}.
    \end{array} \right. 
    \end{aligned}
\end{equation}
For $\{I_U, I_D\}$ indexed by $j{=}\{0,1\}$, we have:
\begin{equation}\label{eq:C2E_ud}
    \begin{aligned}
    \left\{
    \begin{array}{rcl}
    x & = & \frac{W}{2} \cdot cot\theta sin\phi,\\
    y & = & \frac{H}{2} \cdot cot\theta cos(\phi + j\pi).\\
    \end{array} \right. 
    \end{aligned}
\end{equation}
RGB images and semantic labels are captured simultaneously. In order to ensure the diversity of semantics, 
we benefit from $8$ open-source city maps and set $100{\sim}120$ initial collection points in every map. Our virtual collection vehicle drives according to the simulator traffic rules.
{We sample every $50$ frames and keep the first $10$ key-frames of images at each initial collection point.}
{To ensure the diversity of collected data, we modulate the weather and time conditions. Specifically, the weather conditions include \emph{sunny} ($25\%$), \emph{cloudy} ($25\%$), \emph{foggy} ($25\%$), and \emph{rainy} ($25\%$) cases. Besides, the time changes are \emph{daytime} ($85\%$) and \emph{nighttime} ($15\%$). Both illuminations are involved in all weather conditions, yielding more challenging cases such as \emph{rainy night}. The whole SynPASS dataset contains $9,080$ 
panoramic RGB images and semantic labels with a resolution of $1,024{\times}2,048$. Some examples are shown in Fig.~\ref{fig:vis_synpass}, and detailed statistic is in Table~\ref{table:synpass_stat}.} 

The distributions of SynPASS, the panoramic DensePASS~\cite{densepass}, and the pinhole Cityscapes~\cite{cityscapes} datasets are depicted in Fig.~\ref{fig:dataset_comp}. The class-wise pixel numbers are accumulated over all images in the respective datasets. Apart from the overlapping $13$ classes, the SynPASS dataset has $22$ classes in total, including $9$ additional classes: \texttt{other}, \texttt{roadline}, \texttt{ground}, \texttt{bridge}, \texttt{railtrack}, \texttt{groundrail}, \texttt{static}, \texttt{dynamic}, and \texttt{water}. It provides more semantic categories to enrich the 360$^\circ$ scene understanding. 

\input{tables/tab1_synpass_stat}

\section{Experiments}
\label{sec:experiments}

\subsection{Datasets and Settings}
We experiment with six datasets, including two source domains and one target domain in respective indoor and outdoor scenes:
\begin{compactitem}
\item[(1)] Indoor \emph{Panoramic} and \emph{Real} dataset as the \emph{target} domain: Stanford2D3D~\cite{stanford2d3d} Panoramic (\textbf{SPan}) has $1,413$ panoramas and $13$ classes.
{Results are averaged by the official three folds, following~\cite{stanford2d3d} unless otherwise stated.}
\item[(2)] Indoor \emph{Pinhole} and \emph{Real} dataset as the first \emph{source} domain: Stanford2D3D~\cite{stanford2d3d} Pinhole~(\textbf{SPin}) has $70,496$ pinhole images and the same $13$ classes as its panoramic dataset. 
\item[(3)] Indoor \emph{Panoramic} and \emph{Synthetic} dataset as the second \emph{source} domain: Structured3D~\cite{structured3d}~(\textbf{S3D}) has $21,835$ synthetic panoramic images and $29$ classes. 
\item[(4)] Outdoor \emph{Panoramic} and \emph{Real} dataset as the \emph{target} domain: DensePASS~\cite{densepass}~(\textbf{DP}) collected from cities around the world has $2,000$ images for transfer optimization and $100$ labeled images for testing, annotated with $19$ classes. 
{\item[(5)] Outdoor \emph{Pinhole} and \emph{Real} dataset as the first \emph{source} domain: Cityscapes~\cite{cityscapes}~(\textbf{CS}) has $2,979$ and $500$ images in the training and validation sets, and has the same $19$ classes as DensePASS. }
\item[(6)] Outdoor \emph{Panoramic} and \emph{Synthetic} dataset as the second \emph{source} domain: SynPASS~(\textbf{SP}) contains $9,080$ panoramic images and $22$ categories.
{More details are in Sec.~\ref{sec:SynPASS}.}
\end{compactitem}

\noindent\textbf{Four domain adaptation settings} are investigated:
\begin{compactitem}
\item[(1)] Indoor \textsc{Pin2Pan}: \textbf{SPin}~{\large \MVRightarrow}~\textbf{SPan}.
\item[(2)] Indoor \textsc{Syn2Real}: \textbf{S3D}~{\large \MVRightarrow}~\textbf{SPan}.
\item[(3)] Outdoor \textsc{Pin2Pan}: \textbf{CS}~{\large \MVRightarrow}~\textbf{DP}.
\item[(4)] Outdoor \textsc{Syn2Real}: \textbf{SP}~{\large \MVRightarrow}~\textbf{DP}.
\end{compactitem}

\noindent\textbf{Overlapping classes.}
To compare \textsc{Pin2Pan} and \textsc{Syn2Real} adaptations, only the overlapping classes are involved. The indoor datasets have $8$ classes, and the outdoor datasets have $13$ classes. Thus, the adaptation settings are reformed as:
\begin{compactitem}
\item[(1)] Indoor \textsc{Pin2Pan}: \textbf{SPin8}~{\large \MVRightarrow}~\textbf{SPan8}.
\item[(2)] Indoor \textsc{Syn2Real}: \textbf{S3D8}~{\large \MVRightarrow}~\textbf{SPan8}.
\item[(3)] Outdoor \textsc{Pin2Pan}: \textbf{CS13}~{\large \MVRightarrow}~\textbf{DP13}.
\item[(4)] Outdoor \textsc{Syn2Real}: \textbf{SP13}~{\large \MVRightarrow}~\textbf{DP13}.
\end{compactitem}

\noindent\textbf{Implementation settings.}
We train our models with $4$ A100 GPUs with an initial learning rate of $5e^{-5}$, which is scheduled by the poly strategy with power $0.9$ over $200$ epochs. The optimizer is AdamW~\cite{adam} with epsilon $1e^{-8}$, weight decay $1e^{-4}$, and batch size is $4$ on each GPU. The images are augmented by the random resize with ratio $0.5${--}$2.0$, random horizontal flipping, and random cropping to $512{\times}512$.
{For outdoor datasets, the resolution is $1,080{\times}1,080$ and the batch size is $1$.}
When adapting the models from \textsc{Pin2Pan}, the resolution of indoor pinhole and panoramic images are $1,080{\times}1,080$ and $1,024{\times}512$ for training. {In \textsc{Syn2Real}, the resolution of synthetic panoramic images is $1,024{\times}512$.}
The outdoor pinhole- and synthetic images are set to $1,024{\times}512$ and the panoramic images are with a resolution of $2,048{\times}400$. The image sizes of indoor and outdoor validation sets are $2,048{\times}1024$ and $2,048{\times}400$, respectively. Adaptation models are trained within $10K$ iterations on one GPU.

\input{tables/tab2_sota_synpass}

\subsection{\textbf{SynPASS Benchmark}}
In order to study the performance of panoramic semantic segmentation of current existing approaches and our approach Trans4PASS+ on the proposed synthetic dataset, the SynPASS benchmark with the full $22$ classes is established. As shown in Table~\ref{table:synpass_benchmark}, we conduct experiments for panoramic semantic segmentation on the SynPASS dataset using either CNN-based approaches (\eg, Fast-SCNN~\cite{fastscnn}, DeepLabv3+~\cite{deeplabv3+}, and HRNet~\cite{hrnet}) or transformer-based approaches (\eg, PVT~\cite{pvt}, SegFormer~\cite{segformer}, and the proposed Trans4PASS and Trans4PASS+). All the investigations among transformer-based approaches are conducted considering the trade-off between efficiency and model size within a fair comparison. The models are trained on the overall training set and their performances are reported in different weather and day/night conditions. {Compared with existing approaches, Trans4PASS+ (Tiny) surpasses HRNet with the best performance among all the listed CNN-based methods by ${+}5.37\%$ in mIoU on the validation set.}
{Trans4PASS and Trans4PASS+ consistently outperform PVT and SegFormer in all conditions. Compared to SegFormer-B2, our small model achieves respective ${+}4.12\%$ and ${+}3.48\%$ gains on the validation- and test set. The largest improvement lies in the \emph{rainy} condition with a ${+}5.03\%$ gain. The results showcase that our models have a strong capability to capture panoramic segmentation cues on the synthetic dataset even considering different weather and day/night scenarios.}

{Compared to Trans4PASS, the advanced Trans4PASS+ model performs more accurately in all conditions and clearly elevates the overall mIoU scores on both validation- and testing sets.}
According to Table~\ref{table:synpass_stat}, the samples are equally distributed among different scenarios and Trans4PASS+ also yields balanced segmentation performance across different kinds of weather and illuminating conditions, which demonstrates the robustness of our new model in different scenarios. {The results of all the investigated models illustrate that there is still remarkable improvement space on the newly established benchmark, since the best performance is $40.72\%$ on the SynPASS test set, indicating that the new benchmark is challenging due to its high diversity.}

\input{tables/tab3_sota_CS_DP}

\subsection{\textbf{\textsc{Pin2Pan}} and \textbf{\textsc{Syn2Real}} Gaps}
\label{sec:pin2pan_and_syn2real_gaps}
\noindent\textbf{\textsc{Pin2Pan} gaps.}
We first quantify the \textsc{Pin2Pan} domain gap in outdoor scenarios by assessing ${>}15$ off-the-shelf convolutional- and transformer-based semantic segmentation models learned from Cityscapes.\footnote[1]{MMSegmentation: https://github.com/open-mmlab/mmsegmentation.} Table~\ref{table:outdoor_domain_gap} presents the results evaluated on Cityscapes~\cite{cityscapes} and DensePASS~\cite{densepass} validation sets.
CNN-based methods such as PSPNet~\cite{pspnet} and DANet~\cite{danet} experience significant performance drops (${\sim}50\%$) when transferred to panoramic data.
Transformer-based models like SETR~\cite{setr} and SegFormer~\cite{segformer} still exhibit a large mIoU gap of ${\sim}40\%$. MaskFormer~\cite{cheng2021maskformer} and Mask2Former~\cite{cheng2022mask2former} with SegFormer-B2 achieve smaller mIoU gaps, yielding respective mIoU scores of $46.1\%$ and $46.4\%$ at higher complexities ($52.8$ and $68.7$ GFLOPs) on the panoramic domain. Mask2Former with Swin has larger FLOPs and number of parameters, \eg, $97.1$G FLOPs and $68.8$M parameter with Swin-S, obtaining a good result in the source domain. However, using a larger backbone cannot guarantee a better result in pinhole-to-panoramic adaptive segmentation.
In contrast, our Trans4PASS+ (S) model obtains $46.5\%$ in mIoU with only $19.8$ GFLOPs. 
Achieving higher performance in the target domain remains another goal in addition to obtaining smaller performance gaps across domains. 
While utilizing the same Mas2Former head (\eg, Swin-S\MVRightarrow SegFormer-B2\MVRightarrow Trans4PASS+), Trans4PASS+$^*$ obtains a better mIoU score on the DensePASS dataset ($45.8\%$\MVRightarrow$46.4\%$\MVRightarrow$48.4\%$ in mIoU) with lower complexity  ($97.1$\MVRightarrow$68.7$\MVRightarrow$63.2$ GFLOPs), thanks to the distortion-aware DMLPv2 module. 
The overall improvements show that by incorporating distortion awareness during training with pinhole data, our Trans4PASS+ models gain a stronger capacity to handle panorama deformation and object distortion.

\input{tables/tab4_sota_SPin_SPan}

Then, we look into the \textsc{Pin2Pan} domain gap in indoor scenes, as analyzed in Table~\ref{table:indoor_domain_gap} based on the Stanford2D3D dataset~\cite{stanford2d3d}.
{The pinhole- and panoramic images from Stanford2D3D are collected under the same setting, and the \textsc{Pin2Pan} gap is smaller compared to the outdoor scenario.}
Still, in light of other convolutional- and attentional transformer-based architectures, the small Trans4PASS+ variant leads to top mIoU scores of $51.48\%$ and $49.76\%$ for pinhole- and panoramic image semantic segmentation, while its accuracy drop is also largely reduced compared to former state-of-the-art models like Trans4Trans~\cite{zhang2021trans4trans_acvr}.

\input{tables/tab5_s2r_p2p_gaps}

\noindent\textbf{\textsc{Syn2Real} gaps.}
{To measure the \textsc{Syn2Real} domain gap, for outdoor scenes, we leverage our SynPASS~(\textbf{SP13}) and DensePASS~(\textbf{DP13}) datasets with the overlapping $13$ classes.}
{Compared to the previous work~\cite{pvt}, Trans4PASS consistently improves the performance in Table.~\ref{table:comp_domain_gap}-\circled{2}, surpassing the corresponding PVT by ${>}5\%$ on the target domain, and resulting in more robust omni-segmentation as shown in Fig.~\ref{fig:fov}.}
Similarly, for the indoor situation, we experiment on Structured3D~(\textbf{S3D8}) panoramic~\cite{structured3d} and Stanford2D3D panoramic~(\textbf{SPan8}) sets by using their sharing $8$ categories. The results are presented in Table.~\ref{table:comp_domain_gap}-\circled{4}. 
{The advance of Trans4PASS+ with DMLPv2 is pronounced, as it improves ${+}8.25\%$ mIoU compared to the tiny PVT baseline.}
{More in-depth architectural analysis of Trans4PASS+ is in Sec.~\ref{sec:trans4pass_structural_analysis}.}

\noindent\textbf{\textsc{Pin2Pan} \vs \textsc{Syn2Real}.}
{In Table~\ref{table:comp_domain_gap}, we study \textsc{Pin2Pan} and \textsc{Syn2Real} paradigms, to inspect the domain shift and answer the question: \emph{which adaptation scheme is more promising for panoramic semantic segmentation}. Here, a short answer is provided. For the outdoor scenario, the model benefits more from real pinholes than from synthetic panoramas without any adaptation. The \textsc{Pin2Pan}-learned small Trans4PASS+ (\circled{1}) reaches $51.48\%$ in mIoU, while the \textsc{Syn2Real}-transferred variant (\circled{2}) only achieves $43.83\%$.}
{We conjecture that without any domain adaptation, the rich detailed texture cues available in the real pinhole outdoor dataset play an important role in attaining generalizable segmentation.}
For the indoor scenario, as shown in Table~\ref{table:comp_domain_gap}-\circled{3}, the \textsc{Pin2Pan} model also achieves higher performance. %
In contrast, transferring from the synthetic \textbf{S3D8} dataset to real \textbf{SPan8} (Table~\ref{table:comp_domain_gap}-\circled{4}) causes a mIoU gap of ${>}20\%$. Yet, we find that Trans4PASS+, with parallel token mixing, attains smaller \textsc{Syn2Real} mIoU gaps than Trans4PASS. 
More analyses are in Sec.~\ref{sec:pin2pan_and_syn2real_adaptation} and Sec.~\ref{sec:qualitative_analysis}.

\input{tables/tab6_s2r_transfer}

{Moreover, as shown in Table~\ref{table:comp_in_syn2real}, we note that our proposed Trans4PASS+ shows strong zero-shot generalization capacity when only learning from synthetic images and testing on real panoramic data.
It achieves $52.09\%$ in mIoU, which surprisingly outperforms %
the previous state-of-the-art HRNet~\cite{hrnet} ($52.00\%$) trained with both synthetic- and real (extra $1,063$ annotations) datasets as suggested in~\cite{structured3d}.}

\input{tables/tab7_str_abl_1}
\input{tables/tab8_str_abl_2}
\input{tables/tab9_comp_PE_MLP}
\input{tables/tab10_token_mixing}

\subsection{Study of Trans4PASS+ Structure}
\label{sec:trans4pass_structural_analysis}
\noindent\textbf{Ablation of different components.} 
{
We first study the proposed DPE, DMLPv1, and DMLPv2 in Table~\ref{tab:str_abl_1}. 
In the first group without using DPE, both of our DMLP methods achieve better performance with lower computational complexity, which have respective $45.14\%$ and $46.94\%$ in mIoU on the DensePASS dataset. Besides, the new DMLPv2 method can bring additional improvement (${+}1.8\%$ mIoU) as compared to DMLPv1.
In the second group with DPE, consistent performance improvements of both DMLP methods are observed. The DMLPv2 has mIoU of $49.94\%$ with ${+}4.05\%$ gains compared to DMLPv1, which indicates the advantage of DMLPv2 using sufficient token mixing.
However, in cross-group comparisons, both DMLP methods with DPE can obtain substantial performance gains than their non-DPE implementations. Specifically, thanks to the effective DPE, the DMLPv2 method improves the mIoU score from $46.94\%$ to $49.94\%$. Through the ablation study, it is proved that the proposed components are effective, yielding an enhanced Trans4PASS+ model for handling omnidirectional scene segmentation.
}

\begin{figure}[!t]
	\centering
    \includegraphics[width=0.9\columnwidth]{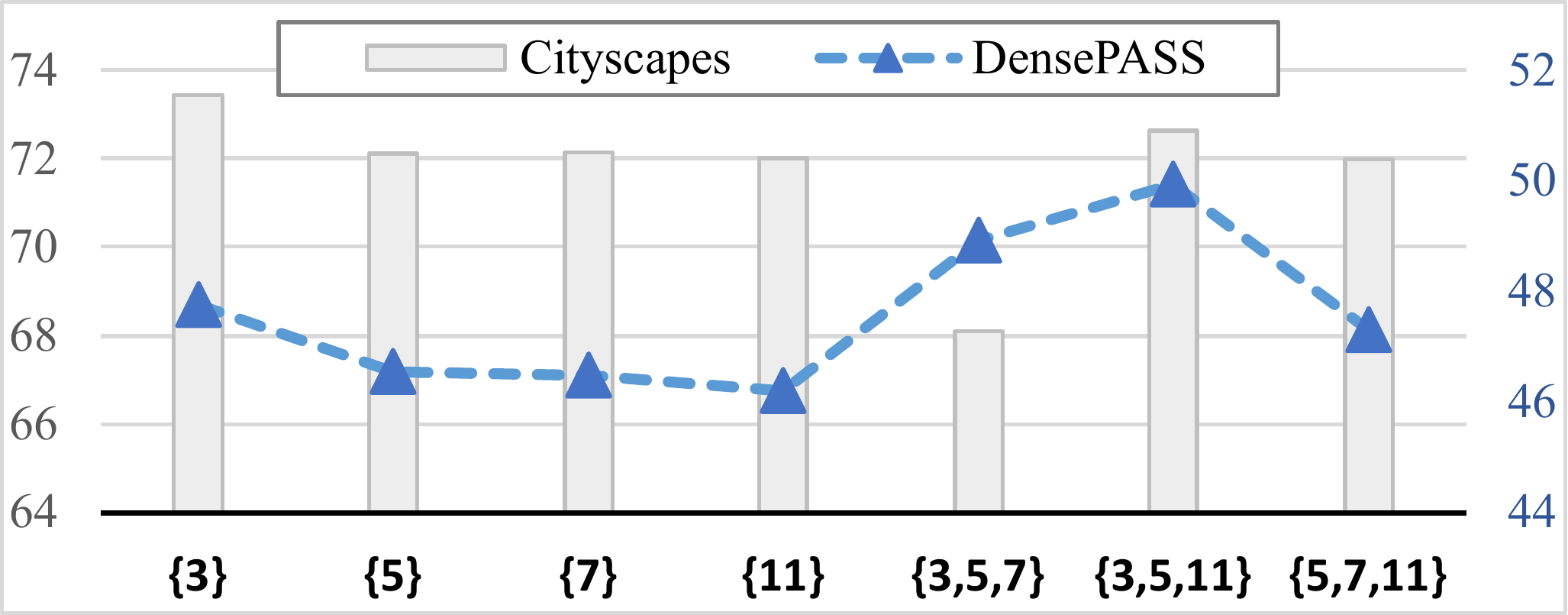}
    \vskip -2ex
	\caption{{Analysis of single-/multi-scale pooling operations and combinations in DMLPv2 module. \{$s$\} means a $s{\times}s$ pooling.}}
    \label{fig:pooling_ablation}
\end{figure}

\noindent\textbf{Ablation of PX and CX.} 
{
In Table~\ref{tab:str_abl_2}, we further study the effectiveness of two token mixers, \ie, the pooling mixer (PX) and the channel mixer (CX) in our advanced DMLPv2. In the case of applying PX and CX separately, there are ${+}7.64\%$ and ${+}8.64\%$ gains compared with the baseline.
However, coupled with both PX and CX, our Trans4PASS+ model with DMLPv2 strikingly boosts the mIoU on DensePASS to $49.94\%$ in mIoU, having ${+}10.92\%$ gains over the baseline. It shows that PX and CX both contribute significantly, forming an effective parallel token mixing to unleash the potential of Trans4PASS+ in handling panoramas.} %

\noindent\textbf{Analysis of multi-scale pooling.} 
{To verify the pooling operation selections, we conduct seven variants in the DMLPv2 module, which include four single pooling and three multi-scale pooling combinations. In Fig.~\ref{fig:pooling_ablation}, we found that the multi-scale pooling of \{3,5,11\} achieves a better performance on the DensePASS dataset.}

\noindent\textbf{Comparison of PE and MLP methods.} 
{
In Table~\ref{tab:comp_PE_MLP}, we perform a comprehensive comparison between various PE and MLP methods. {To ensure a fair comparison, all hyper-parameters like the kernel size ($k{=}7$) and the stride ($s{=}3$) in the deformable convolution are set to be the same as those in the Vanilla PE and our DPE.} Compared to the baseline with vanilla PE and vanilla MLP, our DPE with DMLPv1 improves the panoramic segmentation from $39.02\%$ to $45.89\%$ with ${+}6.87\%$ gains in mIoU on the DensePASS dataset. 
{Compared with vanilla PE, simply using deformable conv for patch embedding cannot ensure improvement due to the lack of pre-training and overlapping operations. In contrast, our DPE includes overlapping and deformable designs and can be trained from scratch, yielding better results in the panoramic domain.}
{To ablate the impacts of different MLP-like modules integrated into the decoder of Trans4PASS+, we compared the DMLPv2 module with vanilla MLP, CycleMLP~\cite{cyclemlp} and ASMLP~\cite{asmlp} modules.}
Our DMLPv2 method is more adaptive as opposed to the fixed offsets in CycleMLP, as depicted in Fig.~\ref{fig:MLPBlocks}. The results also confirm the benefit as DMLPv2 outstrips these modules with a clear margin of $6{\sim}9\%$ in mIoU.
}

\begin{table*}[!t]
	\caption{\textbf{Comparisons and ablation studies of \textsc{Pin2Pan} domain adaptation} in indoor and outdoor scenarios.}
	\renewcommand\arraystretch{1.0}
	\begin{center}
	\vskip-3ex
		\input{tables/table_mpa.tex}
	\end{center}
\vskip-1ex
	\label{tab:pin2pan_in_out}
	\vskip-6ex
\end{table*}

\noindent\textbf{Comparison of token mixing methods.} 
{
In Table~\ref{tab:token_mixing}, we compare our parallel token mixing method with existing methods, including the average-pooling-based mixer from PoolFormer~\cite{yu2021metaformer}, the channel mixer from FAN block~\cite{zhou2022fan}, and a combination of FAN and PoolFormer. They are combined in a similar parallel way but without adding any deformable designs. However, compared to the PoolFormer and the FAN methods, our DMLPv2 method obtains respective ${+}6.76\%$ and ${+}7.40\%$ gains in mIoU on the DensePASS dataset. Besides, our parallel token mixing design further outperforms the combination of PoolFormer and FAN. 
The comparison illustrates that DMLPv2 offers a sweet spot and an optimal path to follow for attaining robust and effective panoramic segmentation against domain shift problems.
}

\subsection{Study of MPA Strategy}
\label{sec:ablation_unsupervised_domain_adaptation}

\subsubsection{Outdoor scenario: Cityscapes$\rightarrow$DensePASS}
\noindent\textbf{Comparison with outdoor state-of-the-art methods.}
Table~\ref{tab:more} shows the adaptation from Cityscapes to DensePASS. {We compare Trans4PASS+ with segmentation frameworks tailored for panoramas, \eg, PASS~\cite{pass} and ECANet~\cite{omnirange}.} 
Both of them are sub-optimal for robust omnidirectional surrounding parsing on the dense $19$-class segmentation benchmark of DensePASS~\cite{densepass}.

Then, we compare MPA-Trans4PASS+ against representative UDA pipelines including some built on adversarial learning such as CLAN~\cite{clan} and P2PDA~\cite{p2pda_trans}, and self-training schemes like CRST~\cite{crst}, SIM~\cite{wang2020differential}, PCS~\cite{yue2021pcs}, HRDA~\cite{hoyer2022hrda}, MIC~\cite{hoyer2023mic}, and DAFormer~\cite{hoyer2021daformer}. {Among these methods, P2PDA is the previous best solution for domain adaptive panoramic segmentation on DensePASS, whereas DAFormer serves as a recent transformer-based domain adaptation method.} {Yet, MPA-Trans4PASS arrives at $56.38\%$. Thanks to DMLPv2 and the SAM-based MPA method, Trans4PASS+ scores the highest $59.43\%$ in mIoU, which outstrips P2PDA-SSL by a large margin of ${+}17.44\%$. Meanwhile, it exceeds the prototypical approach PCS and the transformer-driven DAFormer with ${+}5.60\%$ and ${+}4.87\%$, respectively.}

We further compare multi-supervision methods~\cite{usss,seamless,issafe} which require much more data. USSS~\cite{usss} relies on multi-source SSL, while Seamless-Scene-Segmentation~\cite{seamless} uses instance-specific labels. {ISSAFE~\cite{issafe} merges training data from Cityscapes, KITTI-360~\cite{liao2022kitti360}, and BDD~\cite{bdd}. Their outputs are projected to the $19$ classes of 
DensePASS.} However, these multi-supervision approaches are less effective for panoramic semantic segmentation. {As seen in Table~\ref{tab:more}, our Trans4PASS+ models harvest top segmentation IoU scores on $11$ out of all $19$ categories.} %

\begin{figure}[!t]
	\centering
    \includegraphics[width=1.0\columnwidth]{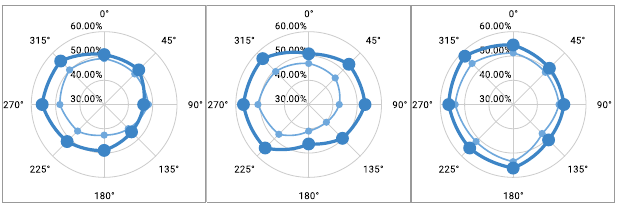}
    \begin{minipage}[t]{.32\columnwidth}
    \centering
    \vskip -4ex
    \subcaption{PVTv2}\label{fig:polar_densepass_pvtv2}
    \end{minipage}%
    \begin{minipage}[t]{.32\columnwidth}
    \centering
    \vskip -4ex
    \subcaption{Trans4PASS}\label{fig:polar_densepass_t4p}
    \end{minipage}%
    \begin{minipage}[t]{.32\columnwidth}
    \centering
    \vskip -4ex
    \subcaption{Trans4PASS+}\label{fig:polar_densepass_t4p+}
    \end{minipage}%
    \vskip -2ex
	\caption{\textbf{Omnidirectional segmentation} before (light blue lines) and after (blue lines) mutual prototypical adaptation. The mIoU (\%) scores in eight directions are reported.} 
	\label{fig:polar_densepass}
\vskip -4ex
\end{figure}

\noindent\textbf{Ablation on DensePASS.}
To test the generalizability, we replace FANet~\cite{fanet} and DANet~\cite{danet} used in P2PDA~\cite{densepass} with Trans4PASS, as displayed in Table~\ref{tab:per_class_densepass}. Trans4PASS comes with ${>}10\%$ gains due to the collected long-range dependencies and distortion-aware features. {In the second and third ablation groups of Table~\ref{tab:per_class_densepass}, our tiny and small Trans4PASS+ models with MPA and SAM can achieve respective $57.67\%$ and $59.43\%$ in mIoU. %
These results certify that MPA works collaboratively with pseudo labels rectified by SAM and offers a complementary feature alignment incentive.}

\noindent\textbf{Omnidiretional segmentation.}
To showcase the effectiveness of MPA on omnidirectional segmentation, the panorama is divided into $8$ directions and mIoU scores are calculated in each direction separately. {The polar diagram in Fig.~\ref{fig:polar_densepass} demonstrates that MPA consistently and reliably improves the adaptation performance while using PVTv2~\cite{pvtv2}, TransPASS, or Trans4PASS+.} %

\subsubsection{Indoor scenario: SPin$\rightarrow$SPan}
\noindent\textbf{Comparison with indoor state-of-the-art methods.}
Table~\ref{tab:supervised_s2d3d_pan} shows adaptation from Stanford2D3D~\cite{stanford2d3d} pinhole to panoramic domains (\textbf{SPin}$\rightarrow$\textbf{SPan}). {Surprisingly, Trans4PASS+ (${\sim}14$M parameters) outperforms existing fully-supervised and transfer-learning methods that use ResNet-101 (${\sim}44$M parameters).} For example, the versatile HoHoNet~\cite{hohonet} obtains $52.0\%$, whereas some methods~\cite{unet,gauge_equivariant,spherical_unstructured_grids,orientation} use RGB-D input to exploit cross-modal complementary information. Still, our lighter TransPASS+ achieves $52.3\%$ while being unsupervised. However, our supervised counterpart can reach $54.1\%$. These results further verify the distortion adaptability of the proposed Trans4PASS+ architecture for panoramic semantic understanding.

\begin{table*}[!t]
\footnotesize
\centering
\caption{\small {\textbf{\textsc{Pin2Pan} and \textsc{Syn2Real} domain adaptation results and comparisons} in both indoor and outdoor scenarios.}} %
\vskip -2ex

\input{tables/table_mpa_s2r_p2p.tex}
\vskip -3ex
\label{table:da_s2r_p2p}
\vskip-2ex
\end{table*}

\noindent{\textbf{Comparison with spherical models.} In Table~\ref{tab:supervised_s2d3d_pan}, we compare Trans4PASS+ with models proposed to address the deformation of spherical data. For example, distortion-aware Tangent~\cite{tangent} and HoHoNet~\cite{hohonet} models tailored for 360-degree obtain respectively $45.6\%$ and $52.0\%$. Thanks to our deformable modules, Trans4PASS+ can better process panoramas and has better scores.} 

\noindent\textbf{Ablation on Stanford2D3D.}
Table~\ref{tab:per_class_s2d3d-pan} presents the ablation study conducted in the fold-$1$ data splitting~\cite{stanford2d3d}. Our MPA-Trans4PASS (Tiny) exceeds the previous state-of-the-art P2PDA-driven DANet and it is even better than the one adapted with a PVT-Small backbone. Overall, Trans4PASS+ (Small) achieves the highest mIoU score ($53.49\%$), even reaching the level of the fully-supervised Trans4PASS+ ($54.1\%$) which does have full access to panoramic image annotations of $1,400$ target samples.

\subsection{\textbf{\textsc{Pin2Pan}} and \textbf{\textsc{Syn2Real}} Adaptation}
\label{sec:pin2pan_and_syn2real_adaptation}
The results and comparisons between {\textsc{Pin2Pan}} and {\textsc{Syn2Real} adaptation paradigms are detailed in Table~\ref{table:da_s2r_p2p}. %

\noindent\textbf{Comparison in the outdoor scenario.}
In Sec.~\ref{sec:pin2pan_and_syn2real_gaps}, we have briefly assessed the comparison between \textsc{Pin2Pan} and \textsc{Syn2Real} performance. {In Table~\ref{tab:da_out_s2r_p2p}, we inspect this in greater detail by using the two $13$-class benchmarks. Before adaptation, \textsc{Pin2Pan} models generally perform better than their corresponding \textsc{Syn2Real} ones.}
{This is due to that the detailed texture information available in the pinhole datasets, provides important cues for segmentation.
Yet, when looking into per-class results, we find that \textsc{Syn2Real} often achieves higher performance on \emph{sidewalk} ($40.52\%$ vs. $42.36\%$). \emph{Sidewalks} can get stretched and appear at multiple positions across the $360^\circ$, which is uncommon in pinhole data, and thereby they are difficult for source-only \textsc{Pin2Pan} models.} In \textsc{Syn2Real}, the spatial distribution- and position priors available in the panoramic synthetic dataset, can help context-aware models to better detect \emph{sidewalks}. %
After adaptation, \textsc{Pin2Pan} model largely improves the accuracy of \emph{sidewalk}, from $40.52\%$ to $54.74\%$, and outperforms the \textsc{Syn2Real}-adapted one which has $51.39\%$. In mIoU, \textsc{Pin2Pan}-adapted model has a better result with $55.24\%$ than the \textsc{Syn2Real} one with $50.88\%$. {This result reveals that \textsc{Pin2Pan} setting benefits more from pinholes, while the \textsc{Syn2Real} one benefits more from the mutual adaptation.}  %

\noindent\textbf{Comparison in the indoor scenario.}
Table~\ref{tab:da_in_s2r_p2p} shows that \textsc{Pin2Pan} yields better performances in both adaptation-free and MPA settings. The \textsc{Syn2Real} indoor models come with unsatisfactory performance on the segmentation of \emph{sofa} due to their different appearances in synthetic and real scenes. It further affects the overall mIoU after adaptation, which shows the challenges of adapting panoramic segmentation models from the synthetic to the real indoor domain.  %
{Nonetheless, MPA improves the performance via \textsc{Pin2Pan} domain adaptation, yielding $67.16\%$ mIoU.} %

\noindent\textbf{Analysis of using SAM.}
{Fig.~\ref{fig:sam_ablation} demonstrates the analysis of using SAM, SSL, and our MPA for panoramic semantic segmentation. The comparison is conducted with two UDA settings of {SP13}$\rightarrow${DP13} and {CS13}$\rightarrow${DP13}. Solely using SAM as a mask correction method, there is a limited improvement over the source-only model, yielding a ${+}2.66\%$ and a ${+}1.29\%$ gain, respectively. We note that using SAM to enhance pseudo labels for the conventional SSL method cannot guarantee a further boost compared to using SAM solely. One reason is the negative effect of the hard pseudo labels. {However, thanks to the mutual prototypes, our MPA with SAM can bring significant improvements, yielding $50.88\%$ and $55.24\%$ in mIoU, with respective gains of ${+}4.39\%$ and ${+}2.47\%$ over standalone SAM. It proves the effectiveness and advantage of using SAM to eliminate the negative impact of pseudo labels and highlights the robustness of MPA.}
\begin{figure}[!h]
	\centering
    \vskip -2ex
    \includegraphics[width=1.0\columnwidth]{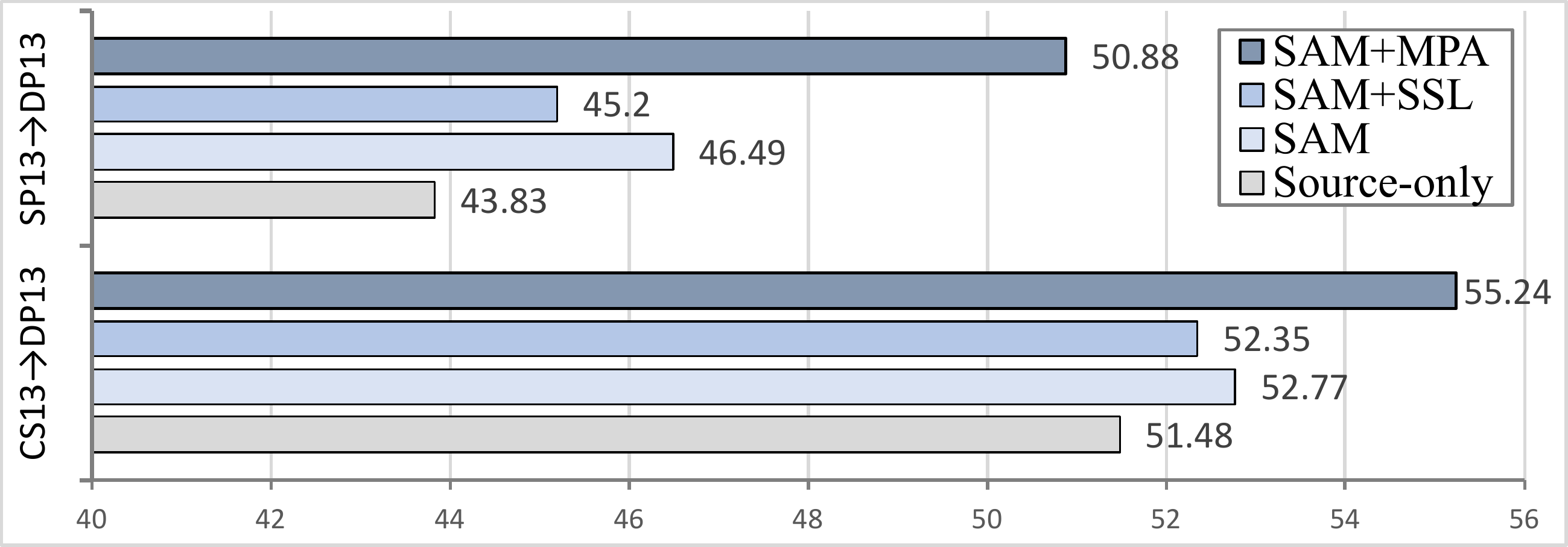}
    \vskip -2ex
	\caption{{Analysis of using SAM~\cite{kirillov2023SAM} for panoramic semantic segmentation, including source-only, SAM-enhanced, SAM+SSL, and our SAM+MPA methods.}}
    \label{fig:sam_ablation}
\end{figure}

\subsection{Qualitative Analysis}
\label{sec:qualitative_analysis}

\noindent\textbf{Panoramic semantic segmentation visualizations.}
In Fig.~\ref{fig:vis_outdoor} and Fig.~\ref{fig:vis_indoor}, Trans4PASS and Trans4PASS+ models can obtain better results than the indoor~\cite{pvt} and outdoor~\cite{segformer} baseline models. {In outdoor cases (Fig.~\ref{fig:vis_outdoor}), Trans4PASS+ obtains better results in \eg, \emph{trucks}, \emph{sidewalks}, and \emph{pedestrians}, while the baseline has difficulty distinguishing distorted objects.}
In indoor cases (Fig.~\ref{fig:vis_indoor}), the objects like \emph{doors} and \emph{tables} are difficult for the baseline, but our Trans4PASS+ predicts correctly.

\begin{figure*}[!t]
	\begin{subfigure}[b]{1.0\textwidth}   
		\centering 
		\includegraphics[width=1.0\textwidth]{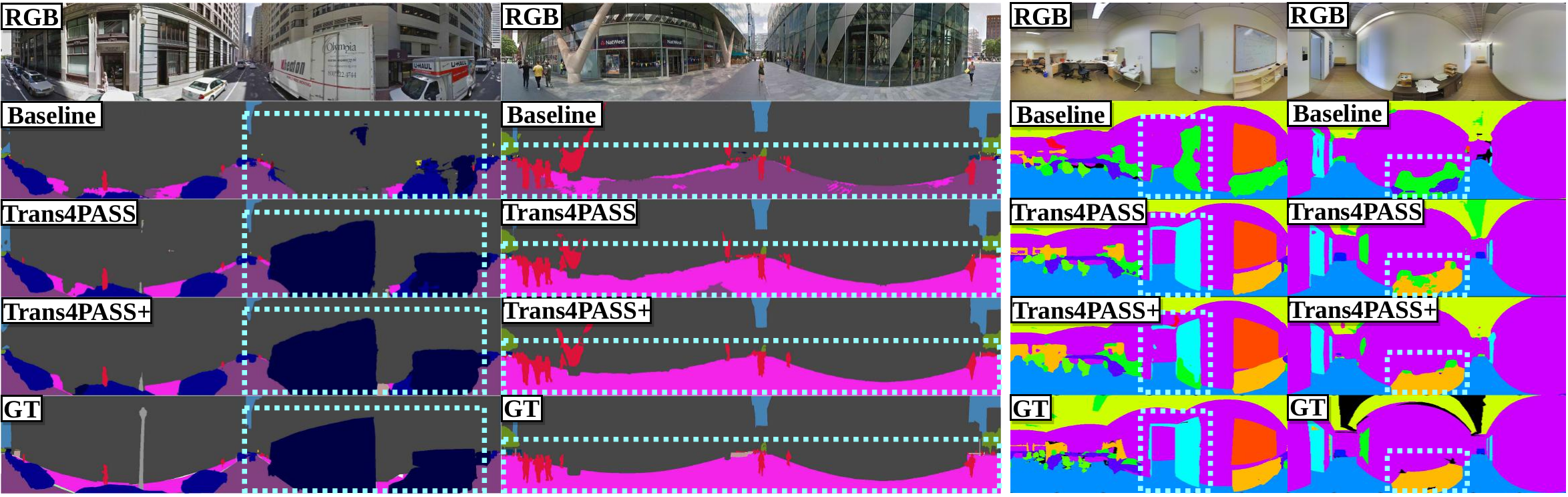}
	    \begin{minipage}[t]{.65\textwidth}
        \centering
        \vskip -4ex
        \subcaption{\small Segmentation outdoors}\label{fig:vis_outdoor}
        \end{minipage}%
        \begin{minipage}[t]{.35\textwidth}
        \centering
        \vskip -4ex
        \subcaption{\small Segmentation indoors}\label{fig:vis_indoor}
        \end{minipage}%
	\end{subfigure}
    \vskip -2ex
	\caption{\small \textbf{Panoramic semantic segmentation visualizations.} The baseline model~\cite{pvtv2} has no deformable designs. %
	{Zoom in for a better view.}} 
	\label{fig:vis}
\vskip -2ex
\end{figure*}

\begin{figure}[!t]
	\includegraphics[width=1.0\linewidth]{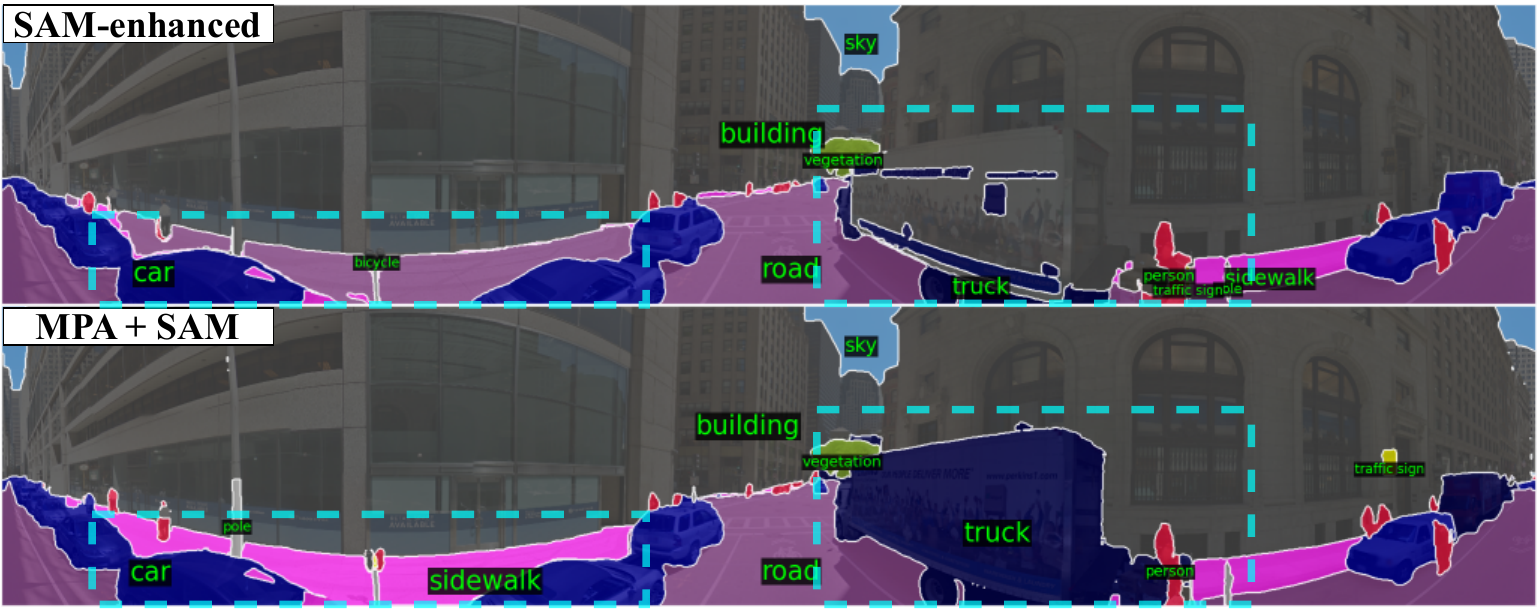}
    \vskip -1ex
	\caption{\small {Visualization of using SAM-enhanced and our SAM-based MPA adaptation methods. Zoom in for a better view.}
	} 
	\label{fig:vis_sam}
\vskip -2ex
\end{figure}

\noindent\textbf{SAM-based visualization.}
{Fig.~\ref{fig:vis_sam} shows a comparison between using SAM alone for correcting the prediction and combining it with our MPA method. The \emph{truck} class can be correctly recognized by the Trans4PASS+ model after being adapted by MPA with SAM, while it is missing in the SAM-enhanced prediction. A similar situation occurs in the \textit{sidewalk} class. The visualization results indicate the effectiveness of our proposed MPA method, which is enhanced by using SAM as the pseudo-label correction.}

\noindent\textbf{\textsc{Pin2Pan} vs. \textsc{Syn2Real}.}
{In Fig.~\ref{fig:vis_p2p_s2r}, we compare the two domain adaptation paradigms. Before MPA, the source-trained \textsc{Pin2Pan} model (\circledbrown{1}) fails to fully detect the \emph{sidewalk}, as the shapes and positional priors of sidewalks in pinhole imagery significantly differ from those in panoramas. In contrast, the source-trained \textsc{Syn2Real} model (\circledbrown{3}) handles the \emph{sidewalk} parsing well. This proves the observation in Fig.~\ref{fig:da_settings}, where the marginal distributions of the synthetic and real domains are close in one dimension encoding information like deformed shapes and positional priors. Yet, the \textsc{Syn2Real} model cannot identify the \emph{traffic signs}, which are recognized by \textsc{Pin2Pan} model that exploits the rich textures from pinhole real scenes, as the pinhole-source and panoramic-target domains are close in another dimension encoding appearance cues.} Yet, after MPA, the \textsc{Pin2Pan} model (\circledbrown{2}) can also seamlessly detect the \emph{sidewalk}, which indicates that our distortion-aware MPA-adapted Trans4PASS+ successfully fixes the large gap in shape-deformations and position-priors.
{However, the adapted \textsc{Syn2Real} model (\circledbrown{4}) has difficulty discovering the \emph{traffic signs}, which lack diverse textures in the simulated data. These segmentation maps corroborate the numerical results in Table~\ref{table:da_s2r_p2p}.}

\begin{figure}[!t]
	\includegraphics[width=1.0\columnwidth]{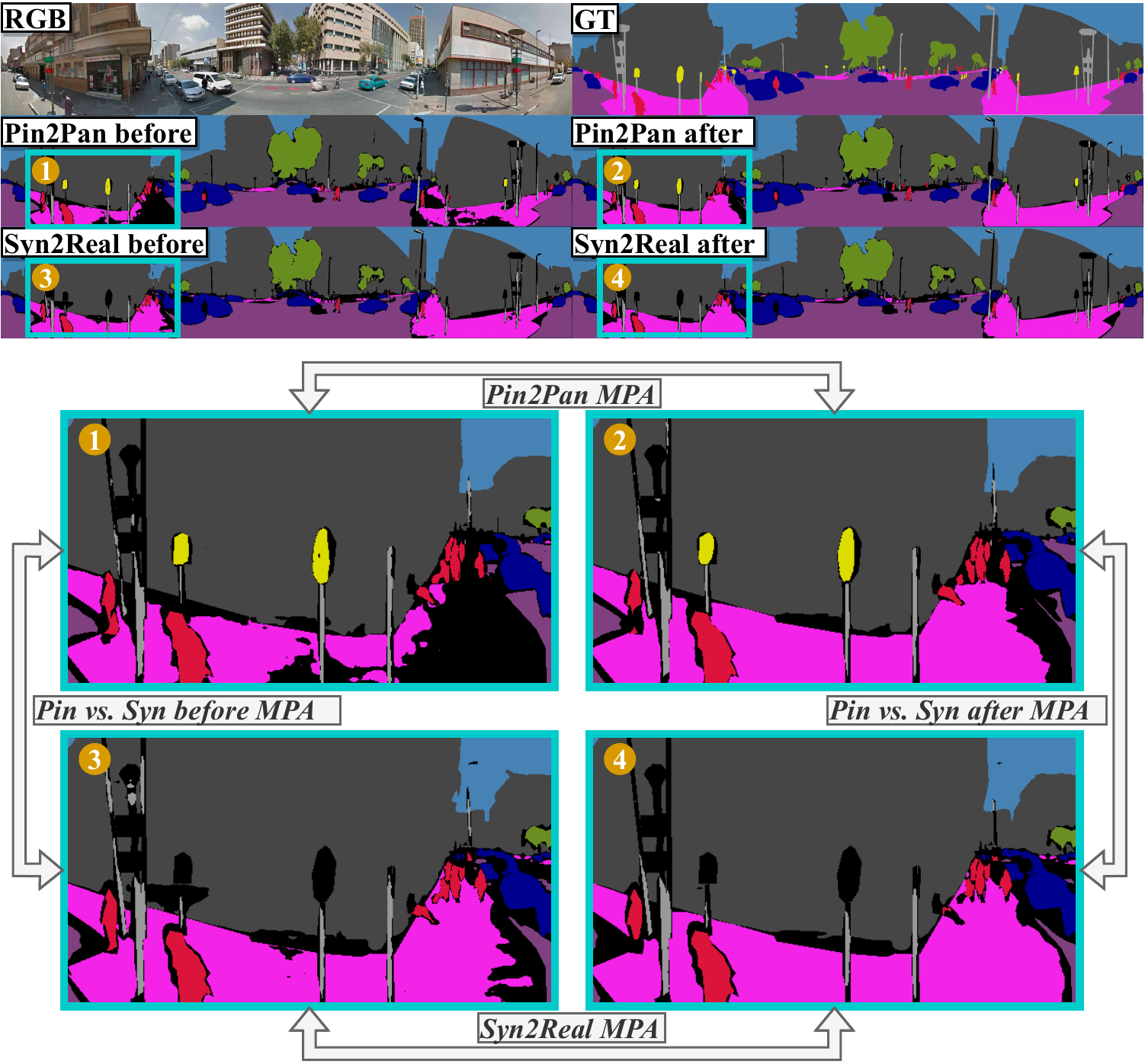}
    \vskip -1ex
	\caption{\small \textbf{\textsc{Pin2Pan} vs. \textsc{Syn2Real} visualizations} before and after MPA, respectively. The black areas indicate misprediction. %
	} 
	\label{fig:vis_p2p_s2r}
\vskip -2ex
\end{figure}

\noindent\textbf{Feature embedding comparison.}
To illustrate the effect of MPA on the feature space, the t-SNE visualization of feature embeddings before and after outdoor \textsc{Pin2Pan} DA is shown in Fig.~\ref{fig:supple_tsne_outdoor}. Each dot is the center of all pixels that share the same class in its image, and these images are from the training set of the respective domain.
{The blue triangle (\protect\marksymbol{triangle}{rblue}) in the source and target domains and the black triangle (\protect\marksymbol{triangle}{black}) in the mutual domain are the respective domain prototype of a certain class.}
Before domain adaptation, the feature embeddings of the source domain, the target domain, and their mutual domain are shown in Fig.~\ref{fig:s_before}, Fig.~\ref{fig:t_before}, and Fig.~\ref{fig:m_before}, respectively, while after adaptation they are shown in Fig.~\ref{fig:s_after}, Fig.~\ref{fig:t_after}, and Fig.~\ref{fig:m_after}, respectively. 
As our proposed MPA method acts on the feature space and provides complementary feature alignment to both domains, their features are supposed to be more closely tied to their mutual prototypes, \ie, both domains go closer to each other bidirectionally. Comparing Fig.~\ref{fig:m_before} and Fig.~\ref{fig:m_after}, the proposed MPA method bridges the domain gap in the feature space and ties the feature distribution closer, such as mutual prototypes of \emph{sidewalk}, \emph{person}, \emph{rider}, and \emph{truck}.

\begin{figure}[t]
	\centering    
	\begin{subfigure}{\columnwidth}
    \includegraphics[width=1.0\columnwidth]{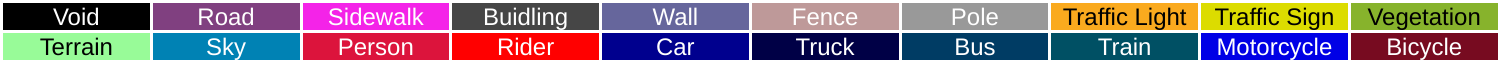}
    \centering \vskip -4ex
    \end{subfigure}
    \begin{subfigure}{\columnwidth}
    \includegraphics[width=1.0\columnwidth]{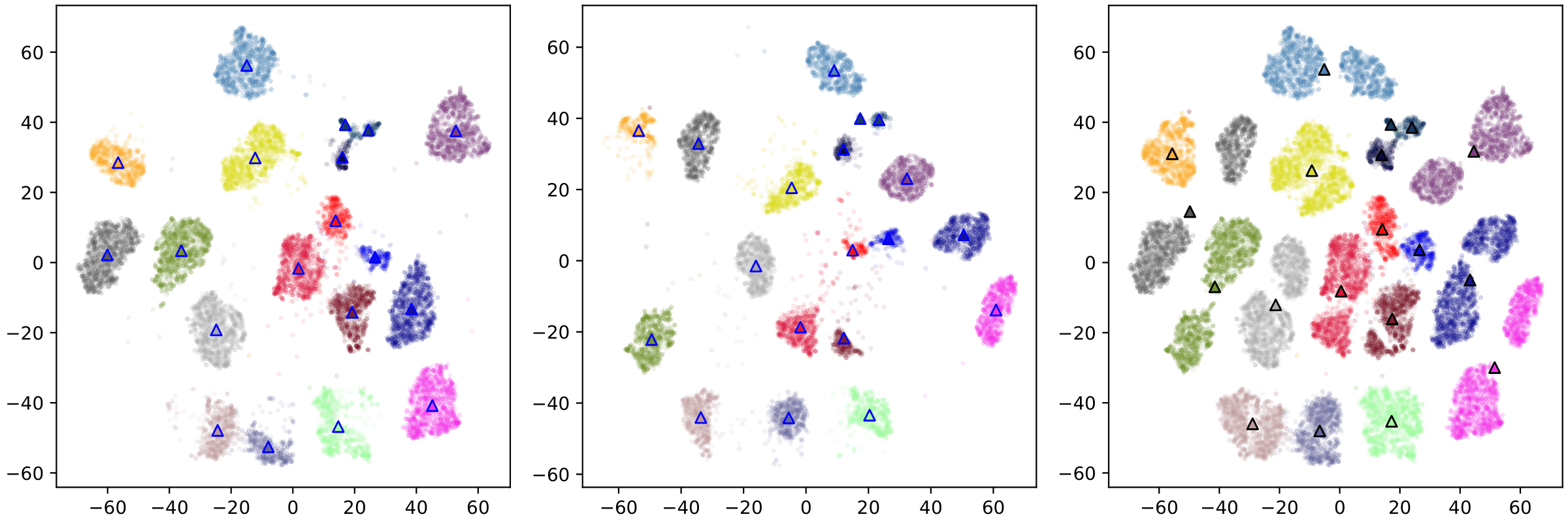}
    \begin{minipage}[t]{.33\columnwidth}
    \centering \vskip -4ex
    \subcaption{\small Source before}\label{fig:s_before}
    \end{minipage}%
    \begin{minipage}[t]{.33\columnwidth}
    \centering \vskip -4ex
    \subcaption{\small Target before}\label{fig:t_before}
    \end{minipage}%
    \begin{minipage}[t]{.33\columnwidth}
    \centering \vskip -4ex
    \subcaption{\small Mutual before}\label{fig:m_before}
    \end{minipage}%
    \end{subfigure}
    \begin{subfigure}{\columnwidth}
    \includegraphics[width=1.0\columnwidth]{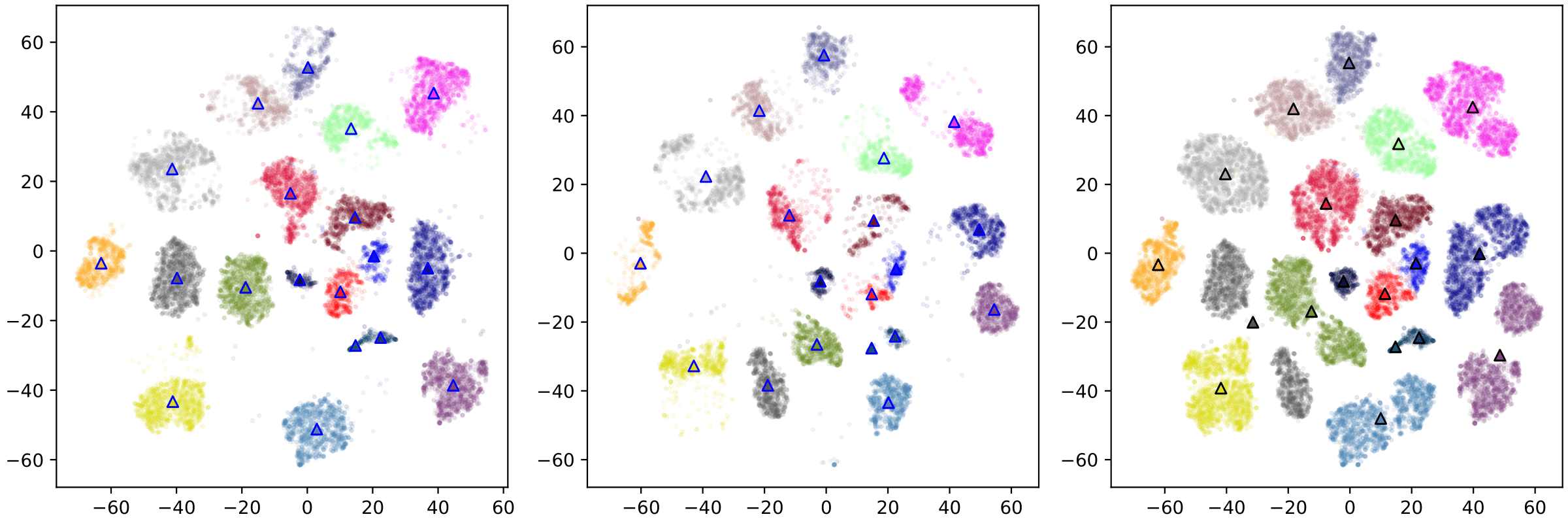}
    \begin{minipage}[t]{.33\columnwidth}
    \centering \vskip -3ex
    \subcaption{\small Source after}\label{fig:s_after}
    \end{minipage}%
    \begin{minipage}[t]{.33\columnwidth}
    \centering \vskip -3ex
    \subcaption{\small Target after}\label{fig:t_after}
    \end{minipage}%
    \begin{minipage}[t]{.33\columnwidth}
    \centering \vskip -3ex
    \subcaption{\small Mutual after}\label{fig:m_after}
    \end{minipage}%
    \end{subfigure}
    \vskip -2ex
	\caption{\textbf{t-SNE visualizations} before and after domain adaptation in outdoor scenes. \protect\marksymbol{triangle}{rblue} are the prototype of source or target domain and \protect\marksymbol{triangle}{black} represents the mutual prototype. {Zoom in for a better view.}} 
    \vskip -2ex
	\label{fig:supple_tsne_outdoor}
\end{figure}

\section{Conclusion}
\label{sec:conclusion}
\input{Tex_content/conclusion}

\ifCLASSOPTIONcompsoc
  \section*{Acknowledgments}
\else
  \section*{Acknowledgment}
\fi
This work was supported in part by the Federal Ministry of Labor and Social Affairs (BMAS) through the AccessibleMaps project under Grant 01KM151112, in part by the University of Excellence through the ``KIT Future Fields'' project, in part by the Helmholtz Association Initiative and Networking Fund on the HAICORE@KIT partition, and in part by Hangzhou SurImage Technology Company Ltd.
\bibliographystyle{IEEEtran}
\bibliography{bib}

\input{Tex_content/appendix}

\end{document}

%% file: Tex_content/abstract.tex
In this paper, we address panoramic semantic segmentation which %
is under-explored due to two critical challenges: {(1) image distortions and object deformations on panoramas; (2) lack of semantic annotations in the $360^\circ$ imagery}.
{To tackle these problems, first, we propose the upgraded Transformer for Panoramic Semantic Segmentation, \ie, Trans4PASS+, equipped with \emph{Deformable Patch Embedding (DPE)} and \emph{Deformable MLP (DMLPv2)} modules for handling object deformations and image distortions whenever (before or after adaptation) and wherever (shallow or deep levels).}
{Second, we enhance the \emph{Mutual Prototypical Adaptation (MPA)} strategy via pseudo-label rectification for unsupervised domain adaptive panoramic segmentation.} Third, aside from Pinhole-to-Panoramic (\textsc{Pin2Pan}) adaptation, we create a new dataset (SynPASS) with $9,080$ panoramic images, facilitating Synthetic-to-Real (\textsc{Syn2Real}) adaptation scheme in $360^\circ$ imagery. Extensive experiments are conducted, which cover indoor and outdoor scenarios, and each of them is investigated with \textsc{Pin2Pan} and \textsc{Syn2Real} regimens. Trans4PASS+ achieves state-of-the-art performances on four domain adaptive panoramic semantic segmentation benchmarks. Code is available at \href{https://github.com/jamycheung/Trans4PASS}{https://github.com/jamycheung/Trans4PASS}.

%% file: Tex_content/related_work.tex
\subsection{Semantic Segmentation}
Dense image semantic segmentation has experienced a steep increase in attention and great progress since Fully Convolutional Networks (FCN)~\cite{fcn} addressed it as an end-to-end per-pixel classification task.
Following FCN, subsequent efforts enhance the segmentation performance
by using encoder-decoder architectures~\cite{segnet,deeplabv3+},
aggregating high-resolution representations~\cite{refinenet,hrnet},
widening receptive fields~\cite{deeplabv2,pspnet,hou2020strip} and collecting contextual priors~\cite{context_encoding,context_prior,jin2021mining}.
Inspired by the non-local blocks~\cite{nonlocal}, self-attention~\cite{attention} is leveraged to establish long-range dependencies~\cite{danet,ccnet,ocnet,liu2020covariance,li2021ctnet} within FCNs.
Then, contemporary architectures appear to substitute convolutional backbones with transformer ones~\cite{vit,touvron2021deit}.
Thus, image understanding can be viewed via a perspective of sequence-to-sequence learning with dense prediction transformers~\cite{swin,pvt,dong2022cswin,li2022contextual_transformer,wu2021p2t} and semantic segmentation transformers~\cite{setr,segmenter,segformer,cheng2021maskformer,gu2022hrvit}.
{More recently, MLP-like architectures~\cite{mlp_mixer,cyclemlp,asmlp,hou2021vision} that alternate spatial- and channel mixing have sparked enormous interest in tackling visual recognition tasks.}

However, most of these methods are designed for narrow-FoV pinhole images and often have large accuracy downgrades when applied in the $360^\circ$ domain for holistic panorama-based perception.
{In this work, we address panoramic semantic segmentation, with a novel distortion-aware transformer architecture that considers a broad FoV already in its design and handles the panorama-specific semantic distribution via parallel MLP-based, channel-wise mixing, and pooling mixing mechanisms.}

\subsection{Panoramic Segmentation}
Capturing wide-FoV scenes, panoramic images~\cite{gao2022review} act as a starting point for a more complete scene understanding. Mainstream outdoor omnidirectional semantic segmentation systems rely on fisheye cameras~\cite{restricted,woodscape,petrovai2022semantic_cameras} or panoramic images~\cite{synthetic,orhan2021semantic_outdoor_panoramic,hu2022distortion_panoramic}. Panoramic panoptic segmentation is also addressed in recent surrounding parsing systems~\cite{pps,mei2022waymo,pps_insights}, where the video segmentation pipeline with the Waymo open dataset~\cite{mei2022waymo} has a coverage of $220^\circ$.
Indoor methods, on the other hand, focus on either distortion-mitigated representations~\cite{distortion_aware,spherical_unstructured_grids,spherephd,equivariant_networks,zheng2022complementary_bidirectional} or multi-tasks schemes~\cite{pano_sfmlearner,hohonet,zhang2021deeppanocontext}.
Yet, most of these works are developed based on the assumption that densely labeled images are implicitly or partially available in the target domain of panoramic images for training a segmentation model.

However, the acquisition of dense pixel-wise labels is extremely labor-intensive and time-consuming, in particular for panoramas with higher complexities and more small objects implicated in wide-FoV observations.
We cut the requirement for labeled target data and circumvent the prohibitively expensive annotation process of determining pixel-level semantics in unstructured real-world surroundings.
{Different from previous works, we look into panoramic semantic segmentation via the lens of unsupervised transfer learning and investigate both Synthetic-to-Real (\textsc{Syn2Real}) and Pinhole-to-Panoramic (\textsc{Pin2Pan}) adaptation strategies to profit from rich, readily available datasets like synthetic panoramic or annotated pinhole datasets.}
In experiments, our panoramic segmentation transformer architecture generalizes to both indoor and outdoor $360^\circ$ scenes.

\subsection{Dynamic and Deformable Vision Transformers}
With the prosperity of vision transformers in the field, some research works develop architectures with dynamic properties.
In earlier works, the anchor-based DPT~\cite{chen2021dpt} and the non-overlapping DAT~\cite{xia2022dat} use deformable designs only in later stages of the encoder and borrow Feature Pyramid Network (FPN) decoders from CNN counterparts.
PS-ViT~\cite{yue2021psvit} utilizes a progressive sampling module to locate discriminative regions, whereas Deformable DETR~\cite{deformable_detr} leverages deformable attention to enhance feature maps.
Further, some methods aim to improve the efficiency of vision transformers by adaptively optimizing the number of informative tokens~\cite{wang2021not_dynamic,rao2021dynamicvit,yin2021avit,xu2021evo_vit} or dynamically modeling relevant dependencies via query grouping~\cite{liu2022dynamic_group_transformer}.
{Unlike these previous works limited to narrow-FoV images, our distortion-aware segmentation transformer is designed for pixel-dense prediction tasks on wide-FoV images, and can better adapt to panoramas by learning to counteract severe deformations in the data.}

\subsection{Unsupervised Domain Adaptation}
Domain adaptation has been thoroughly studied to improve model generalization to unseen domains, \eg, adapting to the real world from synthetic data collections~\cite{ros2016synthia,richter2016playing_gta5}.
Two predominant categories of unsupervised domain adaptation fall either in self-training~\cite{curriculum_da,li2022class_self_labeling,huo2022domain_agnostic_prior,zhu2021improving_efficient_self_training,lai2022decouplenet,xie2022sepico} or adversarial learning~\cite{cycada,adaptsegnet,all_about_structure}.
{Self-training methods usually generate pseudo-labels to gradually adapt through iterative improvement~\cite{pycda}, whereas adversarial solutions build on the idea of GANs~\cite{gan} to conduct image translation~\cite{cycada,li2019bidirectional}, or enforce alignment in layout matching~\cite{contextual_relation_consistent_da} and feature agreement~\cite{clan,densepass}.
Further adaptation flavors, consider model ensembling~\cite{fda,maximum_squares_loss}, category-level alignment~\cite{wang2020differential,jiang2022proca}, adversarial entropy minimization~\cite{intra_da}, and vision transformers~\cite{hoyer2021daformer}.}

Relevant to our task, PIT~\cite{gu2021pit} handles the gap of camera intrinsic parameters with FoV-varying adaptation, whereas P2PDA~\cite{densepass} first tackles \textsc{Pin2Pan} transfer by learning attention correspondences.
Aside from distortion-adaptive architecture design, we revisit panoramic semantic segmentation from a prototype adaptation-based perspective where panoramic knowledge is distilled via class-wise prototypes.
{Differing from recent methods utilizing individual prototypes for source- and target domain~\cite{yue2021pcs,proda}, we present mutual prototypical adaptation, which jointly exploits dual-domain feature embeddings. Besides, we enhance MPA by using SAM~\cite{kirillov2023SAM} to rectify target pseudo labels, which boosts transfer beyond the FoV.
Moreover, we study both \textsc{Pin2Pan} and \textsc{Syn2Real} for learning robust panoramic semantic segmentation.}

%

%% file: tables/tab1_synpass_stat.tex
\begin{table}[!t]
\centering
\setlength\tabcolsep{2.0pt}
\caption{\small {\textbf{SynPASS dataset} for panoramic semantic segmentation, including four adverse weather conditions and two illuminations.}}
\resizebox{\columnwidth}{!}{
\renewcommand{\arraystretch}{1.3}
\begin{tabular}{l|cccc|c} 
\toprule
 & \textbf{Cloudy} & \textbf{Foggy} & \textbf{Rainy} & \textbf{Sunny} & \textbf{ALL} \\ \midrule \midrule[0.5pt]
Split&\textit{train / val / test} & \textit{train / val / test} & \textit{train / val / test} & \textit{train / val / test} & \textit{train / val / test} \\
\#Frames & 1420 / 420 / 430&1420 / 430 / 420&1420 / 430 / 420&1440 / 410 / 420&5700 / 1690 / 1690 \\ \midrule
Split&\textit{day / night} & \textit{day / night} & \textit{day / night} & \textit{day / night} & \textit{day / night} \\
\#Frames & 1980 / 290 & 1710 / 560 & 2040 / 230 & 1970 / 300 & 7700 / 1380\\ \midrule
Total & 2270 & 2270 & 2270 & 2270 & 9080 \\
\bottomrule
\end{tabular}
}
\label{table:synpass_stat}
\end{table}

%% file: tables/tab2_sota_synpass.tex
\begin{table}[!t]
\footnotesize
\centering
\setlength\tabcolsep{2.0pt}
\caption{\small {\textbf{SynPASS benchmark} is evaluated on full 22 classes and is divided into four weather conditions, day- and night-time.}}
\resizebox{\columnwidth}{!}{
\renewcommand{\arraystretch}{0.99}
\begin{tabular}{l|cccc|cc|cc} 
\toprule
\textbf{Method} & \textbf{Cloudy} & \textbf{Foggy} & \textbf{Rainy} & \textbf{Sunny} & \textbf{Day}& \textbf{Night}& \multicolumn{2}{c}{\textbf{ALL}} \\ 
 & \textit{val} & \textit{val} & \textit{val} & \textit{val} & \textit{val} & \textit{val} & \textit{val} & \textit{test} \\
\midrule[0.5pt]\midrule[0.5pt]
Fast-SCNN &30.84&22.68&26.16&27.19&29.68&24.75&26.31&21.30 \\
DeepLabv3+ (MNv2) &38.94&35.19&35.43&37.73&36.01&30.55&36.72&29.66 \\
HRNet (W18Small) &42.92&37.94&37.37&41.45&39.19&32.22&39.80&34.09 \\ \midrule
PVT (Tiny) & 39.92&34.99&34.01&39.84&36.71&27.36&36.83&32.37 \\
PVT (Small) & 40.75&36.14&34.29&40.14&37.92&28.80&37.47&32.68 \\
SegFormer (B1) & 45.34&41.43&40.33&44.36&42.97&33.15&42.68&37.36 \\
SegFormer (B2) & 46.07&40.99&40.10&44.35&44.08&33.99&42.49&37.24 \\
Trans4PASS (Tiny) & 46.90&41.97&41.61&45.52&44.48&34.73&43.68&38.53 \\
Trans4PASS (Small) & 46.74&43.49&43.39&45.94&45.52&37.03&44.80&38.57 \\
Trans4PASS+ (Tiny) & 47.85& 43.38& 43.40& 46.83& 45.99& 36.46&45.17&39.42 \\
\rowcolor{gray!15} Trans4PASS+ (Small) & \textbf{48.85}& \textbf{44.64}& \textbf{45.13}& \textbf{48.29}& \textbf{46.49}& \textbf{37.33}&\textbf{46.61}&\textbf{40.72} \\
\textit{w.r.t. SegFormer (B2)} & \green{+2.78}& \green{+3.65}& \green{+5.03}& \green{+3.94}& \green{+2.41}& \green{+3.34}& \green{+4.12}& \green{+3.48} \\
\bottomrule
\end{tabular}
}
\label{table:synpass_benchmark}
\end{table}

%% file: tables/tab3_sota_CS_DP.tex
\begin{table}[!t]
\centering
\caption{\small {\textbf{Performance gaps} of CNN- and transformer-based models from Cityscapes (\textbf{CS})~$@$~1024${\times}$512 to DensePASS (\textbf{DP}). Trans4PASS+$^*$ models apply DMLPv2 in Mask2Former head.}}
\vskip-1ex
\resizebox{\columnwidth}{!}{
\setlength\tabcolsep{4.0pt}
\renewcommand{\arraystretch}{0.89}
\begin{tabular}{ll|rrrrc}
    \toprule
    \textbf{Network} & \textbf{Backbone} & \textbf{FLOPs (G)}&	\textbf{\#P (M)} & \textbf{CS} & \textbf{DP} & \textbf{mIoU Gaps}
    \\ \midrule[0.5pt]\midrule[0.5pt]
SwiftNet~\cite{swiftnet} & ResNet-18 & 9.3 & 8.5& 75.4 & 25.7 & -49.7 \\ 
Fast-SCNN~\cite{fastscnn}     & Fast-SCNN	   & 0.9 & 1.4& 69.1  &  24.6   & -44.5 \\
ERFNet~\cite{erfnet} & ERFNet    & 14.7 & 2.1& 72.1  & 16.7 &  -55.4  \\ 
FANet~\cite{fanet} & ResNet-34 & 8.1 & n.a& 71.3 & 26.9 & -44.4 \\
PSPNet~\cite{pspnet}            & ResNet-50    & 179.0 & 46.6& 78.6  &  29.5  & -49.1 \\
OCRNet~\cite{ocrnet}            & HRNetV2p-W18 & 53.6 & 12.1& 78.6  & 30.8  & -47.8 \\ 
DeepLabV3+~\cite{deeplabv3+} & ResNet-101   & 254.0 & 60.2& 80.9  &  32.5  & -48.4 \\
DANet~\cite{danet}                 & ResNet-101   & 289.0 & 66.5& 80.4  &  28.5   & -51.9 \\
DNL~\cite{dnl}                   & ResNet-101   & 286.0 & 66.7& 80.4  & 32.1  & -48.3 \\
Semantic-FPN~\cite{panopticfpn} & ResNet-101 & 64.9 & 47.5& 75.8 &  28.8  & -47.0 \\
ResNeSt~\cite{resnest}         & ResNeSt-101  & 283.0 & 69.8& 79.6  &  28.8   & -50.8 \\
OCRNet~\cite{ocrnet}            & HRNetV2p-W48 & 163.0 & 70.4& 80.7  &  32.8  & -47.9 \\ \midrule
SETR-Naive~\cite{setr} & Transformer-L & 363.0 & 306.0& 77.9 & 36.1 & -41.8 \\
SETR-MLA~\cite{setr} & Transformer-L & 367.0 & 311.0& 77.2 & 35.6 & -41.6 \\
SETR-PUP~\cite{setr} & Transformer-L & 417.0 & 310.0& 79.3 & 35.7 & -43.6 \\
SegFormer~\cite{segformer} & SegFormer-B1 & 15.5 & 13.7& 78.5 & 38.5 & -40.0 \\
SegFormer~\cite{segformer} & SegFormer-B2 & 25.3 & 24.7& 81.0 & 42.4 & -38.6 \\
Trans4PASS &  Trans4PASS (T) & 12.0 & 13.9& 79.1 & 41.5 & -37.6   \\
Trans4PASS &  Trans4PASS (S) & 20.0 & 25.0& 81.1 & 44.8 & -36.3  \\
MaskFormer~\cite{cheng2021maskformer} & SegFormer-B1 & 42.9 & 27.9& 66.9 & 35.0 & -31.9 \\
MaskFormer~\cite{cheng2021maskformer} & SegFormer-B2 & 52.8 & 39.0& 78.8 & 46.1 & -32.7 \\
Mask2Former~\cite{cheng2022mask2former} & Swin-T  & 73.3 & 47.4& 81.7 & 39.9 & -41.8\\
Mask2Former~\cite{cheng2022mask2former} & Swin-S  & 97.1 & 68.8& 82.6 & 45.8 & -36.8\\
Mask2Former~\cite{cheng2022mask2former} & SegFormer-B1 & 58.9 & 33.6& 77.6 & 46.5 & -31.1\\
Mask2Former~\cite{cheng2022mask2former} & SegFormer-B2 & 68.7 & 44.6& 80.2 & 46.4 & -33.8\\
Trans4PASS+ &  Trans4PASS+ (T)  & 11.7 & 14.0& 78.6 & 41.6 & -37.0 \\
\rowcolor{gray!15} Trans4PASS+ & Trans4PASS+ (S)  & 19.8 & 25.0& 80.7 & 46.5 & -34.2 \\
Trans4PASS+$^*$&  Trans4PASS+ (T)  & 55.2 & 33.9& 77.8 & 46.8 & -31.0 \\
\rowcolor{gray!15} Trans4PASS+$^*$ & Trans4PASS+ (S)  & 63.2 & 44.9& 79.6 & \textbf{48.4} & -31.2 \\

    \bottomrule
\end{tabular}}
\label{table:outdoor_domain_gap}
\end{table}

%% file: tables/tab4_sota_SPin_SPan.tex
\begin{table}[!t]
\footnotesize
\centering
\caption{\small \textbf{Performance gaps} from Stanford2D3D-Pinhole (\textbf{SPin}) to Stanford2D3D-Panoramic (\textbf{SPan}) dataset on fold-1.}
\vskip -2ex
\resizebox{\columnwidth}{!}{
\renewcommand{\arraystretch}{0.89}
\begin{tabular}{ll|rrc}
    \toprule
    \textbf{Network} & \textbf{Backbone} & \textbf{SPin} & \textbf{SPan} & \textbf{mIoU Gaps}
    \\ \midrule[0.5pt]\midrule[0.5pt]
    Fast-SCNN~\cite{fastscnn} & Fast-SCNN & 41.71 & 26.86 & -14.85 \\
    SwiftNet~\cite{swiftnet} & ResNet-18 & 42.28 & 34.95 & -7.87 \\
    DANet~\cite{danet} & ResNet-50 & 43.33 & 37.76 & -5.57 \\
    DANet~\cite{danet} & ResNet-101 & 40.09 & 31.81 & -8.28 \\ 
    \midrule 
    Trans4Trans-T~\cite{zhang2021trans4trans_acvr} & PVT-T & 41.28 & 24.45 & -16.83 \\ 
    Trans4Trans-S~\cite{zhang2021trans4trans_acvr} & PVT-S & 44.47 & 23.11 & -21.36 \\
    Trans4PASS (T) &  Trans4PASS (T)  & 49.05 & 46.08 & -2.97  \\
    Trans4PASS (S) &  Trans4PASS (S)  & 50.20 & {48.34} & -1.86  \\
    Trans4PASS+ (T) &  Trans4PASS+ (T)  &  48.69 & 46.32 & -2.37 \\
    \rowcolor{gray!15} Trans4PASS+ (S) &  Trans4PASS+ (S)  & {51.48} & \textbf{49.76} & -1.72  \\
    \bottomrule
\end{tabular}}
\label{table:indoor_domain_gap}
\end{table}

%% file: tables/tab5_s2r_p2p_gaps.tex
\begin{table}[!t]
\footnotesize
\centering
\caption{\small {\textbf{\textsc{Syn2Real} \vs \textsc{Pin2Pan} domain gaps.}}}
\vskip -2ex
\resizebox{\columnwidth}{!}{
\setlength\tabcolsep{12.0pt}
\renewcommand{\arraystretch}{0.8}
\begin{tabular}{l|lll}
  \toprule
  \circled{1} \textbf{\textit{Outdoor \textsc{Pin2Pan}}:} & \textbf{CS13}  & \textbf{DP13}   & \textbf{mIoU Gaps}  \\ \midrule
  PVT  & 65.88& 46.19& -19.69\\
  Trans4PASS  & 75.21& 50.96~\gbf{+4.77}& -24.25\\ 
  \rowcolor{gray!15} Trans4PASS+  &  77.04&\textbf{51.48}~\gbf{+5.29}&-25.56 \\ \toprule\toprule
  \circled{2} \textbf{\textit{Outdoor \textsc{Syn2Real}}:} & \textbf{SP13}  & \textbf{DP13}   & \textbf{mIoU Gaps}  \\ \midrule
  PVT  & 52.94& 38.74& -14.20\\ 
  Trans4PASS  & 62.76& 43.18~\gbf{+4.44}& -19.58\\ 
  \rowcolor{gray!15} Trans4PASS+  &  63.21&\textbf{43.83}~\gbf{+5.09}&-19.38 \\\toprule\toprule
  \circled{3} \textbf{\textit{Indoor \textsc{Pin2Pan}}:}  & \textbf{SPin8} & \textbf{SPan8} & \textbf{mIoU Gaps}  \\ \midrule
  PVT  & 60.46 & 57.71 & -2.75 \\
  Trans4PASS  & 66.51 & 62.39~\gbf{+4.68} & -4.12 \\ 
  \rowcolor{gray!15} Trans4PASS+  &  67.28&\textbf{63.73}~\gbf{+6.02}&-3.55 \\ \toprule\toprule
  \circled{4} \textbf{\textit{Indoor \textsc{Syn2Real}}:}  & \textbf{S3D8} & \textbf{SPan8} & \textbf{mIoU Gaps}  \\ \midrule
  PVT  & 66.46 & 45.82 & -20.64 \\
  Trans4PASS  & 77.29 & 51.70~\gbf{+5.88} & -25.59 \\ 
  \rowcolor{gray!15} Trans4PASS+  &  76.04& \textbf{52.09}~\gbf{+6.27} & -23.95 \\\bottomrule
  \end{tabular}
}
\label{table:comp_domain_gap}
\vskip -1ex
\end{table}

%% file: tables/tab6_s2r_transfer.tex
\begin{table}[!t]
\footnotesize
\centering
\setlength\tabcolsep{12.0pt}
\caption{\small \textbf{Comparison of \textsc{Syn2Real} transfer learning} between methods followed Structured3D~\cite{structured3d}. Synthetic: S3D8, Real: SPan8.}
\vskip -2ex
\resizebox{\columnwidth}{!}{
\renewcommand{\arraystretch}{0.89}
\begin{tabular}{ll|ll}
\toprule
\textbf{{Indoor \textsc{Syn2Real}}:} & \textbf{Data}  & \textbf{S3D8} & \textbf{SPan8}  \\ \midrule\midrule
UPerNet (ResNet-50) & Synthetic & - & 28.75 \\
UPerNet (ResNet-50) & Synthetic+Real & - & 49.60 \\
HRNet (W18) & Synthetic & - & 37.92 \\
HRNet (W18) & Synthetic+Real & - & \textbf{52.00} \\ \midrule
Trans4PASS+ (Tiny) & Synthetic & 77.03 & 50.29 \\
\rowcolor{gray!15} Trans4PASS+ (Small) & Synthetic & 76.04 & \textbf{52.09}  \\ \bottomrule
\end{tabular}
}
\label{table:comp_in_syn2real}
\vskip -2ex
\end{table}

%% file: tables/tab7_str_abl_1.tex
\begin{table}[t]
\centering
\caption{\small {Ablation study of Trans4PASS+, including DEP, DMLPv1, and DMLPv2. Models are trained on Cityscapes~(\textbf{CS})~$@$~512${\times}$512 and tested on DensePASS~(\textbf{DP})~$@$~2048${\times}$400. {\#P}: \#parameters in millions.}}
\vskip -2ex
\label{tab:str_abl_1}
\resizebox{\columnwidth}{!}{
\renewcommand{\arraystretch}{0.9}
\begin{tabular}{ccc|cccl}
\toprule
DPE & DMLPv1 & DMLPv2 & GFLOPs & \#P & \textbf{CS} & \textbf{DP} \\
\hline\midrule
 &  &  & 13.27 & 13.66 & 74.93 & 39.02 \\
 & \checkmark &  & 11.82 & 13.92 & 73.10 & 45.14 \\
 &  & \checkmark & 11.53 & 13.94 & 65.86 & 46.94 \\\midrule
\checkmark &  &  & 08.90 & 13.29 & 71.70 & 43.11 \\
\checkmark & \checkmark &  & 12.02 & 13.93 & 72.49 & 45.89 \\
\checkmark & \multicolumn{1}{c}{} & \checkmark & 11.74 & 13.96 & 72.63 & \textbf{49.94}\\
\bottomrule
\end{tabular}
}
\end{table}

%% file: tables/tab8_str_abl_2.tex
\begin{table}[t]
\centering
\caption{\small {Ablation study of PX and CX in DMLPv2 of Trans4PASS+. {CX}: Channel Mixer, {PX}: Pooling Mixer.}}
\vskip -2ex
\label{tab:str_abl_2}
\setlength\tabcolsep{10.0pt}
\renewcommand{\arraystretch}{0.9}
\resizebox{\columnwidth}{!}{
\begin{tabular}{cc|cccc}
\toprule
PX & CX & GFLOPs & \#P & \textbf{CS} & \textbf{DP} \\
\hline \midrule
&  & 13.27 & 13.66 & 74.93 & 39.02 \\
\checkmark &  & 10.22 & 13.60 & 72.62 & 46.66 \\
& \checkmark & 10.96 & 13.78 & 74.07 & 47.66 \\
\checkmark & \checkmark & 11.74 & 13.96 & 72.63 & \textbf{49.94}\\ 
\bottomrule
\end{tabular}

}
\end{table}

%% file: tables/tab9_comp_PE_MLP.tex
\begin{table}[t]
\centering
\caption{\small {Comparison of different PE and MLP methods.}}
\vskip -2ex
\label{tab:comp_PE_MLP}
\setlength\tabcolsep{4.0pt}
\renewcommand{\arraystretch}{0.9}
\resizebox{\columnwidth}{!}{
\begin{tabular}{ll|cccc}
\toprule
PE & MLP & GFLOPs & \#P & \textbf{CS} & \textbf{DP} \\
\hline\midrule
Vanilla PE & Vanilla MLP & 13.27 & 13.66 & 74.93 & 39.02 \\\midrule 
\cellcolor{gray!10}DPT~\cite{chen2021dpt} & DMLPv1 & 13.11 & 13.10 & 69.48 & 36.50 \\
\cellcolor{gray!10}Deformable Conv & DMLPv1& 11.98&	13.93&	71.15&	42.54 \\
\cellcolor{gray!10}Vanilla PE & DMLPv1 & 11.82 & 13.92 & 73.10 & 45.14 \\
\cellcolor{gray!10}DPE & DMLPv1 & 12.02 & 13.93 & 72.49 & 45.89 \\\midrule
DPE & \cellcolor{gray!10}Vanilla MLP & 08.90 & 13.29 & 71.70 & 43.11 \\
DPE & \cellcolor{gray!10}CycleMLP~\cite{cyclemlp} & 09.83 & 13.60 & 73.49 & 40.16 \\
DPE & \cellcolor{gray!10}ASMLP~\cite{asmlp} & 13.40 & 14.19 & 73.65 & 42.05 \\
DPE & \cellcolor{gray!10}DMLPv1 & 12.02 & 13.93 & 72.49 & 45.89 \\
DPE & \cellcolor{gray!10}DMLPv2 & 11.74 & 13.96 & 72.63 & \textbf{49.94}\\\midrule
\cellcolor{gray!10}Deformable Conv & DMLPv2 & 11.70 &13.95 &72.20 &45.74 \\
\cellcolor{gray!10}Vanilla PE & DMLPv2 & 11.54& 13.94& 73.39& 47.08 \\
\cellcolor{gray!10}DPE & DMLPv2 & 11.74 & 13.96 & 72.63 & \textbf{49.94}\\
\bottomrule
\end{tabular}
}
\end{table}

%% file: tables/tab10_token_mixing.tex
\begin{table}[t]
\centering
\caption{\small {Comparison of different token mixing methods.}}
\vskip -2ex
\label{tab:token_mixing}
\setlength\tabcolsep{8.0pt}
\renewcommand{\arraystretch}{0.9}
\resizebox{\columnwidth}{!}{
\begin{tabular}{l|cccc}
\toprule
 & GFLOPs & \#P & \textbf{CS} & \textbf{DP} \\
\hline\midrule
PoolFormer~\cite{yu2021metaformer} & 09.47 & 13.47 & 70.52 & 43.18 \\
FAN~\cite{zhou2022fan} & 10.96 & 13.81 & 71.15 & 42.54 \\
PoolFormer+FAN & 10.96 & 13.81 & 72.21 & 45.97 \\
DMLPv2 & 11.74 & 13.96 & 72.63 & \textbf{49.94}\\
\bottomrule
\end{tabular}
}
\end{table}

%% file: tables/table_mpa.tex
\begin{subtable}[ht]{\textwidth}
\setlength{\tabcolsep}{4pt}
    \begin{center}
	\resizebox{\textwidth}{!}{      
    \renewcommand{\arraystretch}{0.9}
	\begin{tabular}{ @{}l | c | c c c c c c c c c c c c c c c c c c c@{}}
        \toprule[1pt]
        Method & \rotatebox{90}{mIoU} &  \rotatebox{90}{Road} &  \rotatebox{90}{S.walk} &  \rotatebox{90}{Build.} & \rotatebox{90}{Wall} &  \rotatebox{90}{Fence} &  \rotatebox{90}{Pole} & \rotatebox{90}{Tr. light} &  \rotatebox{90}{Tr. sign}&  \rotatebox{90}{Veget.} &  \rotatebox{90}{Terrain} &  \rotatebox{90}{Sky} & \rotatebox{90}{Person} &  \rotatebox{90}{Rider} & \rotatebox{90}{Car} &  \rotatebox{90}{Truck}& \rotatebox{90}{Bus}& \rotatebox{90}{Train}& \rotatebox{90}{M.cycle}&  \rotatebox{90}{Bicycle}\\
        \toprule[1pt]
        ERFNet~\cite{erfnet} & 16.65 & 63.59 & 18.22 & 47.01 & 9.45 & 12.79 & 17.00 & 8.12 & 6.41 & 34.24 & 10.15 & 18.43 & 4.96 & 2.31 & 46.03 & 3.19 & 0.59 & 0.00 & 8.30 & 5.55 \\
        PASS (ERFNet)~\cite{pass} & 23.66 & 67.84 & 28.75 & 59.69 & 19.96 & 29.41 & 8.26 & 4.54 & 8.07 & 64.96 & 13.75 & 33.50 & 12.87 & 3.17 & 48.26 & 2.17 & 0.82 & 0.29 & 23.76& 19.46 \\
        ECANet (Omni-supervised)~\cite{omnirange} & 43.02 & {81.60} & 19.46 & 81.00 & 32.02 & 39.47 & 25.54 & 3.85 & 17.38 & 79.01 & 39.75 & {94.60} & 46.39 & 12.98 & {81.96} & 49.25 & 28.29 & 0.00 & 55.36 & 29.47 \\
        \midrule
        CLAN (Adversarial)~\cite{clan} & 31.46 & 65.39 & 21.14 & 69.10 & 17.29 & 25.49 & 11.17 & 3.14 & 7.61 & 71.03 & 28.19 & 55.55 & 18.86 & 2.76 & 71.60 & 26.42 & 17.99 & 59.53 & 9.44 & 15.91\\
        MIC (Self-training)~\cite{hoyer2023mic} & 31.65 &62.94 &26.19 &71.90 &11.41 &19.44 &20.16 &01.77 &11.41 &69.06 &25.24 &93.76 &30.01 &07.64 &65.00 &26.97 &0.78 &19.28 &27.98 &10.50 \\
        CRST (Self-training)~\cite{crst} & 31.67 & 68.18 & 15.72 & 76.78 & 14.06 & 26.11 & 9.90 & 0.82 & 2.66 & 69.36 & 21.95 & 80.06 & 9.71 & 1.25 & 65.12 & 38.76 & 27.22 & 48.85 & 7.10 & 18.08\\
        HRDA (Adversarial)~\cite{hoyer2022hrda} & 38.84 & 69.77& 29.57& 73.90& 18.40& 18.99& 19.73& 3.09& 8.34& 72.69& 18.63& 93.11& 37.82& 12.82& 73.65& 50.22& 21.39& 59.35& 46.15& 10.29\\
        P2PDA (Adversarial)~\cite{p2pda_trans} & 41.99 & 70.21 & 30.24 & 78.44 & 26.72 & 28.44 & 14.02 & 11.67 & 5.79 & 68.54 & 38.20 & 85.97 & 28.14 & 0.00 & 70.36 & 60.49 & 38.90 & 77.80 & 39.85 & 24.02\\
        SIM (Self-training)~\cite{wang2020differential} & 44.58 & 68.16 & 32.59 & 80.58 & 25.68 & 31.38 & 23.60 & 19.39 & 14.09 & 72.65 & 26.41 & 87.88 & 41.74 & 16.09 & 73.56 & 47.08 & 42.81 & 56.35 & 47.72 & 39.30\\
        PCS (Self-training)~\cite{yue2021pcs} & 53.83 & 78.10 & 46.24 & 86.24 & 30.33 & {45.78} & 34.04 & 22.74 & 13.00 & {79.98} & 33.07 & 93.44 & 47.69 & 22.53 & 79.20 & 61.59 & 67.09 & 83.26 & 58.68 & 39.80 \\
        DAFormer (Self-training)~\cite{hoyer2021daformer} &54.56 & 71.96& 27.70& 87.49& \textbf{36.70}& 45.17& 35.55& \textbf{28.58}& 13.69& 79.24& 26.62& \textbf{94.95}& 54.59& 21.51& 77.99& \textbf{70.90}& 56.40& \textbf{94.47}& 65.34& {47.73} \\
        \midrule
        USSS (IDD)~\cite{usss} & 26.98 & 68.85 & 5.41 & 67.39 & 15.10 & 21.79 & 13.18 & 0.12 & 7.73 & 70.27 & 8.84 & 85.53 & 22.05 & 1.71 & 58.69 & 16.41 & 12.01 & 0.00 & 23.58 & 13.90 \\        
        USSS (Mapillary)~\cite{usss} & 30.87 & 71.01 & 31.85 & 76.79 & 12.13 & 23.61 & 11.93 & 3.23 & 10.15 & 73.11 & 31.24 & 89.59 & 16.05 & 3.86 & 65.27 & 24.46 & 18.72 & 0.00 & 9.08 & 14.48\\
        Seamless (Mapillary)~\cite{seamless} & 34.14 & 59.26 & 24.48 & 77.35 & 12.82 & 30.91 & 12.63 & 15.89 & 17.73 & 75.61 & 33.30 & 87.30 & 19.69 & 4.59 & 63.94 & 25.81 & 57.16 & 0.00 & 11.59 & 19.04 \\
        SwiftNet (Cityscapes)~\cite{swiftnet} & 25.67 & 50.73 & 32.76 & 70.24 & 12.63 & 24.02 & 18.79 & 7.18 & 4.01 & 64.93 & 23.70 & 84.29 & 14.91 & 0.97 & 43.46 & 8.92 & 0.04 & 4.45 & 12.77 & 8.77 \\
        SwiftNet (Merge3)~\cite{issafe} & 32.04 & 68.31 & 38.59 & 81.48 & 15.65 & 23.91 & 20.74 & 5.95 & 0.00 & 70.64 & 25.09 & 90.93 & 32.66 & 0.00 & 66.91 & 42.30 & 5.97 & 0.07 & 6.85 & 12.66 \\
        \midrule
        Trans4PASS (S) (ours) & {55.25} & 78.39 & 41.62 & {86.47} & 31.56 & 45.47 & 34.02 & 22.98 & 18.33 & 79.63 & 41.35 & 93.80 & 49.02 & 22.99 & 81.05 & 67.43 & {69.64} & 86.04 & 60.85 & 39.20\\
        Trans4PASS (S) (ours)* & {56.38} & 79.91 & 42.68 & 86.26 & 30.68 & 42.32 & \textbf{36.61} & {24.81} & {19.64} & 78.80 & {44.73} & 93.84 & {50.71} & {24.39} & 81.72 & {68.86} & 66.18 & {88.62} & {63.87} & {46.62}\\
        \rowcolor{gray!15} Trans4PASS+ (S) (ours) & {57.03} & 79.74& 50.27& {86.59}& 29.57& 44.38& 29.82& 24.81& 18.51& 79.25& \textbf{45.91}& 93.34& 52.68& 25.00& 81.15& 70.35& \textbf{76.80}& 89.08& 63.83& 42.45\\
        \rowcolor{gray!15} Trans4PASS+ (S) (ours)* & \textbf{59.43} & \textbf{82.02}& \textbf{55.24}& \textbf{86.71}& 28.97& \textbf{47.94}& 30.67& 28.03& \textbf{19.75}& \textbf{80.01}& 45.42& 94.20& \textbf{56.35}& \textbf{37.79}& \textbf{84.16}& 70.44& 72.95& 91.40& \textbf{67.92}& \textbf{49.12} \\
        \bottomrule[1pt]
        \end{tabular}
    }
    \end{center}
    \vskip-1ex
    \caption{\textbf{Per-class results on DensePASS.} Comparison with state-of-the-art panoramic segmentation \cite{omnirange,pass}, domain adaptation~\cite{clan,yue2021pcs,crst,p2pda_trans,wang2020differential,hoyer2021daformer,hoyer2023mic,hoyer2022hrda}, and multi-supervision methods~\cite{usss,seamless,issafe}. 
    *~denotes performing Multi-Scale (MS) evaluation.}
    \label{tab:more}
\end{subtable}
%
\begin{subtable}[ht]{0.31\textwidth}
	\begin{center}
	\footnotesize
	\setlength{\tabcolsep}{1mm}
	\resizebox{\textwidth}{!}{
    \renewcommand{\arraystretch}{1.0}
    \begin{tabular}{ l | c | c }
    \toprule
    \textbf{Network} & \textbf{Method} &  \textbf{mIoU(\%)} \\
    \toprule
    FANet & P2PDA & 35.67 \\
    DANet & P2PDA & 41.99 \\
    Trans4PASS (T) & P2PDA & 51.05 \\
    Trans4PASS (S) & P2PDA & 52.91 \\
    \midrule 
    Trans4PASS (T) & -  & 45.89  \\
    \rowcolor{gray!15} Trans4PASS+ (T) & -  & 49.94  \\
    Trans4PASS (T) & MPA + SSL + MS  & {54.72} \\
    \rowcolor{gray!15} Trans4PASS+ (T) & MPA + SAM + MS  & 57.67 \\
    \midrule 
    Trans4PASS (S) & -  & 48.73 \\
    \rowcolor{gray!15} Trans4PASS+ (S) & -  & 53.07 \\
    Trans4PASS (S) & MPA + SSL + MS  & {56.38} \\
    \rowcolor{gray!15} Trans4PASS+ (S) & MPA + SAM + MS  &  \textbf{59.43} \\
    \bottomrule[1pt]
    \end{tabular}
	}
	\captionsetup{font={small}}
    \caption{\textbf{Adaptation results on DensePASS.}}
    \label{tab:per_class_densepass}
	\end{center}
\end{subtable}
\hspace{3pt}
\begin{subtable}[ht]{0.32\textwidth}
	\begin{center}
	\setlength{\tabcolsep}{1.0mm}
	\resizebox{\textwidth}{!}{
    \renewcommand{\arraystretch}{1.05}
    \begin{tabular}{ c| l | c | c}
    \toprule[1pt]
    &\textbf{Method} & \textbf{Input} & \textbf{mIoU(\%)} \\ \toprule[1pt]
    {\multirow{9}{*}{\rotatebox[origin=c]{90}{\textit{Supervised}}}} 
    &GaugeNet~\cite{gauge_equivariant} &RGB-D &39.4  \\
    &UGSCNN~\cite{spherical_unstructured_grids}& RGB-D  & 38.3 \\
    &HexRUNet~\cite{orientation} &RGB-D & 43.3 \\ \cline{2-4}
    &Tangent~\cite{tangent} (ResNet-101) & RGB  &45.6  \\
    &HoHoNet~\cite{hohonet} (ResNet-101) & RGB  & 52.0  \\
    &Trans4PASS (Small) & RGB & 52.1 \\
    &Trans4PASS (Small+MS) & RGB & \textbf{53.0} \\
    &\cellcolor{gray!15}Trans4PASS+ (Small+MS) & \cellcolor{gray!15}RGB & \cellcolor{gray!15}\textbf{54.1} \\
    \midrule
    {\multirow{4}{*}{\rotatebox[origin=c]{90}{\textit{UDA}}}} 
    &Trans4PASS (Source only) & RGB & 48.1 \\
    &Trans4PASS (MPA)   & RGB  & 50.8 \\
    &Trans4PASS (MPA+MS)   & RGB  & \textbf{51.2} \\
    &\cellcolor{gray!15}Trans4PASS+ (MPA+MS)   & \cellcolor{gray!15}RGB  & \cellcolor{gray!15}\textbf{52.3}\\
    \bottomrule[1pt]
    \end{tabular}
	}
	\captionsetup{font={small}}
    \caption{\textbf{Comparison on SPan} avg. of 3 folds.}
    \label{tab:supervised_s2d3d_pan}
	\end{center}
\end{subtable}
\hspace{3pt}
\begin{subtable}[ht]{0.31\textwidth}
	\begin{center}
    \scriptsize
	\setlength{\tabcolsep}{2mm}
	\resizebox{0.94\textwidth}{!}{
    \renewcommand{\arraystretch}{0.97}
    \begin{tabular}{ l | c | c }
    \toprule
    \textbf{Network} & \textbf{Method} &  \textbf{mIoU(\%)} \\
    \toprule
    PVT-Tiny & P2PDA & 39.66 \\
    DANet & P2PDA & 42.26 \\
    PVT-Small & P2PDA & 43.10 \\
    \midrule
    Trans4PASS (T) &-& 46.08 \\
    \rowcolor{gray!15} Trans4PASS+ (T) &-& 47.05 \\
    Trans4PASS (T) & MPA & 47.48 \\
    \rowcolor{gray!15} Trans4PASS+ (T) & MPA & {48.61}\\\midrule
    Trans4PASS (S) &-& 48.34 \\
    \rowcolor{gray!15} Trans4PASS+ (S) &-& 49.76 \\
    Trans4PASS (S) & MPA  & {52.15} \\
    \rowcolor{gray!15} Trans4PASS+ (S) & MPA  & \textbf{53.49} \\
    \bottomrule
    \end{tabular}
	}
	\captionsetup{font={small}}
	\caption{\textbf{Adaptation results on SPan}~$@$~fold-1.
	}
	\label{tab:per_class_s2d3d-pan}
	\end{center}
\end{subtable}

%% file: tables/table_mpa_s2r_p2p.tex
\begin{subtable}[ht]{\linewidth}
\setlength{\tabcolsep}{4pt}
\begin{center}
\resizebox{\linewidth}{!}{      
\renewcommand{\arraystretch}{0.92}
\begin{tabular}{l|c|c|ccccccccccccc}
\toprule[1pt]
\textbf{Network} & \textbf{Method} & \rotatebox{0}{mIoU} &  \rotatebox{0}{Road} &  \rotatebox{0}{S.walk} &  \rotatebox{0}{Build.} & \rotatebox{0}{Wall} &  \rotatebox{0}{Fence} &  \rotatebox{0}{Pole} & \rotatebox{0}{Tr. light} &  \rotatebox{0}{Tr. sign}&  \rotatebox{0}{Veget.} &  \rotatebox{0}{Terrain} &  \rotatebox{0}{Sky} & \rotatebox{0}{Person} &  \rotatebox{0}{Car}\\
\toprule[1pt]
\rowcolor{gray!15} \multicolumn{15}{l}{{(1) \textit{Outdoor} \textsc{Pin2Pan}: \textbf{{CS13}{$\rightarrow$}{DP13} }}} & \\ 
Trans4PASS+ (S) & Source-only & 51.48 & 76.45 & 40.52 & \textbf{86.16} & 28.70 & 43.77 & 26.93 & 15.75 & 16.71 & \textbf{79.91} & 32.48 & \textbf{93.76} & 49.21 & \textbf{78.87} \\
Trans4PASS+ (S) & MPA & \textbf{55.24} & \textbf{82.25} & \textbf{54.74} & 85.80 & \textbf{31.55} & \textbf{47.24} & \textbf{31.44} & \textbf{21.95} & \textbf{17.45} & 79.05 & \textbf{45.07} & 93.42 & \textbf{50.12} & 78.04 \\\midrule
\rowcolor{gray!15} \multicolumn{15}{l}{{(2) \textit{Outdoor} \textsc{Syn2Real}: \textbf{{SP13}{$\rightarrow$}{DP13}}}}  & \\ 
Trans4PASS+ (S) & Source-only & 43.83 & 70.26 & 42.36 & 80.22 & 12.88 & 20.55 & 19.32 & 17.01 & 03.44 & 71.43 & 31.28 & 90.14 & 44.64 & 66.21 \\
Trans4PASS+ (S) & MPA & \textbf{50.88} & \textbf{77.74} & \textbf{51.39} & \textbf{82.53} & \textbf{29.33} & \textbf{43.37} & \textbf{25.18} & \textbf{20.09} & \textbf{08.37} & \textbf{76.36} & \textbf{41.56} & \textbf{91.07} & \textbf{45.43} & \textbf{68.98} \\ 
\bottomrule
\end{tabular}
}
\end{center}
\vskip-1ex
\caption{{\textbf{Per-class results in CS13$\rightarrow$DP13 and SP13$\rightarrow$DP13, before and after MPA.}}}
\vskip-1ex
\label{tab:da_out_s2r_p2p}
\end{subtable}

\begin{subtable}[ht]{\linewidth}
\setlength{\tabcolsep}{10pt}
\begin{center}
\resizebox{\linewidth}{!}{      
\renewcommand{\arraystretch}{0.92}
\begin{tabular}{l|c|c|cccccccc}
\toprule[1pt]
\textbf{Network} & \textbf{Method} & \rotatebox{0}{mIoU}& \rotatebox{0}{Ceiling}& \rotatebox{0}{Chair}& \rotatebox{0}{Door}& \rotatebox{0}{Floor}& \rotatebox{0}{Sofa}& \rotatebox{0}{Table}& \rotatebox{0}{Wall}& \rotatebox{0}{Window} \\ \toprule[1pt]
\rowcolor{gray!15} \multicolumn{11}{l}{{(3) \textit{Indoor} \textsc{Pin2Pan}: \textbf{{SPin8}{$\rightarrow$}{SPan8}}}}  \\ 
Trans4PASS+ (S) & Source-only &63.73 & \textbf{90.63} & 62.30 & 24.79 & \textbf{92.62} & 35.73 & \textbf{73.16} & 78.74 & 51.78 \\
Trans4PASS+ (S) & MPA &\textbf{67.16} & 90.04 & \textbf{64.04} & \textbf{42.89} & 91.74 & \textbf{38.34} & 71.45 & \textbf{81.24} & \textbf{57.54} \\ \midrule
\rowcolor{gray!15} \multicolumn{11}{l}{{(4) \textit{Indoor} \textsc{Syn2Real}: \textbf{{S3D8}{$\rightarrow$}{SPan8}}}} \\ 
Trans4PASS+ (S) & Source-only & 51.75 & {85.37} & 49.90 & 09.63 & {89.75} & \textbf{21.40} & {32.10} & \textbf{71.49} & 54.34 \\
Trans4PASS+ (S) & MPA &\textbf{52.73} & \textbf{85.86} & \textbf{52.89} & \textbf{15.30} & \textbf{90.74} & {07.83} & \textbf{37.78} & {71.15} & \textbf{60.24} \\ \bottomrule
\end{tabular}
}
\end{center}
\vskip-1ex
\caption{{\textbf{Per-class results in SPin8$\rightarrow$SPan8 and S3D8$\rightarrow$SPan8, before and after MPA.}}}
\vskip-1ex
\label{tab:da_in_s2r_p2p}
\end{subtable}

%% file: Tex_content/conclusion.tex
In this paper, we propose a universal framework with two variants of the \emph{Transformer for PAnoramic Semantic Segmentation (Trans4PASS)} architecture to revitalize $360^\circ$ scene understanding. The \emph{Deformable Patch Embedding (DPE)} and the \emph{Deformable MLP (DMLP)} modules empower Trans4PASS with distortion awareness. A \emph{Mutual Prototypical Adaptation (MPA)} strategy is introduced for transferring semantic information from the label-rich source domain to the label-scarce target domain, by combining source labels and target pseudo-label for feature alignment in feature and output space. A new dataset, termed \emph{SynPASS}, is created. It enables the supervised training of panoramic segmentation models, and it further provides an alternative Synthetic-to-Real (\textsc{Syn2Real}) domain adaptation paradigm, which is compared to the Pinhole-to-Panoramic (\textsc{Pin2Pan}) adaptation scenario. The framework obtains state-of-the-art accuracy on four competitive domain adaptive panoramic semantic segmentation benchmarks.
In the future, we will explore the combination of cubemap and equirectangular projections, along with the fusion of LiDAR data and panoramic images. Furthermore, it would be interesting to combine two sources, such as pinhole and synthetic datasets, and investigate multi-source domain adaptive panoramic segmentation.

%

%% file: Tex_content/appendix.tex
\clearpage
\appendices
\counterwithin{figure}{section}
\counterwithin{table}{section}

\section{More Quantitative Results}
\label{appendix:more_quantitative_results}

\subsection{Analysis of hyper-parameters}
As the spatial correspondence problem indicated in~\cite{dai2017deformable}, if the deformable convolution is added to the shallow or middle layers, the spatial structures are susceptible to fluctuation~\cite{restricted}. To solve this issue, the regional restriction of learned offsets is used to stabilize the training of our early-stage and four-stage Deformable Patch Embedding (DPE) module. Table~\ref{table:effect_r} shows that $r{=}4$ has a better result, and we set it as default in our experiments. 

\begin{table}[h]
%
\footnotesize
\setlength\tabcolsep{10.0pt}
\centering
\caption{\small \textbf{Effect of regional restriction ($r$)} on DensePASS.}
\vskip -2ex
\resizebox{\columnwidth}{!}{
\begin{tabular}{r|rrrrr}
    \toprule
     & None & r=1 & r=2 & r=4 & r=8 \\ \midrule[0.5pt]
     mIoU(\%) & 45.74 & 44.51& 45.59& \textbf{45.89}& 45.57\\
    \bottomrule
\end{tabular}}
\label{table:effect_r}
\vskip-2ex
\end{table}

\begin{figure}[h]
	\centering
    \includegraphics[width=0.99\columnwidth]{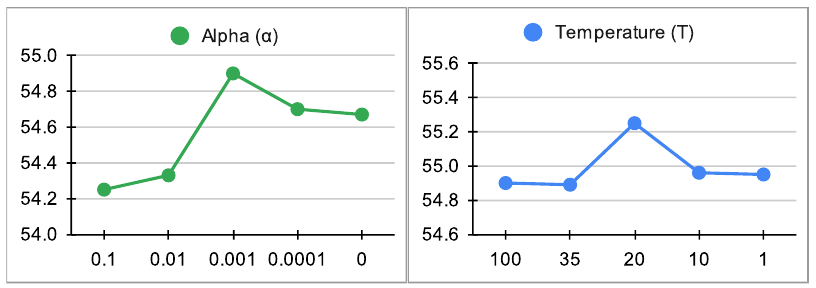}
    \begin{minipage}[t]{.5\columnwidth}
    \centering
    \vskip -3ex
    \subcaption{\small mIoU(\%) {--} $\alpha$}\label{fig:hyper_alpha}
    \end{minipage}%
    \begin{minipage}[t]{.5\columnwidth}
    \centering
    \vskip -3ex
    \subcaption{\small mIoU(\%) {--} $\mathcal{T}$}\label{fig:hyper_temperature}
    \end{minipage}%
    \vskip -2ex
	\caption{ \small \textbf{Analysis of hyper-parameters} on DensePASS. } 
	\label{fig:hyper_params}
\end{figure}
We analyze the weight $\alpha$ and the temperature $\mathcal{T}$ as shown in Fig.~\ref{fig:hyper_alpha} and Fig.~\ref{fig:hyper_temperature}. The \emph{Mutual Prototypical Adaptation (MPA)} loss and the source- and target segmentation losses are combined by the weight $\alpha$. As $\alpha$ decreases from $0.1$ to $0$, we set the temperature $\mathcal{T}{=}35$ in the MPA loss and evaluate the mIoU($\%$) results on the DensePASS dataset~\cite{densepass}. If $\alpha{=}0$, the final loss is equivalent to that of the SSL-based method, \ie, the MPA loss is excluded. When $\alpha{=}0.001$ for combining both, MPA and SSL, Trans4PASS obtains a better result. 
We further investigate the effect of the temperature $\mathcal{T}$ in the MPA loss. As shown in Fig.~\ref{fig:hyper_temperature}, the performance is not sensitive to the distillation temperature, which illustrates the robustness of our MPA method. Nevertheless, we found that MPA performs better when the temperature is lower, so $\mathcal{T}{=}20$ is set as default.

\subsection{Computational complexity}
We report the complexity of Deformable Patch Embedding~(DPE) and two Deformable MLP~(DMLP) modules on DensePASS in Table~\ref{table:complex_dpe_dmlp}. The comparison indicates that our methods have better results with the same order of complexity.

\begin{table}[h]
\footnotesize
\setlength\tabcolsep{1.0pt}
\centering
\caption{\small {\textbf{Computational complexity}. GFLOPs~@$512{\times}512$.}}
\vskip -2ex
\resizebox{\columnwidth}{!}{
\begin{tabular}{@{}c|c>{\columncolor[gray]{0.9}}c|cccc>{\columncolor[gray]{0.9}}c}
    \toprule
     & \textbf{PE}~\cite{segformer} & \textbf{DPE} & \textbf{DPT}~\cite{chen2021dpt} & \textbf{CycleMLP}~\cite{cyclemlp} & \textbf{ASMLP}~\cite{asmlp} & \textbf{DMLP} & \textbf{DMLPv2} \\ \midrule[0.5pt]
     GFLOPs & 0.16 & 0.36 & 7.65 & 1.25 & 4.83 & 3.45 & 3.13\\
     \#Params(M) & 0.01 & 0.02 & 2.90 & 0.45 & 1.04 & 0.79 & 0.80 \\\midrule[0.5pt]
     mIoU(\%) & 45.14  & \textbf{49.94} & 36.50 & 40.16 & 42.05 & {45.89} & \textbf{49.94}\\
    \bottomrule
\end{tabular}}
\label{table:complex_dpe_dmlp}
\vskip-2ex
\end{table}

\section{More Qualitative Results}
\label{appendix:apd_more_qualitative_results}

\subsection{Panoramic semantic segmentation} 
\label{appendix:apd_pano_seg_vis}
\begin{figure*}[!t]
%
	\begin{subfigure}[b]{1.0\textwidth}   
	\centering 
	\includegraphics[width=1.0\textwidth]{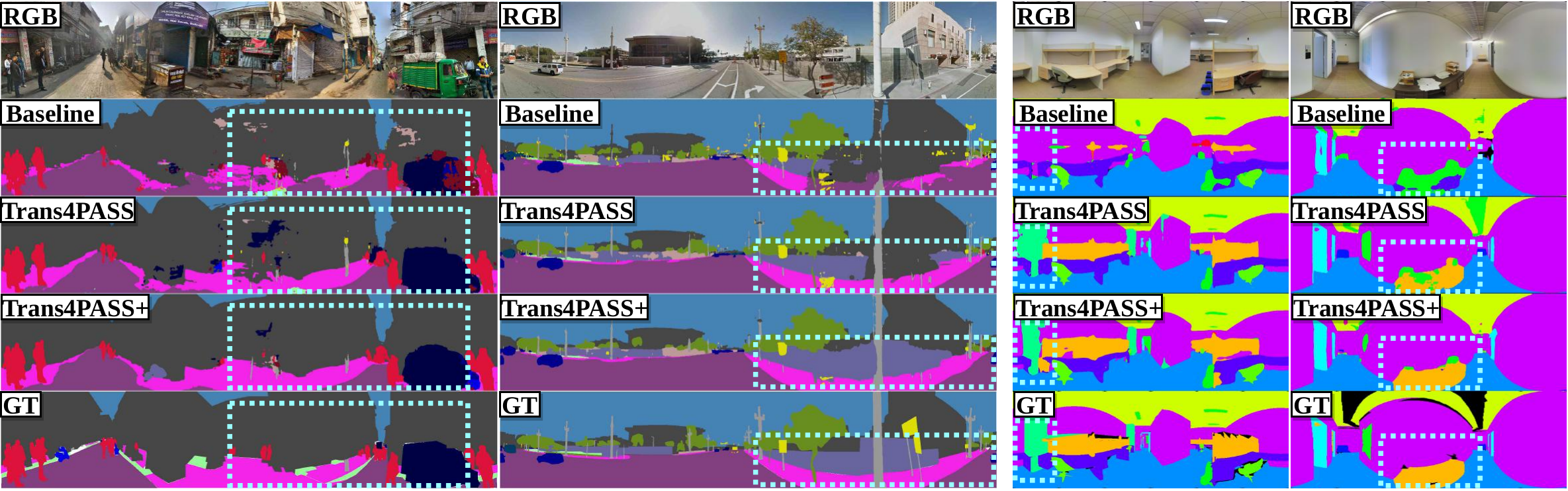}
    \begin{minipage}[t]{.65\textwidth}
    \centering
    \vskip -3.5ex
    \subcaption{\small Segmentation outdoors}\label{fig:vis_apd_outdoor}
    \end{minipage}%
    \begin{minipage}[t]{.35\textwidth}
    \centering
    \vskip -3.5ex
    \subcaption{\small Segmentation indoors}\label{fig:vis_apd_indoor}
    \end{minipage}%
    \end{subfigure}
    \vskip -2ex
	\caption{\small \textbf{More panoramic semantic segmentation visualizations.} {Zoom in for a better view.}} 
	\label{fig:vis_apd}
\end{figure*}

{To verify the proposed model, more qualitative comparisons based on the DensePASS dataset are displayed in Fig.~\ref{fig:vis_apd}.} Specifically, Trans4PASS models can better segment deformed foreground objects, such as \emph{trucks} in Fig.~\ref{fig:vis_apd_outdoor}. Apart from the foreground object, Trans4PASS models yield high-quality segmentation results in the distorted background categories, \eg, \emph{fence} and \emph{sidewalk}.

For indoor scenarios, more qualitative comparisons are shown in Fig.~\ref{fig:vis_apd_indoor}, which are from the fold-$1$ Stanford2D3D-Panoramic dataset~\cite{stanford2d3d}. Our models produce better segmentation results in those categories, such as \emph{columns} and \emph{tables}, while the baseline model can hardly identify these deformed objects. 

\begin{figure}[!t]
	\begin{subfigure}[b]{1.0\columnwidth}   
		\centering 
		\includegraphics[width=1.0\columnwidth]{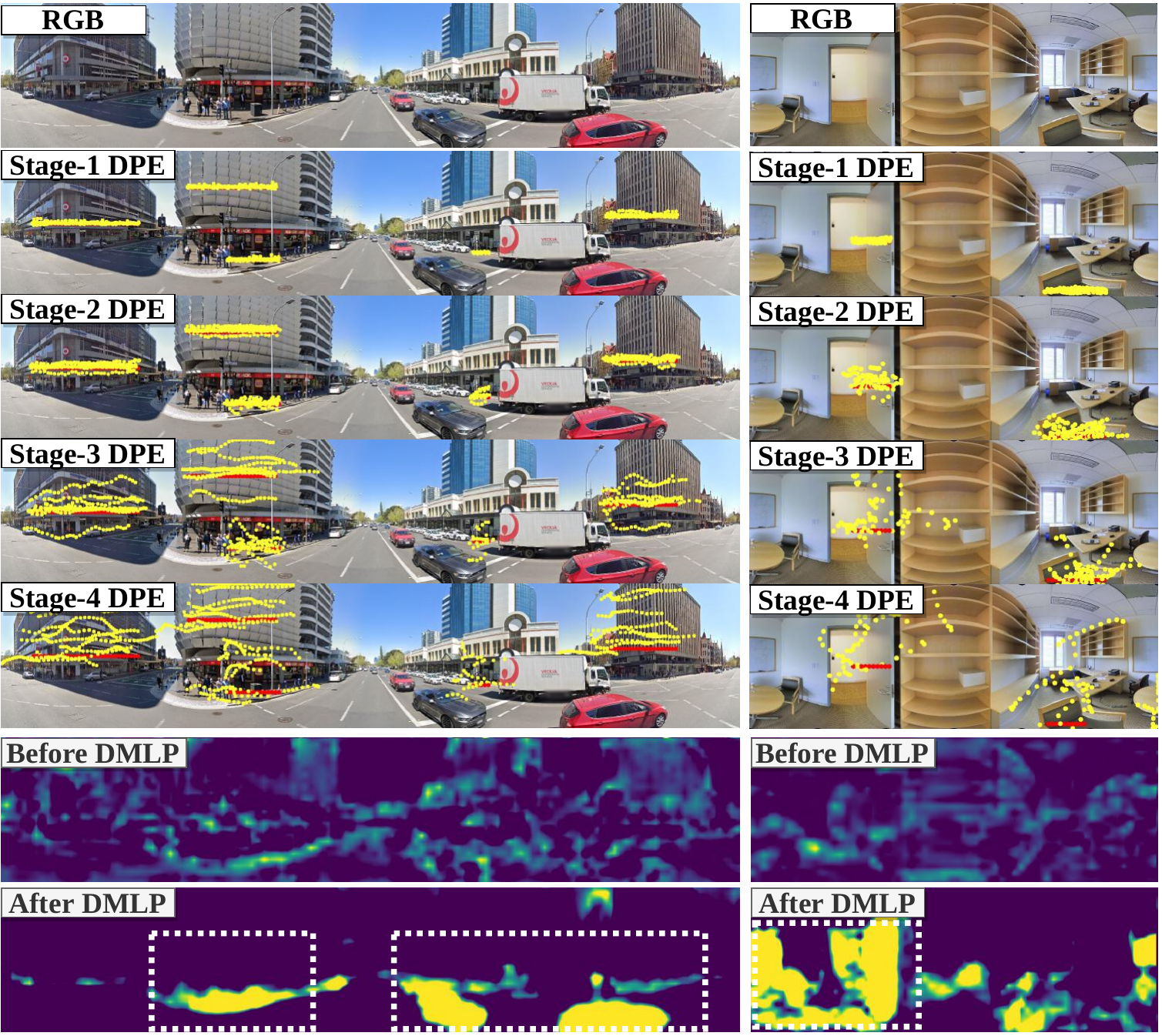}
	    \begin{minipage}[t]{.65\columnwidth}
        \centering
        \vskip -3.5ex
        \subcaption{\small outdoors}\label{fig:vis_dpe_dmlp_outdoor}
        \end{minipage}%
        \begin{minipage}[t]{.35\columnwidth}
        \centering
        \vskip -3.5ex
        \subcaption{\small indoors}\label{fig:vis_dpe_dmlp_indoor}
        \end{minipage}%
	\end{subfigure}
    \vskip -2ex
	\caption{\small \textbf{DPE and DMLP visualizations.} The \textcolor{yellow}{$\bullet$} dots in four stages are sampling points shifted by learned offsets \textit{w.r.t.} the \textcolor{red}{$\bullet$} patch center of DPE (from decoder). The bottom two rows show the $\text{\#}75$ channel maps of stage-$3$ before and after DMLP. {Zoom in for a better view.}} 
	\label{fig:vis_dpe_dmlp}
\vskip -3ex
\end{figure}

\subsection{DPE and DMLP visualizations}
To investigate the effectiveness of two distortion-aware designs, the visualizations of DPE and DMLP (v1) are shown in Fig.~\ref{fig:vis_dpe_dmlp}. The RGB images and DPE from four stages of Trans4PASS are visualized in the top five rows in Fig.~\ref{fig:vis_dpe_dmlp}, where the red dots are the centers of the $s{\times}s$ patch sequence and the $s^{2}$ yellow dots are the learned offsets from DPE. The offsets result that each pixel is adaptive to distorted objects and space, such as the deformed \emph{building} and \emph{sidewalk} in Stage-$4$ DPE in the outdoor case (Fig.~\ref{fig:vis_dpe_dmlp}-(a)) and the \textit{chairs} in the indoor case (Fig.~\ref{fig:vis_dpe_dmlp}-(b)). Furthermore, two feature map pairs from the $75^{th}$ channel before and after DMLP are displayed in the bottom two rows in Fig.~\ref{fig:vis_dpe_dmlp}. Compared to the feature maps before DMLP, the feature maps are enhanced by the DMLP-based token mixer and present semantically recognizable responses, \eg on regions of distorted \emph{sidewalks} or objects of deformed \emph{cars}.

\begin{table*}[!t]
    \caption{\textbf{Per-class results} on the \emph{test} set of the SynPASS benchmark.}
	\vskip-3ex
    \renewcommand\arraystretch{1.0}
	\begin{center}
\input{tables/table_apd_synpass.tex}
	\end{center}
	\label{tab:apd_synpass_per_class}
	\vskip-2ex
\end{table*}

\subsection{Segmentation on the SynPASS benchmark}
Table.~\ref{tab:apd_synpass_per_class} presents per-class results on the SynPASS benchmark. The small Trans4PASS+ model obtains $39.16\%$ mIoU and sufficient improvements, as compared to the CNN-based HRNet model ($+5.07\%$) and the SegFormer model ($+1.92\%$). Besides, our Trans4PASS models achieve top scores on $17$ of $22$ classes. However, there is still a lot to be excavated on the SynPASS benchmark, such as the \emph{wall}, \emph{ground}, \emph{bridge}, and \emph{dynamic} categories, which are challenging cases in the synthetic panoramic images.

\begin{figure*}[!t]
    \centering
	\includegraphics[width=0.8\textwidth]{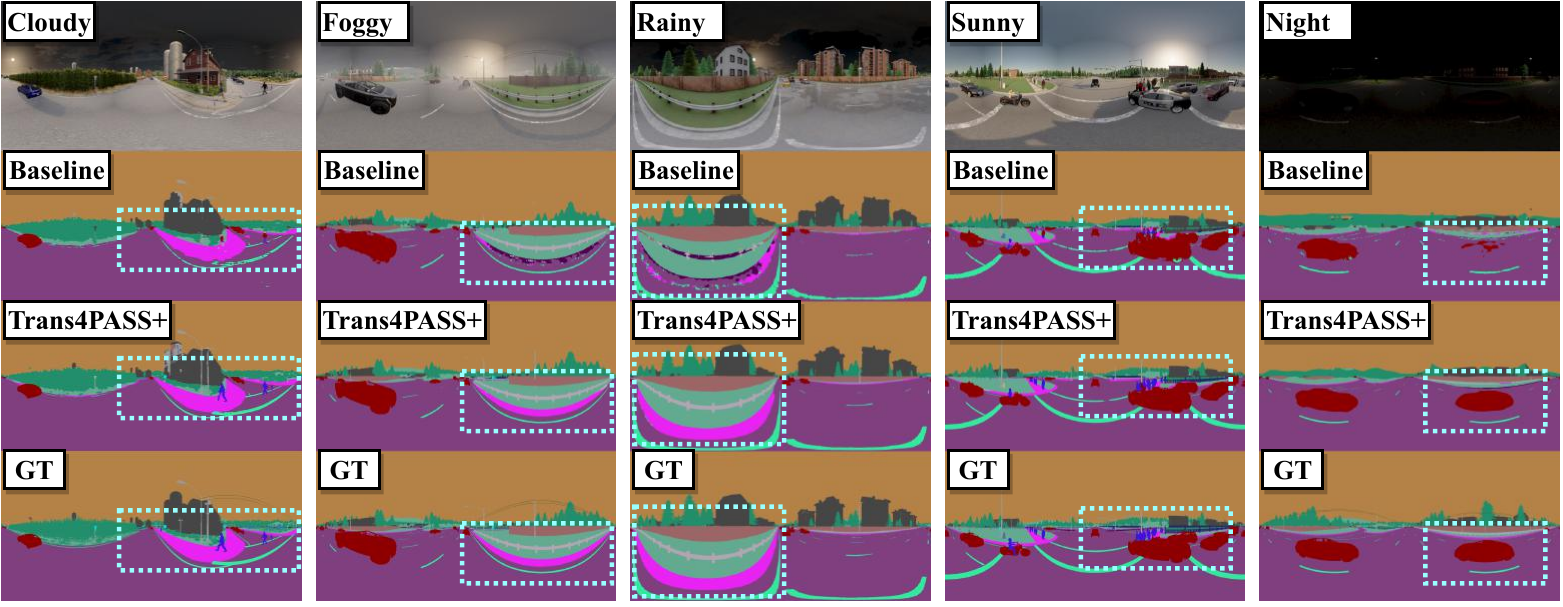}
    \vskip -2ex
	\caption{\small \textbf{SynPASS segmentation visualizations}. {Zoom in for a better view.}} 
	\label{fig:vis_synpass_apd}
\vskip -2ex
\end{figure*}

A montage of panoramic semantic segmentation results generated from the validation set of the SynPASS dataset is presented in Fig.~\ref{fig:vis_synpass_apd}. %
Compared with the baseline PVTv2 model~\cite{pvtv2}, our Trans4PASS+ model is more robust against adverse situations and obtains more accurate segmentation results, such as the \emph{pedestrian} in cloudy and sunny scenes, the \emph{sidewalk} in the foggy and rainy scenes, and the \emph{vehicles} in the night scenes.

\subsection{\textsc{Pin2Pan} vs. \textsc{Syn2Real} visualization}
\label{appendix:apd_pin2pan_vs_syn2real_vis}
\begin{figure*}[!t]
    \centering
	\includegraphics[width=0.8\textwidth]{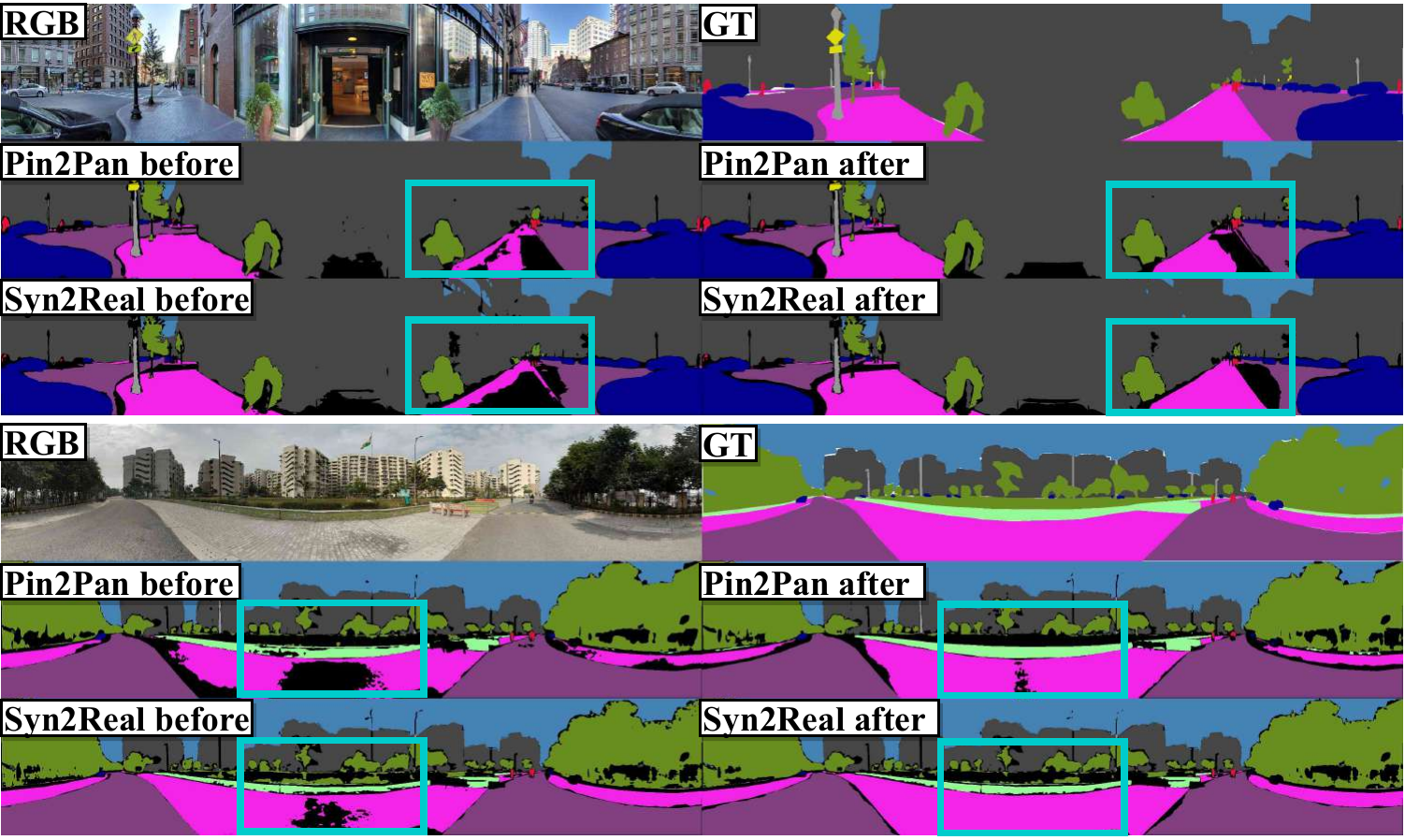}
    \vskip -2ex
	\caption{\small \textbf{More \textsc{Pin2Pan} vs. \textsc{Syn2Real} visualizations} before and after MPA, respectively. {Zoom in for a better view.}} 
	\label{fig:vis_p2p_s2r_apd}
\vskip -3ex
\end{figure*}

For a more comprehensive analysis of the two different adaptation paradigms, additional visualization samples are shown in Fig.~\ref{fig:vis_p2p_s2r_apd}. In the first case, before MPA, there is not a significant deformation in the target-domain \emph{sidewalk} highlighted by the blue box, thus, the pinhole-trained model obtains more accurate segmentation results than the synthetic-trained model.
{That means if no distortion appears, the pinhole-trained model benefits more from the same realistic scene appearance as the target domain and can perform better than the synthetic-trained model.}
{However, in the second case before MPA, the situation is reversed due to the existence of distortion in the highlighted \emph{sidewalk} from the target domain, which appears in an uncommon position compared to that in the pinhole domain.} At this point, the synthetic-source trained model benefits more from the similar shape and position prior as in the target domain \emph{sidewalk}. Nonetheless, after our MPA, both paradigms obtain more complete and accurate segmentation results. This verifies the effectiveness of our proposed mutual prototypical adaptation strategy, which jointly uses ground-truth labels from the source and pseudo-labels from the target, and drives the domain alignment on the feature and output spaces.

\subsection{Failure Case Analysis}
\label{appendix:failure_case_analysis}
\begin{figure*}[!t]
    \centering
	\includegraphics[width=0.8\textwidth]{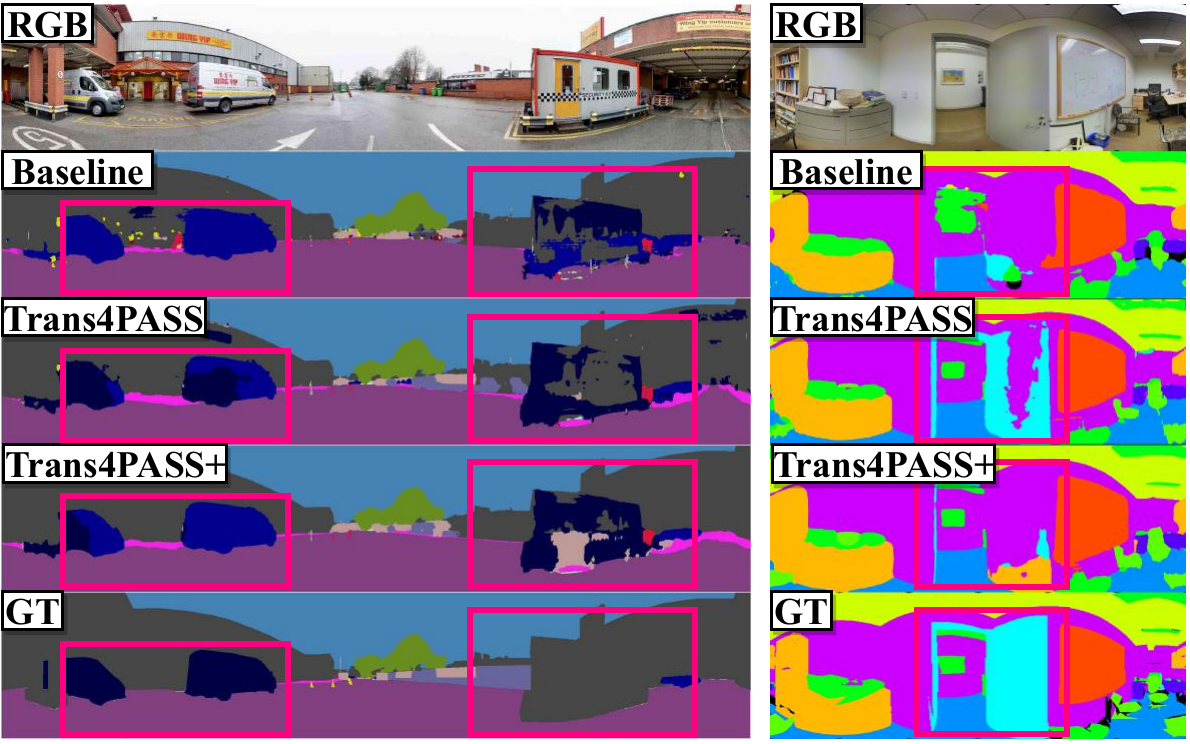}
    \vskip -2ex
	\caption{\small \textbf{Failure case visualizations from indoor and outdoor scenarios.}} 
	\label{fig:vis_fail}
\vskip -3ex
\end{figure*}

{Some failure cases of panoramic semantic segmentation are presented in Fig.~\ref{fig:vis_fail}.
Some erroneous segmentation samples from the three models are presented in Fig.~\ref{fig:vis_fail}. In the outdoor scene, while the PVTv2 baseline~\cite{pvtv2} recognizes the \emph{truck} as a \emph{car}, the Trans4PASS model can only segment a part of the \emph{truck}. All three models have difficulty segmenting the \emph{building} that looks similar to a \emph{truck}. In the indoor scene, the baseline and the Trans4PASS+ model fail to differentiate between the distorted \emph{door} and \emph{wall}, as both are similar in appearance and shape in this case. This issue can potentially be addressed by using complementary panoramic depth information to obtain discriminative features.}

%% file: tables/table_apd_synpass.tex
\vskip-3ex
\setlength{\tabcolsep}{4pt}
    \begin{center}
	\resizebox{\textwidth}{!}{      
    \renewcommand{\arraystretch}{1.0}
	\begin{tabular}{ @{}l | c | c c c c c c c c c c c c c c c c c c c c c c@{}}
        \toprule[1pt]
        Method & \rotatebox{90}{mIoU} &\rotatebox{90}{Building}&\rotatebox{90}{Fence}&\rotatebox{90}{Other}&\rotatebox{90}{Pedestrian}&\rotatebox{90}{Pole}&\rotatebox{90}{RoadLine}&\rotatebox{90}{Road}&\rotatebox{90}{SideWalk}&\rotatebox{90}{Vegetation}&\rotatebox{90}{Vehicles}&\rotatebox{90}{Wall}&\rotatebox{90}{TrafficSign}&\rotatebox{90}{Sky}&\rotatebox{90}{Ground}&\rotatebox{90}{Bridge}&\rotatebox{90}{RailTrack}&\rotatebox{90}{GroundRail}&\rotatebox{90}{TrafficLight}&\rotatebox{90}{Static}&\rotatebox{90}{Dynamic}&\rotatebox{90}{Water}&\rotatebox{90}{Terrain}\\
        \toprule[1pt]
        Fast-SCNN (Fast-SCNN) &21.30&64.14&10.22&00.00&00.08&07.77&26.68&80.63&33.87&60.93&32.91&00.14&00.00&89.61&\textbf{01.37}&00.01&03.80&20.81&00.00&01.02&00.00&00.01&34.68\\
        DeepLabv3+ (MobileNetv2) &29.66&75.14&11.50&00.00&25.04&11.05&39.90&89.19&43.73&62.44&62.41&01.59&00.00&91.86&00.54&00.01&26.13&43.72&08.30&12.67&01.06&05.03&41.21\\
        HRNet (W18Small) & 34.09& 75.94&28.76&00.00&29.59&23.74&59.21&91.68&52.63&63.94&64.63&00.70&00.00&93.02&00.81&00.01&\textbf{27.42}&65.54&08.08&16.32&01.10&03.80&42.99\\ \midrule
        PVT (Tiny) &  32.37&74.83&19.94&00.24&21.82&13.15&62.59&93.14&49.09&67.27&46.44&{01.69}&09.63&96.09&00.18&\textbf{02.64}&08.81&61.11&14.09&12.04&00.99&05.05&51.33\\
        PVT (Small) & 32.68&78.02&27.12&\textbf{00.27}&23.48&16.51&59.81&92.87&50.21&66.22&43.50&01.12&08.67&96.34&00.44&00.15&02.82&63.88&13.78&15.15&01.58&08.78&48.29 \\
        SegFormer (B1) & 37.36&78.24&20.59&00.00&38.28&21.09&68.72&94.50&59.72&68.43&67.51&00.83&09.86&96.08&00.56&01.38&20.79&69.59&23.38&19.91&01.38&08.97&52.07 \\
        SegFormer (B2) & 37.24&79.25&23.58&00.00&40.01&20.14&65.28&92.80&46.92&68.64&77.45&01.42&15.00&96.33&00.57&00.58&02.68&67.60&25.86&20.80&01.99&\textbf{20.92}&51.53\\
        Trans4PASS (Tiny) & 38.53&79.17&\textbf{28.18}&00.13&36.04&23.69&69.16&95.51&{61.71}&69.77&71.12&01.53&16.98&96.50&00.56&01.60&15.22&70.48&{26.03}&23.11&02.08&09.24&49.77\\
        Trans4PASS (Small) & 38.57&80.02&24.56&00.07&41.49&25.23&72.00&95.89&59.88&69.07&77.08&01.04&13.72&96.69&00.67&00.73&05.60&\textbf{72.56}&25.93&22.45&02.78&08.34&{52.65}\\
        Trans4PASS+ (Tiny) & 39.42&79.63& 24.45& 00.21& 44.23& \textbf{26.71}& 70.32& 95.86& 61.80& 69.25& 78.85& 01.09& 13.81& 97.12& 00.91& 03.48& 19.32& 72.44& 21.08& 25.56& 02.67& 05.03& 53.20 \\
        \rowcolor{gray!15} Trans4PASS+ (Small) & \textbf{40.72}&\textbf{80.91}& 20.78& 00.23& \textbf{45.36}& 24.08& \textbf{72.51}& \textbf{96.79}& \textbf{67.15}& \textbf{70.46}& \textbf{81.39}& \textbf{04.28}& \textbf{26.19}& \textbf{97.21}& 01.24& 01.74& 16.56& 67.08& \textbf{28.64}& \textbf{23.68}& \textbf{03.35}& 08.48& \textbf{57.57} \\

        \bottomrule[1pt]
        \end{tabular}
    }
    \end{center}

%% file: main.bbl
\begin{thebibliography}{100}
\providecommand{\url}[1]{#1}
\csname url@samestyle\endcsname
\providecommand{\newblock}{\relax}
\providecommand{\bibinfo}[2]{#2}
\providecommand{\BIBentrySTDinterwordspacing}{\spaceskip=0pt\relax}
\providecommand{\BIBentryALTinterwordstretchfactor}{4}
\providecommand{\BIBentryALTinterwordspacing}{\spaceskip=\fontdimen2\font plus
\BIBentryALTinterwordstretchfactor\fontdimen3\font minus \fontdimen4\font\relax}
\providecommand{\BIBforeignlanguage}[2]{{%
\expandafter\ifx\csname l@#1\endcsname\relax
\typeout{** WARNING: IEEEtran.bst: No hyphenation pattern has been}%
\typeout{** loaded for the language `#1'. Using the pattern for}%
\typeout{** the default language instead.}%
\else
\language=\csname l@#1\endcsname
\fi
#2}}
\providecommand{\BIBdecl}{\relax}
\BIBdecl

\bibitem{wildpass}
K.~Yang, X.~Hu, and R.~Stiefelhagen, ``Is context-aware {CNN} ready for the surroundings? {P}anoramic semantic segmentation in the wild,'' \emph{TIP}, vol.~30, pp. 1866--1881, 2021.

\bibitem{Garanderie_2018_ECCV}
G.~P. de~La~Garanderie, A.~A. Abarghouei, and T.~P. Breckon, ``Eliminating the blind spot: Adapting {3D} object detection and monocular depth estimation to 360$^\circ$ panoramic imagery,'' in \emph{ECCV}, 2018.

\bibitem{densepass}
C.~Ma, J.~Zhang, K.~Yang, A.~Roitberg, and R.~Stiefelhagen, ``{DensePASS:} {Dense} panoramic semantic segmentation via unsupervised domain adaptation with attention-augmented context exchange,'' in \emph{ITSC}, 2021.

\bibitem{gao2022review}
S.~Gao, K.~Yang, H.~Shi, K.~Wang, and J.~Bai, ``Review on panoramic imaging and its applications in scene understanding,'' \emph{TIM}, vol.~71, pp. 1--34, 2022.

\bibitem{xu2018predicting_head_movement}
M.~Xu, Y.~Song, J.~Wang, M.~Qiao, L.~Huo, and Z.~Wang, ``Predicting head movement in panoramic video: A deep reinforcement learning approach,'' \emph{TPAMI}, vol.~41, no.~11, pp. 2693--2708, 2019.

\bibitem{xu2021spherical}
Y.~Xu, Z.~Zhang, and S.~Gao, ``Spherical {DNNs} and their applications in 360$^\circ$ images and videos,'' \emph{TPAMI}, vol.~44, no.~10, pp. 7235--7252, 2022.

\bibitem{ai2022deep_omnidirectional_survey}
H.~Ai, Z.~Cao, J.~Zhu, H.~Bai, Y.~Chen, and L.~Wang, ``Deep learning for omnidirectional vision: A survey and new perspectives,'' \emph{arXiv preprint arXiv:2205.10468}, 2022.

\bibitem{hohonet}
C.~Sun, M.~Sun, and H.-T. Chen, ``{HoHoNet:} 360 indoor holistic understanding with latent horizontal features,'' in \emph{CVPR}, 2021.

\bibitem{omnirange}
K.~Yang, J.~Zhang, S.~Rei{\ss}, X.~Hu, and R.~Stiefelhagen, ``Capturing omni-range context for omnidirectional segmentation,'' in \emph{CVPR}, 2021.

\bibitem{pass}
K.~Yang, X.~Hu, L.~M. Bergasa, E.~Romera, and K.~Wang, ``{PASS:} {Panoramic} annular semantic segmentation,'' \emph{T-ITS}, vol.~21, no.~10, pp. 4171--4185, 2020.

\bibitem{zhang2022trans4pass}
J.~Zhang, K.~Yang, C.~Ma, S.~Rei{\ss}, K.~Peng, and R.~Stiefelhagen, ``Bending reality: Distortion-aware transformers for adapting to panoramic semantic segmentation,'' in \emph{CVPR}, 2022.

\bibitem{pvt}
W.~Wang, E.~Xie, X.~Li, D.~Fan, K.~Song, D.~Liang, T.~Lu, P.~Luo, and L.~Shao, ``Pyramid vision transformer: A versatile backbone for dense prediction without convolutions,'' in \emph{ICCV}, 2021.

\bibitem{cityscapes}
M.~Cordts, M.~Omran, S.~Ramos, T.~Rehfeld, M.~Enzweiler, R.~Benenson, U.~Franke, S.~Roth, and B.~Schiele, ``The cityscapes dataset for semantic urban scene understanding,'' in \emph{CVPR}, 2016.

\bibitem{setr}
S.~Zheng, J.~Lu, H.~Zhao, X.~Zhu, Z.~Luo, Y.~Wang, Y.~Fu, J.~Feng, T.~Xiang, P.~H.~S. Torr, and L.~Zhang, ``Rethinking semantic segmentation from a sequence-to-sequence perspective with transformers,'' in \emph{CVPR}, 2021.

\bibitem{clan}
Y.~Luo, L.~Zheng, T.~Guan, J.~Yu, and Y.~Yang, ``Taking a closer look at domain shift: Category-level adversaries for semantics consistent domain adaptation,'' in \emph{CVPR}, 2019.

\bibitem{crst}
Y.~Zou, Z.~Yu, X.~Liu, B.~V. K.~V. Kumar, and J.~Wang, ``Confidence regularized self-training,'' in \emph{ICCV}, 2019.

\bibitem{kirillov2023SAM}
A.~Kirillov, E.~Mintun, N.~Ravi, H.~Mao, C.~Rolland, L.~Gustafson, T.~Xiao, S.~Whitehead, A.~C. Berg, W.~Lo, P.~Doll{\'{a}}r, and R.~B. Girshick, ``Segment anything,'' \emph{arXiv preprint arXiv:2304.02643}, 2023.

\bibitem{stanford2d3d}
I.~Armeni, S.~Sax, A.~R. Zamir, and S.~Savarese, ``Joint {2D-3D-semantic} data for indoor scene understanding,'' \emph{arXiv preprint arXiv:1702.01105}, 2017.

\bibitem{structured3d}
J.~Zheng, J.~Zhang, J.~Li, R.~Tang, S.~Gao, and Z.~Zhou, ``{Structured3D:} {A} large photo-realistic dataset for structured {3D} modeling,'' in \emph{ECCV}, 2020.

\bibitem{p2pda_trans}
J.~Zhang, C.~Ma, K.~Yang, A.~Roitberg, K.~Peng, and R.~Stiefelhagen, ``Transfer beyond the field of view: Dense panoramic semantic segmentation via unsupervised domain adaptation,'' \emph{T-ITS}, vol.~23, no.~7, pp. 9478--9491, 2022.

\bibitem{yu2021metaformer}
W.~Yu, M.~Luo, P.~Zhou, C.~Si, Y.~Zhou, X.~Wang, J.~Feng, and S.~Yan, ``{MetaFormer} is actually what you need for vision,'' in \emph{CVPR}, 2022.

\bibitem{cyclemlp}
S.~Chen, E.~Xie, C.~Ge, D.~Liang, and P.~Luo, ``{CycleMLP:} {A} {MLP-like} architecture for dense prediction,'' in \emph{ICLR}, 2022.

\bibitem{asmlp}
D.~Lian, Z.~Yu, X.~Sun, and S.~Gao, ``{AS-MLP:} {An} axial shifted {MLP} architecture for vision,'' in \emph{ICLR}, 2022.

\bibitem{zhou2022fan}
D.~Zhou, Z.~Yu, E.~Xie, C.~Xiao, A.~Anandkumar, J.~Feng, and J.~M. Alvarez, ``Understanding the robustness in vision transformers,'' in \emph{ICML}, 2022.

\bibitem{chen2021dpt}
Z.~Chen, Y.~Zhu, C.~Zhao, G.~Hu, W.~Zeng, J.~Wang, and M.~Tang, ``{DPT:} {Deformable} patch-based transformer for visual recognition,'' in \emph{MM}, 2021.

\bibitem{hoyer2021daformer}
L.~Hoyer, D.~Dai, and L.~Van~Gool, ``{DAFormer:} {Improving} network architectures and training strategies for domain-adaptive semantic segmentation,'' in \emph{CVPR}, 2022.

\bibitem{fcn}
J.~Long, E.~Shelhamer, and T.~Darrell, ``Fully convolutional networks for semantic segmentation,'' in \emph{CVPR}, 2015.

\bibitem{segnet}
V.~Badrinarayanan, A.~Kendall, and R.~Cipolla, ``{SegNet:} {A} deep convolutional encoder-decoder architecture for image segmentation,'' \emph{TPAMI}, vol.~39, no.~12, pp. 2481--2495, 2017.

\bibitem{deeplabv3+}
L.-C. Chen, Y.~Zhu, G.~Papandreou, F.~Schroff, and H.~Adam, ``Encoder-decoder with atrous separable convolution for semantic image segmentation,'' in \emph{ECCV}, 2018.

\bibitem{refinenet}
G.~Lin, A.~Milan, C.~Shen, and I.~Reid, ``{RefineNet:} {Multi-path} refinement networks for high-resolution semantic segmentation,'' in \emph{CVPR}, 2017.

\bibitem{hrnet}
J.~Wang, K.~Sun, T.~Cheng, B.~Jiang, C.~Deng, Y.~Zhao, D.~Liu, Y.~Mu, M.~Tan, X.~Wang, W.~Liu, and B.~Xiao, ``Deep high-resolution representation learning for visual recognition,'' \emph{TPAMI}, vol.~43, no.~10, pp. 3349--3364, 2021.

\bibitem{deeplabv2}
L.-C. Chen, G.~Papandreou, I.~Kokkinos, K.~Murphy, and A.~L. Yuille, ``{DeepLab:} {Semantic} image segmentation with deep convolutional nets, atrous convolution, and fully connected {CRFs},'' \emph{TPAMI}, vol.~40, no.~4, pp. 834--848, 2018.

\bibitem{pspnet}
H.~Zhao, J.~Shi, X.~Qi, X.~Wang, and J.~Jia, ``Pyramid scene parsing network,'' in \emph{CVPR}, 2017.

\bibitem{hou2020strip}
Q.~Hou, L.~Zhang, M.-M. Cheng, and J.~Feng, ``Strip pooling: Rethinking spatial pooling for scene parsing,'' in \emph{CVPR}, 2020.

\bibitem{context_encoding}
H.~Zhang, K.~J. Dana, J.~Shi, Z.~Zhang, X.~Wang, A.~Tyagi, and A.~Agrawal, ``Context encoding for semantic segmentation,'' in \emph{CVPR}, 2018.

\bibitem{context_prior}
C.~Yu, J.~Wang, C.~Gao, G.~Yu, C.~Shen, and N.~Sang, ``Context prior for scene segmentation,'' in \emph{CVPR}, 2020.

\bibitem{jin2021mining}
Z.~Jin, T.~Gong, D.~Yu, Q.~Chu, J.~Wang, C.~Wang, and J.~Shao, ``Mining contextual information beyond image for semantic segmentation,'' in \emph{ICCV}, 2021.

\bibitem{nonlocal}
X.~Wang, R.~Girshick, A.~Gupta, and K.~He, ``Non-local neural networks,'' in \emph{CVPR}, 2018.

\bibitem{attention}
A.~Vaswani, N.~Shazeer, N.~Parmar, J.~Uszkoreit, L.~Jones, A.~N. Gomez, L.~Kaiser, and I.~Polosukhin, ``Attention is all you need,'' in \emph{NeurIPS}, 2017.

\bibitem{danet}
J.~Fu, J.~Liu, H.~Tian, Y.~Li, Y.~Bao, Z.~Fang, and H.~Lu, ``Dual attention network for scene segmentation,'' in \emph{CVPR}, 2019.

\bibitem{ccnet}
Z.~Huang, X.~Wang, L.~Huang, C.~Huang, Y.~Wei, and W.~Liu, ``{CCNet:} {C}riss-cross attention for semantic segmentation,'' in \emph{ICCV}, 2019.

\bibitem{ocnet}
Y.~Yuan, L.~Huang, J.~Guo, C.~Zhang, X.~Chen, and J.~Wang, ``{OCNet:} {Object} context for semantic segmentation,'' \emph{IJCV}, vol. 129, no.~8, pp. 2375--2398, 2021.

\bibitem{liu2020covariance}
Y.~Liu, Y.~Chen, P.~Lasang, and Q.~Sun, ``Covariance attention for semantic segmentation,'' \emph{TPAMI}, vol.~44, no.~4, pp. 1805--1818, 2022.

\bibitem{li2021ctnet}
Z.~Li, Y.~Sun, L.~Zhang, and J.~Tang, ``{CTNet:} {Context-based} tandem network for semantic segmentation,'' \emph{TPAMI}, vol.~44, no.~12, pp. 9904--9917, 2022.

\bibitem{vit}
A.~Dosovitskiy, L.~Beyer, A.~Kolesnikov, D.~Weissenborn, X.~Zhai, T.~Unterthiner, M.~Dehghani, M.~Minderer, G.~Heigold, S.~Gelly, J.~Uszkoreit, and N.~Houlsby, ``An image is worth 16x16 words: Transformers for image recognition at scale,'' in \emph{ICLR}, 2021.

\bibitem{touvron2021deit}
H.~Touvron, M.~Cord, M.~Douze, F.~Massa, A.~Sablayrolles, and H.~J{\'e}gou, ``Training data-efficient image transformers \& distillation through attention,'' in \emph{ICML}, 2021.

\bibitem{swin}
Z.~Liu, Y.~Lin, Y.~Cao, H.~Hu, Y.~Wei, Z.~Zhang, S.~Lin, and B.~Guo, ``Swin transformer: Hierarchical vision transformer using shifted windows,'' in \emph{ICCV}, 2021.

\bibitem{dong2022cswin}
X.~Dong, J.~Bao, D.~Chen, W.~Zhang, N.~Yu, L.~Yuan, D.~Chen, and B.~Guo, ``{CSWin} transformer: {A} general vision transformer backbone with cross-shaped windows,'' in \emph{CVPR}, 2022.

\bibitem{li2022contextual_transformer}
Y.~Li, T.~Yao, Y.~Pan, and T.~Mei, ``Contextual transformer networks for visual recognition,'' \emph{TPAMI}, vol.~45, no.~2, pp. 1489--1500, 2023.

\bibitem{wu2021p2t}
Y.-H. Wu, Y.~Liu, X.~Zhan, and M.-M. Cheng, ``{P2T:} {Pyramid} pooling transformer for scene understanding,'' \emph{TPAMI}, 2022.

\bibitem{segmenter}
R.~Strudel, R.~Garcia, I.~Laptev, and C.~Schmid, ``Segmenter: Transformer for semantic segmentation,'' in \emph{ICCV}, 2021.

\bibitem{segformer}
E.~Xie, W.~Wang, Z.~Yu, A.~Anandkumar, J.~M. Alvarez, and P.~Luo, ``{SegFormer:} {Simple} and efficient design for semantic segmentation with transformers,'' in \emph{NeurIPS}, 2021.

\bibitem{cheng2021maskformer}
B.~Cheng, A.~G. Schwing, and A.~Kirillov, ``Per-pixel classification is not all you need for semantic segmentation,'' in \emph{NeurIPS}, 2021.

\bibitem{gu2022hrvit}
J.~Gu, H.~Kwon, D.~Wang, W.~Ye, M.~Li, Y.-H. Chen, L.~Lai, V.~Chandra, and D.~Z. Pan, ``Multi-scale high-resolution vision transformer for semantic segmentation,'' in \emph{CVPR}, 2022.

\bibitem{mlp_mixer}
I.~O. Tolstikhin, N.~Houlsby, A.~Kolesnikov, L.~Beyer, X.~Zhai, T.~Unterthiner, J.~Yung, A.~Steiner, D.~Keysers, J.~Uszkoreit, M.~Lucic, and A.~Dosovitskiy, ``{MLP-mixer:} {An} {all-MLP} architecture for vision,'' in \emph{NeurIPS}, 2021.

\bibitem{hou2021vision}
Q.~Hou, Z.~Jiang, L.~Yuan, M.-M. Cheng, S.~Yan, and J.~Feng, ``Vision permutator: A permutable {MLP-like} architecture for visual recognition,'' \emph{TPAMI}, vol.~45, no.~1, pp. 1328--1334, 2023.

\bibitem{restricted}
L.~Deng, M.~Yang, H.~Li, T.~Li, B.~Hu, and C.~Wang, ``Restricted deformable convolution-based road scene semantic segmentation using surround view cameras,'' \emph{T-ITS}, vol.~21, no.~10, pp. 4350--4362, 2020.

\bibitem{woodscape}
S.~K. Yogamani, C.~Witt, H.~Rashed, S.~Nayak, S.~Mansoor, P.~Varley, X.~Perrotton, D.~O'Dea, P.~P{\'{e}}rez, C.~Hughes, J.~Horgan, G.~Sistu, S.~Chennupati, M.~Uric{\'{a}}r, S.~Milz, M.~Simon, and K.~Amende, ``{WoodScape:} {A} multi-task, multi-camera fisheye dataset for autonomous driving,'' in \emph{ICCV}, 2019.

\bibitem{petrovai2022semantic_cameras}
A.~Petrovai and S.~Nedevschi, ``Semantic cameras for 360-degree environment perception in automated urban driving,'' \emph{T-ITS}, vol.~23, no.~10, pp. 17\,271--17\,283, 2022.

\bibitem{synthetic}
Y.~Xu, K.~Wang, K.~Yang, D.~Sun, and J.~Fu, ``Semantic segmentation of panoramic images using a synthetic dataset,'' in \emph{SPIE}, 2019.

\bibitem{orhan2021semantic_outdoor_panoramic}
S.~Orhan and Y.~Bastanlar, ``Semantic segmentation of outdoor panoramic images,'' \emph{SIVP}, vol.~16, no.~3, pp. 643--650, 2022.

\bibitem{hu2022distortion_panoramic}
X.~Hu, Y.~An, C.~Shao, and H.~Hu, ``Distortion convolution module for semantic segmentation of panoramic images based on the image-forming principle,'' \emph{TIM}, vol.~71, pp. 1--12, 2022.

\bibitem{pps}
A.~Jaus, K.~Yang, and R.~Stiefelhagen, ``Panoramic panoptic segmentation: Towards complete surrounding understanding via unsupervised contrastive learning,'' in \emph{IV}, 2021.

\bibitem{mei2022waymo}
J.~Mei, A.~Z. Zhu, X.~Yan, H.~Yan, S.~Qiao, Y.~Zhu, L.-C. Chen, H.~Kretzschmar, and D.~Anguelov, ``Waymo open dataset: Panoramic video panoptic segmentation,'' in \emph{ECCV}, 2022.

\bibitem{pps_insights}
A.~Jaus, K.~Yang, and R.~Stiefelhagen, ``Panoramic panoptic segmentation: Insights into surrounding parsing for mobile agents via unsupervised contrastive learning,'' \emph{T-ITS}, vol.~24, no.~4, pp. 4438--4453, 2023.

\bibitem{distortion_aware}
K.~Tateno, N.~Navab, and F.~Tombari, ``Distortion-aware convolutional filters for dense prediction in panoramic images,'' in \emph{ECCV}, 2018.

\bibitem{spherical_unstructured_grids}
C.~M. Jiang, J.~Huang, K.~Kashinath, Prabhat, P.~Marcus, and M.~Nie{\ss}ner, ``Spherical {CNNs} on unstructured grids,'' in \emph{ICLR}, 2019.

\bibitem{spherephd}
Y.~Lee, J.~Jeong, J.~Yun, W.~Cho, and K.-J. Yoon, ``{SpherePHD:} {Applying} {CNNs} on a spherical {PolyHeDron} representation of 360{\textdegree} images,'' in \emph{CVPR}, 2019.

\bibitem{equivariant_networks}
M.~Shakerinava and S.~Ravanbakhsh, ``Equivariant networks for pixelized spheres,'' in \emph{ICML}, 2021.

\bibitem{zheng2022complementary_bidirectional}
Z.~Zheng, C.~Lin, L.~Nie, K.~Liao, Z.~Shen, and Y.~Zhao, ``Complementary bi-directional feature compression for indoor 360{\textdegree} semantic segmentation with self-distillation,'' \emph{arXiv preprint arXiv:2207.02437}, 2022.

\bibitem{pano_sfmlearner}
M.~Liu, S.~Wang, Y.~Guo, Y.~He, and H.~Xue, ``{Pano-SfMLearner:} {Self-Supervised} multi-task learning of depth and semantics in panoramic videos,'' \emph{SPL}, vol.~28, pp. 832--836, 2021.

\bibitem{zhang2021deeppanocontext}
C.~Zhang, Z.~Cui, C.~Chen, S.~Liu, B.~Zeng, H.~Bao, and Y.~Zhang, ``{DeepPanoContext:} {Panoramic} {3D} scene understanding with holistic scene context graph and relation-based optimization,'' in \emph{ICCV}, 2021.

\bibitem{xia2022dat}
Z.~Xia, X.~Pan, S.~Song, L.~E. Li, and G.~Huang, ``Vision transformer with deformable attention,'' in \emph{CVPR}, 2022.

\bibitem{yue2021psvit}
X.~Yue, S.~Sun, Z.~Kuang, M.~Wei, P.~H.~S. Torr, W.~Zhang, and D.~Lin, ``Vision transformer with progressive sampling,'' in \emph{ICCV}, 2021.

\bibitem{deformable_detr}
X.~Zhu, W.~Su, L.~Lu, B.~Li, X.~Wang, and J.~Dai, ``Deformable {DETR}: Deformable transformers for end-to-end object detection,'' in \emph{ICLR}, 2021.

\bibitem{wang2021not_dynamic}
Y.~Wang, R.~Huang, S.~Song, Z.~Huang, and G.~Huang, ``Not all images are worth 16x16 words: Dynamic vision transformers with adaptive sequence length,'' in \emph{NeurIPS}, 2021.

\bibitem{rao2021dynamicvit}
Y.~Rao, W.~Zhao, B.~Liu, J.~Lu, J.~Zhou, and C.-J. Hsieh, ``{DynamicViT:} {Efficient} vision transformers with dynamic token sparsification,'' in \emph{NeurIPS}, 2021.

\bibitem{yin2021avit}
H.~Yin, A.~Vahdat, J.~M. Alvarez, A.~Mallya, J.~Kautz, and P.~Molchanov, ``{A-ViT:} {Adaptive} tokens for efficient vision transformer,'' in \emph{CVPR}, 2022.

\bibitem{xu2021evo_vit}
Y.~Xu, Z.~Zhang, M.~Zhang, K.~Sheng, K.~Li, W.~Dong, L.~Zhang, C.~Xu, and X.~Sun, ``{Evo-ViT:} {Slow-fast} token evolution for dynamic vision transformer,'' in \emph{AAAI}, 2021.

\bibitem{liu2022dynamic_group_transformer}
K.~Liu, T.~Wu, C.~Liu, and G.~Guo, ``Dynamic group transformer: A general vision transformer backbone with dynamic group attention,'' in \emph{IJCAI}, 2022.

\bibitem{ros2016synthia}
G.~Ros, L.~Sellart, J.~Materzynska, D.~Vazquez, and A.~M. Lopez, ``The {SYNTHIA} dataset: {A} large collection of synthetic images for semantic segmentation of urban scenes,'' in \emph{CVPR}, 2016.

\bibitem{richter2016playing_gta5}
S.~R. Richter, V.~Vineet, S.~Roth, and V.~Koltun, ``Playing for data: Ground truth from computer games,'' in \emph{ECCV}, 2016.

\bibitem{curriculum_da}
Y.~Zhang, P.~David, and B.~Gong, ``Curriculum domain adaptation for semantic segmentation of urban scenes,'' in \emph{ICCV}, 2017.

\bibitem{li2022class_self_labeling}
R.~Li, S.~Li, C.~He, Y.~Zhang, X.~Jia, and L.~Zhang, ``Class-balanced pixel-level self-labeling for domain adaptive semantic segmentation,'' in \emph{CVPR}, 2022.

\bibitem{huo2022domain_agnostic_prior}
X.~Huo, L.~Xie, H.~Hu, W.~Zhou, H.~Li, and Q.~Tian, ``Domain-agnostic prior for transfer semantic segmentation,'' in \emph{CVPR}, 2022.

\bibitem{zhu2021improving_efficient_self_training}
Y.~Zhu, Z.~Zhang, C.~Wu, Z.~Zhang, T.~He, H.~Zhang, R.~Manmatha, M.~Li, and A.~J. Smola, ``Improving semantic segmentation via efficient self-training,'' \emph{TPAMI}, 2021.

\bibitem{lai2022decouplenet}
X.~Lai, Z.~Tian, X.~Xu, Y.~Chen, S.~Liu, H.~Zhao, L.~Wang, and J.~Jia, ``{DecoupleNet:} {Decoupled} network for domain adaptive semantic segmentation,'' in \emph{ECCV}, 2022.

\bibitem{xie2022sepico}
B.~Xie, S.~Li, M.~Li, C.~H. Liu, G.~Huang, and G.~Wang, ``{SePiCo:} {Semantic-guided} pixel contrast for domain adaptive semantic segmentation,'' \emph{TPAMI}, vol.~45, no.~7, pp. 9004--9021, 2023.

\bibitem{cycada}
J.~Hoffman, E.~Tzeng, T.~Park, J.~Zhu, P.~Isola, K.~Saenko, A.~A. Efros, and T.~Darrell, ``{CyCADA:} {Cycle-consistent} adversarial domain adaptation,'' in \emph{ICML}, 2018.

\bibitem{adaptsegnet}
Y.-H. Tsai, W.-C. Hung, S.~Schulter, K.~Sohn, M.-H. Yang, and M.~Chandraker, ``Learning to adapt structured output space for semantic segmentation,'' in \emph{CVPR}, 2018.

\bibitem{all_about_structure}
W.-L. Chang, H.-P. Wang, W.-H. Peng, and W.-C. Chiu, ``All about structure: Adapting structural information across domains for boosting semantic segmentation,'' in \emph{CVPR}, 2019.

\bibitem{pycda}
Q.~Lian, L.~Duan, F.~Lv, and B.~Gong, ``Constructing self-motivated pyramid curriculums for cross-domain semantic segmentation: A non-adversarial approach,'' in \emph{ICCV}, 2019.

\bibitem{gan}
I.~J. Goodfellow, J.~Pouget{-}Abadie, M.~Mirza, B.~Xu, D.~Warde{-}Farley, S.~Ozair, A.~C. Courville, and Y.~Bengio, ``Generative adversarial nets,'' in \emph{NeurIPS}, 2014.

\bibitem{li2019bidirectional}
Y.~Li, L.~Yuan, and N.~Vasconcelos, ``Bidirectional learning for domain adaptation of semantic segmentation,'' in \emph{CVPR}, 2019.

\bibitem{contextual_relation_consistent_da}
J.~Huang, S.~Lu, D.~Guan, and X.~Zhang, ``Contextual-relation consistent domain adaptation for semantic segmentation,'' in \emph{ECCV}, 2020.

\bibitem{fda}
Y.~Yang and S.~Soatto, ``{FDA:} {Fourier} domain adaptation for semantic segmentation,'' in \emph{CVPR}, 2020.

\bibitem{maximum_squares_loss}
M.~Chen, H.~Xue, and D.~Cai, ``Domain adaptation for semantic segmentation with maximum squares loss,'' in \emph{ICCV}, 2019.

\bibitem{wang2020differential}
Z.~Wang, M.~Yu, Y.~Wei, R.~Feris, J.~Xiong, W.~Hwu, T.~S. Huang, and H.~Shi, ``Differential treatment for stuff and things: A simple unsupervised domain adaptation method for semantic segmentation,'' in \emph{CVPR}, 2020.

\bibitem{jiang2022proca}
Z.~Jiang, Y.~Li, C.~Yang, P.~Gao, Y.~Wang, Y.~Tai, and C.~Wang, ``Prototypical contrast adaptation for domain adaptive semantic segmentation,'' in \emph{ECCV}, 2022.

\bibitem{intra_da}
F.~Pan, I.~Shin, F.~Rameau, S.~Lee, and I.~S. Kweon, ``Unsupervised intra-domain adaptation for semantic segmentation through self-supervision,'' in \emph{CVPR}, 2020.

\bibitem{gu2021pit}
Q.~Gu, Q.~Zhou, M.~Xu, Z.~Feng, G.~Cheng, X.~Lu, J.~Shi, and L.~Ma, ``{PIT:} {Position-invariant} transform for {cross-FoV} domain adaptation,'' in \emph{ICCV}, 2021.

\bibitem{yue2021pcs}
X.~Yue, Z.~Zheng, S.~Zhang, Y.~Gao, T.~Darrell, K.~Keutzer, and A.~L. Sangiovanni{-}Vincentelli, ``Prototypical cross-domain self-supervised learning for few-shot unsupervised domain adaptation,'' in \emph{CVPR}, 2021.

\bibitem{proda}
P.~Zhang, B.~Zhang, T.~Zhang, D.~Chen, Y.~Wang, and F.~Wen, ``Prototypical pseudo label denoising and target structure learning for domain adaptive semantic segmentation,'' in \emph{CVPR}, 2021.

\bibitem{resnet}
K.~He, X.~Zhang, S.~Ren, and J.~Sun, ``Deep residual learning for image recognition,'' in \emph{CVPR}, 2016.

\bibitem{lai2017semantic}
W.-S. Lai, Y.~Huang, N.~Joshi, C.~Buehler, M.-H. Yang, and S.~B. Kang, ``Semantic-driven generation of hyperlapse from 360 degree video,'' \emph{TVCG}, vol.~24, no.~9, pp. 2610--2621, 2018.

\bibitem{dai2017deformable}
J.~Dai, H.~Qi, Y.~Xiong, Y.~Li, G.~Zhang, H.~Hu, and Y.~Wei, ``Deformable convolutional networks,'' in \emph{ICCV}, 2017.

\bibitem{hu2018SE_net}
J.~Hu, L.~Shen, and G.~Sun, ``Squeeze-and-excitation networks,'' in \emph{CVPR}, 2018.

\bibitem{chen2020simclr_v2}
T.~Chen, S.~Kornblith, K.~Swersky, M.~Norouzi, and G.~Hinton, ``Big self-supervised models are strong semi-supervised learners,'' in \emph{NeurIPS}, 2020.

\bibitem{liao2022kitti360}
Y.~Liao, J.~Xie, and A.~Geiger, ``{KITTI-360:} {A} novel dataset and benchmarks for urban scene understanding in {2D} and {3D},'' \emph{TPAMI}, vol.~45, no.~3, pp. 3292--3310, 2023.

\bibitem{testolina2022selma}
P.~Testolina, F.~Barbato, U.~Michieli, M.~Giordani, P.~Zanuttigh, and M.~Zorzi, ``{SELMA:} {Semantic} large-scale multimodal acquisitions in variable weather, daytime and viewpoints,'' \emph{T-ITS}, 2023.

\bibitem{sekkat2022synwoodscape}
A.~R. Sekkat, Y.~Dupuis, V.~R. Kumar, H.~Rashed, S.~K. Yogamani, P.~Vasseur, and P.~Honeine, ``{SynWoodScape:} {Synthetic} surround-view fisheye camera dataset for autonomous driving,'' \emph{RA-L}, vol.~7, no.~3, pp. 8502--8509, 2022.

\bibitem{omniscape}
A.~R. Sekkat, Y.~Dupuis, P.~Vasseur, and P.~Honeine, ``The {OmniScape} dataset,'' in \emph{ICRA}, 2020.

\bibitem{carla}
A.~Dosovitskiy, G.~Ros, F.~Codevilla, A.~Lopez, and V.~Koltun, ``{CARLA:} {An} open urban driving simulator,'' in \emph{CoRL}, 2017.

\bibitem{adam}
D.~P. Kingma and J.~Ba, ``Adam: A method for stochastic optimization,'' in \emph{ICLR}, 2015.

\bibitem{fastscnn}
R.~P.~K. Poudel, S.~Liwicki, and R.~Cipolla, ``{Fast-SCNN:} {F}ast semantic segmentation network,'' in \emph{BMVC}, 2019.

\bibitem{swiftnet}
M.~Orsic, I.~Kreso, P.~Bevandic, and S.~Segvic, ``In defense of pre-trained {ImageNet} architectures for real-time semantic segmentation of road-driving images,'' in \emph{CVPR}, 2019.

\bibitem{erfnet}
E.~Romera, J.~M. Alvarez, L.~M. Bergasa, and R.~Arroyo, ``{ERFNet:} {Efficient} residual factorized {ConvNet} for real-time semantic segmentation,'' \emph{T-ITS}, vol.~19, no.~1, pp. 263--272, 2018.

\bibitem{fanet}
P.~Hu, F.~Perazzi, F.~C. Heilbron, O.~Wang, Z.~Lin, K.~Saenko, and S.~Sclaroff, ``Real-time semantic segmentation with fast attention,'' \emph{RA-L}, vol.~6, no.~1, pp. 263--270, 2021.

\bibitem{ocrnet}
Y.~Yuan, X.~Chen, and J.~Wang, ``Object-contextual representations for semantic segmentation,'' in \emph{ECCV}, 2020.

\bibitem{dnl}
M.~Yin, Z.~Yao, Y.~Cao, X.~Li, Z.~Zhang, S.~Lin, and H.~Hu, ``Disentangled non-local neural networks,'' in \emph{ECCV}, 2020.

\bibitem{panopticfpn}
A.~Kirillov, R.~Girshick, K.~He, and P.~Doll{\'a}r, ``Panoptic feature pyramid networks,'' in \emph{CVPR}, 2019.

\bibitem{resnest}
H.~Zhang, C.~Wu, Z.~Zhang, Y.~Zhu, Z.~Zhang, H.~Lin, Y.~Sun, T.~He, J.~Mueller, R.~Manmatha, M.~Li, and A.~J. Smola, ``{ResNeSt:} {Split-attention} networks,'' in \emph{CVPRW}, 2022.

\bibitem{cheng2022mask2former}
B.~Cheng, I.~Misra, A.~G. Schwing, A.~Kirillov, and R.~Girdhar, ``Masked-attention mask transformer for universal image segmentation,'' in \emph{CVPR}, 2022.

\bibitem{zhang2021trans4trans_acvr}
J.~Zhang, K.~Yang, A.~Constantinescu, K.~Peng, K.~M{\"u}ller, and R.~Stiefelhagen, ``{Trans4Trans:} {Efficient} transformer for transparent object segmentation to help visually impaired people navigate in the real world,'' in \emph{ICCVW}, 2021.

\bibitem{hoyer2023mic}
L.~Hoyer, D.~Dai, H.~Wang, and L.~Van~Gool, ``{MIC:} {Masked} image consistency for context-enhanced domain adaptation,'' in \emph{CVPR}, 2023.

\bibitem{hoyer2022hrda}
L.~Hoyer, D.~Dai, and L.~Van~Gool, ``{HRDA:} {Context-aware} high-resolution domain-adaptive semantic segmentation,'' in \emph{ECCV}, 2022.

\bibitem{usss}
T.~Kalluri, G.~Varma, M.~Chandraker, and C.~V. Jawahar, ``Universal semi-supervised semantic segmentation,'' in \emph{ICCV}, 2019.

\bibitem{seamless}
L.~Porzi, S.~R. Bul{\`o}, A.~Colovic, and P.~Kontschieder, ``Seamless scene segmentation,'' in \emph{CVPR}, 2019.

\bibitem{issafe}
J.~Zhang, K.~Yang, and R.~Stiefelhagen, ``{ISSAFE:} {I}mproving semantic segmentation in accidents by fusing event-based data,'' in \emph{IROS}, 2020.

\bibitem{gauge_equivariant}
T.~Cohen, M.~Weiler, B.~Kicanaoglu, and M.~Welling, ``Gauge equivariant convolutional networks and the icosahedral {CNN},'' in \emph{ICML}, 2019.

\bibitem{orientation}
C.~Zhang, S.~Liwicki, W.~Smith, and R.~Cipolla, ``Orientation-aware semantic segmentation on icosahedron spheres,'' in \emph{ICCV}, 2019.

\bibitem{tangent}
M.~Eder, M.~Shvets, J.~Lim, and J.-M. Frahm, ``Tangent images for mitigating spherical distortion,'' in \emph{CVPR}, 2020.

\bibitem{bdd}
F.~Yu, H.~Chen, X.~Wang, W.~Xian, Y.~Chen, F.~Liu, V.~Madhavan, and T.~Darrell, ``{BDD100K:} {A} diverse driving dataset for heterogeneous multitask learning,'' in \emph{CVPR}, 2020.

\bibitem{pvtv2}
W.~Wang, E.~Xie, X.~Li, D.~Fan, K.~Song, D.~Liang, T.~Lu, P.~Luo, and L.~Shao, ``{PVT v2:} {Improved} baselines with pyramid vision transformer,'' \emph{CVM}, vol.~8, no.~3, pp. 415--424, 2022.

\bibitem{unet}
O.~Ronneberger, P.~Fischer, and T.~Brox, ``{U-net:} convolutional networks for biomedical image segmentation,'' in \emph{MICCAI}, 2015.

\end{thebibliography}
